\documentclass[12pt,journal,draftcls,a4paper,twoside,onecolumn]{IEEEtran}
\usepackage{dsfont}
\usepackage{amsmath}
\usepackage{amssymb}
\usepackage{amsfonts}
\usepackage{graphicx}
\usepackage{amsmath}
\usepackage[all,poly]{xy}
\usepackage{multirow}
\usepackage{color}
\usepackage[noadjust]{cite}
\usepackage{subfigure}
\usepackage{mathrsfs}
\usepackage{algorithmic}
\usepackage[linesnumbered,lined,ruled]{algorithm2e}
\usepackage{cancel}
\usepackage{url}
\usepackage{lipsum}
\usepackage{stfloats}
\usepackage{epsfig}
\usepackage{bm,balance}
\usepackage{amsthm}
\usepackage{gensymb}
\usepackage{url}

\usepackage{epstopdf}

\usepackage{setspace}
\newtheorem*{remark*}{Remark}

\newcommand{\bs}{\boldsymbol}

\def\bfa{{\mathbf{a}}}

\def\bfe{{\mathbf{e}}}

\def\bfn{{\mathbf{n}}}

\def\bfx{{\mathbf{x}}}
\def\bfy{{\mathbf{y}}}

\def\bfA{{\mathbf{A}}}

\def\bfE{{\mathbf{E}}}

\def\bfM{{\mathbf{M}}}
\def\bfN{{\mathbf{N}}}

\def\bfP{{\mathbf{P}}}

\def\bfX{{\mathbf{X}}}

\def\calA{{\mathcal{A}}}

\def\calO{{\mathcal{O}}}

\newcommand{\argmin}{\mathrm{arg}\min}

\newcounter{algo}
\renewcommand{\thealgo}{\arabic{algo}}

\graphicspath{{figures/}}

\title{Unsupervised Nonlinear Spectral Unmixing based on a Multilinear Mixing Model}
\author{\IEEEauthorblockN{Qi Wei, \IEEEmembership{Member,~IEEE},
Marcus Chen, \IEEEmembership{Member,~IEEE},\\
Jean-Yves Tourneret, \IEEEmembership{Senior Member,~IEEE},
Simon Godsill, \IEEEmembership{Member,~IEEE}}
\thanks{Qi Wei and Simon Godsill are with Department of Engineering, University of Cambridge, CB2 1PZ, Cambridge, UK (e-mail: \{qi.wei, sjg\}@eng.cam.ac.uk).
Marcus Chen is with the Department of Mathematics, Nanyang Technological University, Singapore, 639798 (e-mail: marcuschen@pmail.ntu.edu.sg).
Jean-Yves Tourneret is with IRIT/INP-ENSEEIHT, University of Toulouse, Toulouse, France (e-mail: jean-yves.tourneret@enseeiht.fr).}
}

\begin{document}
\maketitle

\begin{abstract}
In the community of remote sensing, nonlinear mixing models 
have recently received particular attention in hyperspectral image processing.
In this paper, we present a novel nonlinear spectral unmixing method following the 
recent multilinear mixing model of \cite{Heylen2016MLM}, which includes an 
infinite number of terms related to interactions between different endmembers. The proposed unmixing 
method is unsupervised in the sense that the endmembers are estimated jointly with 
the abundances and other parameters of interest, i.e., the transition probability 
of undergoing further interactions. Non-negativity and  sum-to-one constraints
are imposed on abundances while only non-negativity is considered for endmembers. The resulting unmixing 
problem is formulated as a constrained nonlinear optimization problem, which is solved by a block coordinate
descent strategy, consisting of updating the endmembers, abundances and transition probability iteratively.
The proposed method is evaluated and compared with linear unmixing methods for synthetic and
real hyperspectral datasets acquired by the AVIRIS sensor. The advantage of using non-linear unmixing as opposed
to linear unmixing is clearly shown in these examples.
\end{abstract}


\section{Introduction}
\label{sec:intro}



Spectral unmixing (SU) aims at decomposing a set of $n$ multivariate measurements (or pixel vectors)
$\bfX = \left[\bfx^1,\ldots,\bfx^{n}\right]$ into a collection of $m$ elementary signatures
$\bfE=\left[\bfe_1,\cdots,\bfe_m\right]$, usually referred to as \emph{endmembers}, and estimating
the relative proportions $\bfA=\left[\bfa^1,\ldots,\bfa^n\right]$ of these signatures, called \emph{abundances}. 
SU has been advocated as a relevant multivariate analysis technique in various applicative areas, including remote sensing \cite{Averbuch2012}, planetology \cite{Themelis2012pss}, microscopy \cite{Dobigeon2012ultra}, spectroscopy \cite{Carteret2009} and gene expression analysis \cite{Huang2011plos}. In particular, a great interest has been demonstrated when analyzing multi-band (e.g., hyperspectral) images, for instance for pixel classification \cite{Chang1998b}, material quantification \cite{Wang2006} and subpixel detection \cite{Manolakis2001}.

\subsection{Linear Mixture Model}
\label{sec:proposed}
The linear mixture model (LMM) assumes that each image pixel is a linear combination of all the endmembers present in this pixel.
The LMM model has been widely used in the remote sensing community and can be expressed as (see, e.g., \cite{Keshava2002})

\begin{equation}
\label{eq:obs_general}
\bfx^i =  \bfE \bfa^i + \bfn^i
\end{equation}
where
\begin{itemize}
\item $\bfx^i$ is a $d \times 1$ vector representing the measured reflectance for the $i$th pixel,
\item $\bfE \in\mathbb{R}^{d \times m}$ is a non-negative matrix whose columns 
$\bfe_1,\cdots,\bfe_m$ correspond to $m$ endmember signatures and span the space
where the data $\bfx^1,\ldots,\bfx^{n}$ reside,
\item $\bfa^i$ is an $m \times 1$ non-negative vector, which includes the fractional abundances (coefficients) 
for the $i$th pixel (that sum to $1$),
\item $\bfn^i \in \mathbb{R}^{d}$ is the additive Gaussian noise.
\end{itemize}

By arranging all pixels of the observed scenario lexicographically, the LMM model can be written as 
\begin{equation}
\label{eq:obs_general}
\bf X =  EA + N
\end{equation}
where $\bfX \in \mathbb{R}^{d \times n}$, $\bfA \in \mathbb{R}^{m \times n}$ and $\bfN \in \mathbb{R}^{d \times n}$
are the reflectance, abundance and noise matrices, $n$ is the number of observations, and $d$ is the
number of spectral bands.

The spectral unmixing problem based on the LMM is generally formulated as the following constrained
least squares problem.
\begin{align}
  \label{eq:fcld_vector}
  \begin{aligned}
  &\min_{\bfa^i}  \Vert{{\bfx^i - \bfE \bfa^i}}\Vert_2^2\\
   & \text{subject to (s.t.)}  \quad \bfa^i \geq 0 \quad \textrm{and} \quad {\bf 1}^T_m{\bfa^i} = 1
   \end{aligned}
\end{align}
where ${\bf 1}_m$ is an $m \times 1$ vector with all ones. Using all observed data and matrix notations, 
the optimization problem can be written as
\begin{align}
  \label{eq:fcld_matrix}
  \begin{aligned}
  &\min_{\bf A}  \Vert{{\bf X - E A}}\Vert_F^2\\
   & \text{s.t.}  \quad \textbf{A}\geq 0 \quad \textrm{and} \quad{\bf 1}^T_m{\bf A} = {\bf 1}_n^T
   \end{aligned}
\end{align}
where $\textbf{A}\geq 0$ has to be understood in the element-wise sense, 
meaning that all the coefficients are non-negative. Note
that $\|\cdot\|_F$ is the Frobenius norm,
which is defined as 
\begin{equation*}
\|\bfX\|_F=\sqrt{\text{trace}(\bfX^H \bfX)}
\end{equation*}
where $\bfX^H$ denotes the conjugate transpose of $\bfX$ and trace($\bfM$)
is the trace of the matrix $\bfM$ \cite{Horn2012}. 

\subsection{Nonlinear Mixture Model}
Due to its simple and intuitive physical interpretation as well as
tractable estimation process, the LMM has been widely used for unmixing, 
and has shown interesting results in various applications.
However, there exist many scenarios in which the LMM is not appropriate
and can be advantageously replaced by a nonlinear mixing model \cite{Dobigeon2014,Heylen2016MLM}.
One notable example is the case of scenes with large geometrical structures such as
buildings or trees, where shadowing and mutual illumination involve 
multiple light scattering effects.
Another example is the case of mineral mixtures (also referred to as intimate mixtures),
where an incoming light ray can interact many times with the different mineral
grains, and the single interactions assumed in the LMM can
even become relatively rare. Furthermore, the LMM only
considers reflection and disregards optical transmission, which
can become quite important in vegetation and mineral mixtures.
To solve these problems, nonlinear mixture models have been 
proposed as interesting alternatives to overcome the inherent 
limitations of the LMM. These models include the Hapke model \cite{Hapke1981bidirectional},
the generalized bilinear model (GBM) \cite{Halimi2011,Yokoya2014nonlinear},
the linear-quadratic model \cite{Meganem2014linear},
the post-nonlinear mixing model (PNMM) \cite{Altmann2012supervised} and the multi-linear mixing (MLM) 
model \cite{Heylen2016MLM}.

In this work, we focus on the recently proposed MLM model \cite{Heylen2016MLM} as follows
\begin{equation}
\begin{split}
\bfx^i &=(1-P^i)\bfy^i+(1-P^i)P^i(\bfy^i \odot \bfy^i)\\
& +(1-P^i)(P^i)^2(\bfy^i \odot \bfy^i\odot \bfy^i)\cdots
\label{eq:MLM_org}
\end{split}
\end{equation}
where $\bfx^i$ represents the observed reflectance, $\bfy^i = \bfE \bfa^i$ is the linear term used in the traditional LMM model,
$\odot$ represents the Hadamard entry-wise product and $P^i$ represents the probability of undergoing further interactions
after each interaction with a material. Thus, $1-P_i$ corresponds to 
the probability of escaping the scene and reaching the observer. 
The MLM model is the first nonlinear model that includes all orders of interactions 
by introducing only a single parameter $P^i$, which describes the probability of further
interactions. Furthermore, the summation in \eqref{eq:MLM_org} can be conveniently simplified as the following fixed-point equation 
\begin{equation}
\bfx^i = (1-P^i)\bfy^i+P^i \bfy^i \odot \bfx^i.
\label{eq:MLM}
\end{equation}
Note that $P^i$ is different from pixel to pixel. For more details concerning the
derivation of the MLM, we refer the reader to 
\cite{Heylen2016MLM}. To achieve nonlinear unmixing, the authors of \cite{Heylen2016MLM}
considered the following optimization problem
\begin{equation}
\begin{split}
&\argmin_{\{\bfa^i, P^i\}} \|{\bfx^i-\frac{(1-P^i)\bfy^i}{1-P^i\bfy^i}}\|_2^2\\
&\textrm{ s.t. } \quad  \bfy^i=\bfE\bfa^i.
\end{split}
\label{eq:Heylen_objective}
\end{equation}
The endmember matrix $\bfE$ was suggested to be estimated using VCA \cite{Nascimento2005},
which is one of the state-of-the-art endmember extraction methods, and to be fixed in the unmixing,
leading to a supervised unmixing method. However, the VCA algorithm is based on the LMM model,
which is different from the MLM. Furthermore, the optimization w.r.t. $\bfa^i$ and $P^i$ is
highly nonlinear and nonconvex, preventing a unique solution from being obtained. 

To overcome the difficulties mentioned above, this work considers three main modifications with respect to
the method in \cite{Heylen2016MLM}. First, the objective function is slightly changed from \eqref{eq:Heylen_objective}, in
order to avoid its highly nonlinearity and nonconvexity w.r.t. the parameters to be estimated. 
This modification significantly decreases the complexity of the optimization problem \eqref{eq:Heylen_objective},
which will be illustrated later. Second, instead of fixing the endmember matrix using VCA, the output of VCA is used as 
an initialization of an algorithm, which estimates the endmember matrix jointly with
the abundances and the transition probability, leading to an unsupervised
nonlinear mixing strategy. Finally, the parameter $P_i$ is constrained to belong to 
the interval [0,1], which is in agreement with its probability interpretation.

\section{Nonlinear Spectral Unmixing: A BCD scheme}
The nonlinear unmixing problem investigated in this work can be formulated as the 
following optimization problem
\begin{equation}
\argmin_{\{\bfa^i, P^i\}_{i=1}^m,\bfE} L(\bfE, \bfA,\bfP)
\label{eq:objective}
\end{equation}
\begin{equation*}
\begin{split}
\textrm{ with } \quad  L(\bfE, \bfA,\bfP) &= \sum_{i=1}^{n}\|\bfx^i - (1-P^i)\bfy^i - P^i\bfy^i \odot \bfx^i \|_2^2\\
\bfy^i&=\bfE\bfa^i \\
\bfa^i &\geq 0 \quad \textrm{and} \quad {\bf 1}^T_m{\bfa^i} = 1\\
0 &\leq \bfE \leq 1\\
0 &\leq P^i \leq 1.
\end{split}
\end{equation*}

To solve the problem \eqref{eq:objective}, we propose to update $\bfa^i$, $\bfE$
and $\bfP^i$ alternately, using a block coordinate descent (BCD) strategy.
Even though \eqref{eq:objective} is a nonconvex problem w.r.t. $\bfa^i$, $\bfE$ and $\bfP^i$ jointly,
it is interesting to note that each sub-problem turns out to be a convex problem which has a 
unique solution. The BCD algorithm is known to converge to a stationary point of the objective function 
to be optimized provided that this objective function has a unique minimum point w.r.t. each 
variable \cite[Prop. 2.7.1]{Bertsekas1999}, which is the case for the criterion in \eqref{eq:objective}. 
Thus, the BCD algorithm introduced in this paper converges to a stationary point of \eqref{eq:objective}. 
Note that the nonlinear unmixing problem \eqref{eq:objective} includes 
a linear unmixing (abundance estimation) step, an endmember extraction step and a 
transition probability estimation step. To ease the notation, we omit the upper indices $i$ 
for $\bfa$ and $\bfP$ hereafter as they can be updated pixel by pixel in parallel.
It is worthy of note that one popular strategy to overcome the nonconvexity could be 
simulation based methods such as Markov Chain Monte Carlo (see \cite{Pereyra2016} for a recent review).
Such an approach would be computationally intensive, but could potentially yield improvement 
in performance and better estimation of the uncertainty inherent in the problem.
However, the major drawback of being computationally expensive for simulation based methods
prevents their effective use in this application.

\subsection{Optimization w.r.t. $\bfa$}
The optimization w.r.t. $\bfa$ is now expressed as 
\begin{equation}
\argmin_{\bfa} \|\bfx - (1-P)\bfE \bfa - P \left(\bfE \bfa\right) \odot \bfx \|_2^2
\end{equation}
\begin{equation*}
\textrm{ s.t. }  \bfa \geq 0 \quad \textrm{and} \quad {\bf 1}^T_m{\bfa} = 1.
\end{equation*}

Straightforward computations lead to the following equivalent optimization problem
\begin{equation}
\argmin_{\bfa} \|\bfx - \tilde{\bfE} \bfa \|_2^2 \textrm{\quad s.t. \quad}  \bfa \geq 0 \quad \textrm{and} \quad {\bf 1}^T_m{\bfa} = 1
\label{eq:opt_a_FCLS}
\end{equation}
where $\tilde{\bfE}= \bfE \odot \left[(1-P) {\bf 1}_{d\times m} +P \bfx {\bf 1}^T_m \right]$.
Thus, the optimization w.r.t. $\bfa$ becomes a standard fully constrained least squares (FCLS) problem 
with a modified endmember matrix $\tilde{\bfE}$. To solve this classical convex 
problem, there exist plenty of methods, e.g., active-set \cite{Heinz2001}, 
ADMM \cite{Boyd2011} (sometimes referred to as SUNSAL \cite{Bioucas2010SUNSAL}),
projection-based methods \cite{Wei2015FastUnmixing}, etc. 
Instead of solving \eqref{eq:opt_a_FCLS} exactly,
we use the gradient projection method \cite{Calamai1987,Combettes2005} to decrease the objective
function defined in \eqref{eq:opt_a_FCLS}. More specifically, by denoting $g(\bfa)=\|\bfx - \tilde{\bfE} \bfa \|_2^2$,
\eqref{eq:opt_a_FCLS} can be rewritten as
\begin{equation}
\argmin_{\bfa} g(\bfa) \textrm{\quad s.t. \quad}  \bfa \in \calA
\end{equation}
where $\calA = \left\{\bfa \in \mathbb{R}^{m} |  \bfa \geq 0 \textrm{ and } {\bf 1}^T_m{\bfa} = 1 \right\}$.
Thus, the gradient of the objective $g(\bfa)$ w.r.t. $\bfa$ can be calculated as 
\begin{equation*}
\nabla_{\bfa} g(\bfa)= \tilde{\bfE}^T(\tilde{\bfE}\bfa-\bfx).
\end{equation*}
Note that the gradient projection method is different from the conventional gradient descent method 
in that each update after a move along the gradient direction $\nabla_{\bfa} g(\bfa)$
is projected onto the convex set $\calA$ to force all the updates to belong to the set of feasible solutions, i.e.,
\begin{equation}
{\bfa}  = \Pi_{\calA} \left(\bfa-\gamma_{\bfa}\nabla_{\bfa} g(\bfa)\right), \quad \varepsilon \leq \gamma_{\bfa} \leq 2/L_{\bfa}-\varepsilon
\label{eq:UpdateA}
\end{equation}
where $\Pi_{\calA}$ denotes the projection operator onto $\calA$,
$\varepsilon \in ]0,\textrm{min}\{1,1/L_{\bfa}\}[$ and $L_{\bfa}=\|\tilde{\bfE}^T \tilde{\bfE}\|_F$ is the Lipschitz constant of $\nabla_{\bfa} g(\bfa)$.
The projection onto the (canonical) simplex $\calA$ can be achieved with a finite algorithm\footnote{A finite algorithm is 
an iterative algorithm which converges in a finite number of steps.}, such as Michelot \cite{Michelot1986},
Duchi \emph{et al.} \cite{Duchi2008}, Condat \cite{Condat2014Fast}, etc. 

The motivation to use this gradient projection algorithm is two-fold. First, the convergence of 
a gradient projection within a BCD scheme is guaranteed (see more details in \cite{Beck2013,Bolte2014,Wright2015}). 
Second, the update \eqref{eq:UpdateA} is less computationally intensive than solving the optimization problem \eqref{eq:opt_a_FCLS} exactly 
which requires iterative updates. In this work, the stepsize $\gamma_{\bfa}$ is fixed to $1/L_{\bfa}$ to
ensure a sufficient decrease of the objective value per iteration. 
The updating scheme for $\bfa$ is summarized in Algorithm \ref{Algo:UpdateA}.
The computational complexity to calculate the abundances for all pixels is of the order $\calO(\max\{d,m\}nm)$.



\begin{algorithm}[h!]
\KwIn{$\bfa$, $\bfE$, $\bfx$, $P$}
\tcc{Calculate the modified endmembers (pixel-wised)}
$\tilde{\bfE} \leftarrow \bfE \odot \left((1-P) {\bf 1}_{d\times m} +P \bfx {\bf 1}^T_m \right)$\;
\tcc{Calculate the Lipschitz constant}
$L_{\bfa} \leftarrow \|\tilde{\bfE}^T \tilde{\bfE}\|_F$\;
\tcc{Gradient projection update}
$\hat{\bfa} \leftarrow \Pi_{\calA} \left(\bfa-\nabla_{\bfa} g(\bfa)/L_{\bfa}\right)$\;
\KwOut{$\hat{\bfa}$}
\caption{Minimization w.r.t. $\bfa$}
\label{Algo:UpdateA}
\end{algorithm}

\subsection{Optimization w.r.t. $P$}
The optimization w.r.t. $P$ can be formulated as 
\begin{equation}
\argmin_{P} \|\bfx - (1-P)\bfy - P\bfy \odot \bfx \|_2^2
\label{eq:opt_P}
\end{equation}
\begin{equation*}
\textrm{ s.t.  } 0 \leq P \leq 1.
\end{equation*}
Obviously, problem \eqref{eq:opt_P} is convex and admits the following closed-form solution
\begin{equation}
\hat{P}=\Pi_{{[0,1]}}\left(\frac{(\bfy-\bfy \odot \bfx)^T(\bfy-\bfx)}{\|\bfy-\bfy \odot \bfx\|_2^2}\right)
\label{eq:sol_P}
\end{equation}
where $[0,1]$ 
corresponds to the box constraints
and $\bf y=Ea$. Furthermore, it is interesting to note that if $0 \leq \bfy \leq 1$, 
the unconstrained solution $\frac{(\bfy-\bfy \odot \bfx)^T(\bfy-\bfx)}{\|\bfy-\bfy \odot \bfx\|_2^2}$ satisfies the box constraint
automatically. Thus, the updates of $P$ can be simplified as 
\begin{equation}
\hat{P}=\frac{(\bfy-\bfy \odot \bfx)^T(\bfy-\bfx)}{\|\bfy-\bfy \odot \bfx\|_2^2}.
\label{eq:sol_P_simply}
\end{equation}
The computational complexity to calculate the probability $P$ for all pixels is of the order $\calO(nd)$.

\subsection{Optimization w.r.t. $\bfE$}
\label{subsec:opt_E}
The optimization of the objective function in \eqref{eq:objective} w.r.t. $\bfE$ can be formulated as
\begin{equation}
\argmin_{\bfE} \|\bfx - (1-P)\bfE \bfa - P \left(\bfE \bfa\right) \odot \bfx \|_2^2
\end{equation}
\begin{equation*}
\textrm{ s.t. }  0 \leq \bfE \leq 1.
\end{equation*}
The above problem can be equivalently rewritten as
\begin{equation}
\argmin_{\bfE} \|\bfx - \left(\bfE \odot \tilde{\bfA}\right){\bf 1}_m\|_2^2 \textrm{\quad s.t. \quad }  0 \leq \bfE \leq 1
\end{equation}
where $\tilde{\bfA}=\left((1-P) {\bf 1}_d  +P \bfx\right)\bfa^T$. Considering all the observed pixels leads to 
\begin{equation}
\argmin_{\bfE}  f(\bfE)  \textrm{\quad s.t. \quad }  0 \leq \bfE \leq 1
\label{eq:obj_E}
\end{equation}
where 
$f(\bfE)=\sum_{i=1}^{n}\|\bfx^i - \left(\bfE \odot \tilde{\bfA}^i \right){\bf 1}_m\|_2^2$
and $\tilde{\bfA}^i=\left((1-P^i) {\bf 1}_d  +P^i \bfx^i\right){(\bfa^{i})^T}$.
The gradient of the objective $f(\bfE)$ can therefore be calculated as follows
\begin{equation}
\begin{split}
\nabla_{\bfE} f(\bfE) &=\sum_{i=1}^n \left[\left(\bfE \odot \tilde{\bfA}^i\right) {\bs 1}_{m} - \bfx^i\right]  {\bs 1}_{m}^T \odot \tilde{\bfA}^i\\
&=\sum_{i=1}^{n}
\left[
\begin{array}{cc}
(\bfe_1 \odot \tilde{\bfa}^i_1 \odot \tilde{\bfa}^i_1 + \cdots + \bfe_m \odot \tilde{\bfa}^i_m \odot \tilde{\bfa}^i_1)^T \\
(\bfe_1 \odot \tilde{\bfa}^i_1 \odot \tilde{\bfa}^i_2 + \cdots + \bfe_m \odot \tilde{\bfa}^i_m \odot \tilde{\bfa}^i_2)^T \\
\vdots \\
(\bfe_1 \odot \tilde{\bfa}^i_1 \odot \tilde{\bfa}^i_m + \cdots + \bfe_m \odot \tilde{\bfa}^i_m \odot \tilde{\bfa}^i_m )^T
\end{array}
\right]^T\\
&-
\sum_{i=1}^{n}\left[
\begin{array}{cc}
(\bfx^i \odot \tilde{\bfa}^i_1)^T \\
(\bfx^i \odot \tilde{\bfa}^i_2)^T \\
\vdots \\
(\bfx^i \odot \tilde{\bfa}^i_m)^T 
\end{array}
\right]^T
\end{split}
\label{eq:Grad_E}
\end{equation}
where $\bfE=[\bfe_1,\cdots,\bfe_m]$ and $\tilde{\bfA}^i=[\tilde{\bfa}_1^i,\cdots,\tilde{\bfa}_m^i]$.
The second order derivative (Hessian matrix) of $f(\bfE)$ w.r.t. $\bfE$ is a tensor and not easy to be expressed explicitly.
Thanks to the Hadamard product, the second order derivative can be computed row by row.
More specifically, for the $i$th row of $\bfE$, denoted as $\bs{\epsilon}^j (\in \mathbb{R}^{1\times m})$, we have 
\begin{equation}
\nabla^2_{\bs{\epsilon}^j} f(\bs{\epsilon}^j)= \sum\limits_{i=1}^n \left[
\begin{array}{ccc}
\tilde{a}_{1,j}^i  \tilde{a}_{1,j}^i &  \cdots & \tilde{a}_{1,j}^i  \tilde{a}_{m,j}^i\\
\tilde{a}_{2,j}^i  \tilde{a}_{1,j}^i &  \cdots & \tilde{a}_{2,j}^i  \tilde{a}_{m,j}^i\\
\vdots\\
\tilde{a}_{m,j}^i  \tilde{a}_{1,j}^i &  \cdots & \tilde{a}_{m,j}^i  \tilde{a}_{m,j}^i
\end{array}
\right]
\label{eq:End_sol}
\end{equation}
where 
$\tilde{a}_{l,j}^i$ represents the element of the matrix $\tilde{A}^i$ located in the $l$th row and
in the $j$th column and $j=1,\cdots,d$. Thus, the Hessian matrix (or Lipschitz constant) of $\nabla_{\bfE} f(\bfE)$ can be computed row by row.


Similar to the update of $\bfA$, the gradient projection method can be implemented as follows
\begin{equation}
\bs{\epsilon}^j = \Pi_{{[0,1]}^{1\times m}}\left\{\bs{\epsilon}^j-\gamma_{\bs{\epsilon}^j}\nabla_{\bs{\epsilon}^j} f(\bs{\epsilon}^j)\right\}, \quad \varepsilon \leq \gamma_{\bs{\epsilon}^j} \leq 2/{L_{\bs{\epsilon}^j}}-\varepsilon
\label{eq:UpdateE}
\end{equation}
where 
$\varepsilon \in ]0,\textrm{min}\{1,1/L_{\bs{\epsilon}^j}\}[$ and $L_{\bs{\epsilon}^j}=\|\nabla^2_{\bs{\epsilon}^j} f(\bs{\epsilon}^j)\|_F$. 
The computational complexity to calculate the endmember matrix $\bfE$ is of the order $\calO(\max\{md,m^2\}n)$.
In this work, the stepsize $\gamma_{\bs{\epsilon}^j}$ is fixed to $1/L_{\bs{\epsilon}^j}$ to
ensure a sufficient decrease of the objective value per iteration. 
The update of $\bfE$ is summarized in Algorithm \ref{Algo:UpdateE}.

%

\begin{algorithm}[h!]
\KwIn{$\bfE$, $\bfa^{1:n}$, $\bfx^{1:n}$, $P^{1:n}$}
	\tcc{Compute the temporary variable ${\tilde{\bfA}^i}$}
	${\tilde{\bfA}^i}  \leftarrow  \left((1-P^i) {\bf 1}_d  +P^i \bfx^i\right){\bfa^{i T}}$ for $i=1,\cdots,n$\;
	\tcc{Compute the gradient}
	$\nabla_{\bfE} f(\bfE)  \leftarrow$ Update $\nabla_{\bfE} f(\bfE)$ cf. \eqref{eq:Grad_E}\; 
	\tcc{Compute the Lipschitz constants row by row}
	$L_{\bs{\epsilon}^j}\leftarrow\|\nabla^2_{\bs{\epsilon}^j} f(\bs{\epsilon}^j)\|_F$ for $j=1,\cdots,d$\;
	\tcc{Compute $\hat{\bfE}$ row by row}
	$\hat{\bfE}  \leftarrow$ Update each row of $\hat{\bfE}$ cf. \eqref{eq:UpdateE}\; 

\KwOut{$\hat{\bfE}$}
\caption{Minimization w.r.t. $\bfE$}
\label{Algo:UpdateE}
\end{algorithm}


\subsection{Summary}
The proposed algorithm is summarized in Algorithm \ref{Algo:BlindNLU}. Note that the updates of $\bfa_i$ and $\bfP_i$
can be implemented for all pixels in parallel, explaining why the updates of $\bfa_i$ and $\bfP_i$ are given in matrix form
in lines 4 and 5 of Algorithm \ref{Algo:BlindNLU}, where $\bfA=\left[\bfa_1,\cdots,\bfa_n\right]$ and $\bfP=\left[P_1,\cdots,P_n\right]$.
Note that the joint estimation problem of $\bfE$, $\bfA$ and $\bfP$ is nonconvex and thus admits multiple local optima.
Thus, in practice, any other prior information is encouraged to be integrated in the estimation problem to alleviate its 
ill-posedness. For example, if we simply fix the endmember matrix \emph{a priori}, the optimization will consist of alternating between
$\bfA$ and $\bfP$, leading to a supervised nonlinear unmixing method, similar to the method investigated in \cite{Heylen2016MLM}.

\begin{algorithm}[h!]
\KwIn{$\bfX$}
\tcc{Initialize $\bfE$, $\bfP$}
$\bfE^{(0)} \leftarrow \textrm{VCA}(\bfX)$\;
$\bfP^{(0)} \leftarrow \bs 0$\;
\For{ $t = 1,2, \ldots$ \KwTo stopping rule}{
    \tcc{Update $\bfA$ cf. Algo. \ref{Algo:UpdateA} or \eqref{eq:opt_a_FCLS}}
    ${\bfA}^{(t)}  \in \{\bfA|\, L(\bfE^{(t-1)}, {\bfA},\bfP^{(t-1)}) \leq L(\bfE^{(t-1)}, {\bfA}^{(t-1)},\bfP^{(t-1)})\} $\;
    \tcc{Update $\bfP$ cf. \eqref{eq:sol_P}}
    ${\bfP}^{(t)}  \in \argmin\limits_{\bfP} L(\bfE^{(t-1)}, {\bfA}^{(t)},\bfP)$\;
    \tcc{Update  $\bfE$ cf. Algo. \ref{Algo:UpdateE}} 
    ${\bfE}^{(t)}  \in \{\bfE|\, L(\bfE, {\bfA}^{(t)},\bfP^{(t)}) \leq L(\bfE^{(t-1)}, {\bfA}^{(t)},\bfP^{(t)})\} $\; 
}
Set $\hat{\bfA}={\bfA}^{(t)}$, $\hat{\bfE}={\bfE}^{(t)}$ and $\hat{\bfP}={\bfP}^{(t)}$\;
\KwOut{$\hat{\bfA}$, $\hat{\bfE}$ and $\hat{\bfP}$}
\caption{Unsupervised Nonlinear Unmixing based on the Multilinear Mixture Model}
\label{Algo:BlindNLU}
\end{algorithm}

\subsection{Convergence Analysis}

The convergence of the proposed nonlinear unmixing algorithm can be analysed under the
framework of the BCD method. More specifically, the proposed nonlinear unmixing algorithm
contains gradient projection steps within a BCD strategy, whose convergence 
has been proved under convexity \cite{Beck2013} and nonconvexity 
assumptions \cite{Chouzenoux2013, Bolte2014} (see \cite{Wright2015} for a recent review). 
Assuming that the objective function $f$ is a continuously differentiable convex function whose gradient is 
Lipschitz, the above method, referred to as block coordinate gradient projection (BCGP) method in \cite{Beck2013}
has been proved to have sub-linear rate of convergence.
In \cite{Bolte2014}, Bolte \emph{et al.} explored the convergence of 
the iterates in a more general framework, which is 
referred to as proximal alternating linearized minimization (PALM). The authors first gave a convergence proof for 
two blocks under nonconvex and nonsmooth assumptions and then generalized it for more than two blocks (see more 
details in \cite[Theorem $1$ and Section 3.6]{Bolte2014}). When the objective function is nonconvex, the sequence of iterates
generated by PALM is guaranteed to converge to a stationary point of the objective function instead of converging to
its optimal value. In this nonlinear unmixing application, the optimization problem is obviously nonconvex
due to the entanglement of $\bfE$ and $\bfA$, which can be regarded as an extended non-negative matrix factorization.
Thus, according to the above analysis, the sequence generated by Algorithm \ref{Algo:BlindNLU}
converges to a stationary point of the objective function $L(\bfE,\bfA,\bfP)$. 





\section{Experiments using Synthetic and Real Data}
\label{sec:simu}
This section studies the performance of the proposed unsupervised nonlinear unmixing algorithm
using both synthetic and real data.
All algorithms have been implemented using MATLAB R2015b on a computer with Intel(R) Core(TM) i7-4790 CPU@3.60GHz and 16GB RAM.
The unmixing results have been evaluated using the figures of merit described in Section \ref{subsec:performance}. 
Several experiments have been conducted using synthetic datasets with controlled ground-truth. 
These experiments are studied in Section \ref{subsec:synthetic}.
Two real datasets associated with different applications are then considered in Section \ref{subsec:real}.

\subsection{Performance Measures}
\label{subsec:performance}


To analyze the quality of the estimated results, we have considered the following
normalized mean square errors (NMSEs) 
\begin{align*}
    \textrm{NMSE}_{\bfA} = \frac{\|\widehat{\bf A}- {\bf A}\|^2_F}{\| {\bf A}\|^2_F}
\end{align*}

\begin{align*}
    \textrm{NMSE}_{\bfE} = \frac{\|\widehat{\bf E}- {\bf E}\|^2_F}{\| {\bf E}\|^2_F}
\end{align*}

\begin{align*}
    \textrm{NMSE}_{\bfP} = \frac{\|\widehat{\bf P}- {\bf P}\|^2_F}{\| {\bf P}\|^2_F}.
\end{align*}

The smaller these NMSEs, the better the quality of the estimation. 
Another quality index is the spectral angle mapper (SAM), which measures the 
spectral distortion between the actual and estimated endmembers. The SAM is defined as
\begin{equation*}
\textrm{SAM}_{\bfE}(\bfe_n,\hat{\bfe}_n)=\textrm{arccos} \left(\frac{\langle\bfe_n,\hat{\bfe}_n\rangle}{ \|\bfe_n\|_2\|\hat{\bfe}_n\|_2}\right).
\label{eq:SAM}
\end{equation*}
The overall SAM is finally obtained by averaging the SAMs computed from all endmembers.
Note that the value of SAM is expressed in degrees and thus belongs to $(-90,90]$.
The smaller the absolute value of SAM, the less important the spectral distortion.

\subsection{Synthetic Data}
\label{subsec:synthetic}
In order to build the endmember matrix $\bfE$, we have randomly selected
four spectral signatures from the United States Geological Survey (USGS) digital spectral 
library\footnote{http://speclab.cr.usgs.gov/spectral.lib06/}.
In this experiment, the number of endmembers is fixed to $m=4$
and the reflectance spectra have $L = 224$ spectral bands
ranging from $383$nm to $2508$nm. The ground-truth of the endmembers
are displayed with solid lines in Fig. \ref{fig:Syn_End_ini}, \ref{fig:Syn_End_LMM} and \ref{fig:Syn_End}.
The abundance matrix $\bfA$ has been generated by drawing vectors distributed according 
to the uniform distribution in the simplex $\calA$ defined by the non-negativity and sum-to-one constraints.
The generated ground-truth of the abundance maps are shown in the first row of Figs. \ref{fig:Syn_Abu}.
The values of $P$ for all pixels have been generated by drawing samples from a uniform distribution on $[0,1]$
and is shown in the top left of Figs. \ref{fig:Syn_Prob}.
Unless indicated, the performance of these algorithms has been
evaluated on a synthetic data of size $100 \times 100 \times 224$ whose signal 
to noise ratio (SNR) has been fixed to SNR=$40$dB.



\subsubsection{Initialization}
\begin{figure}
\centering
\includegraphics[width=0.5\textwidth]{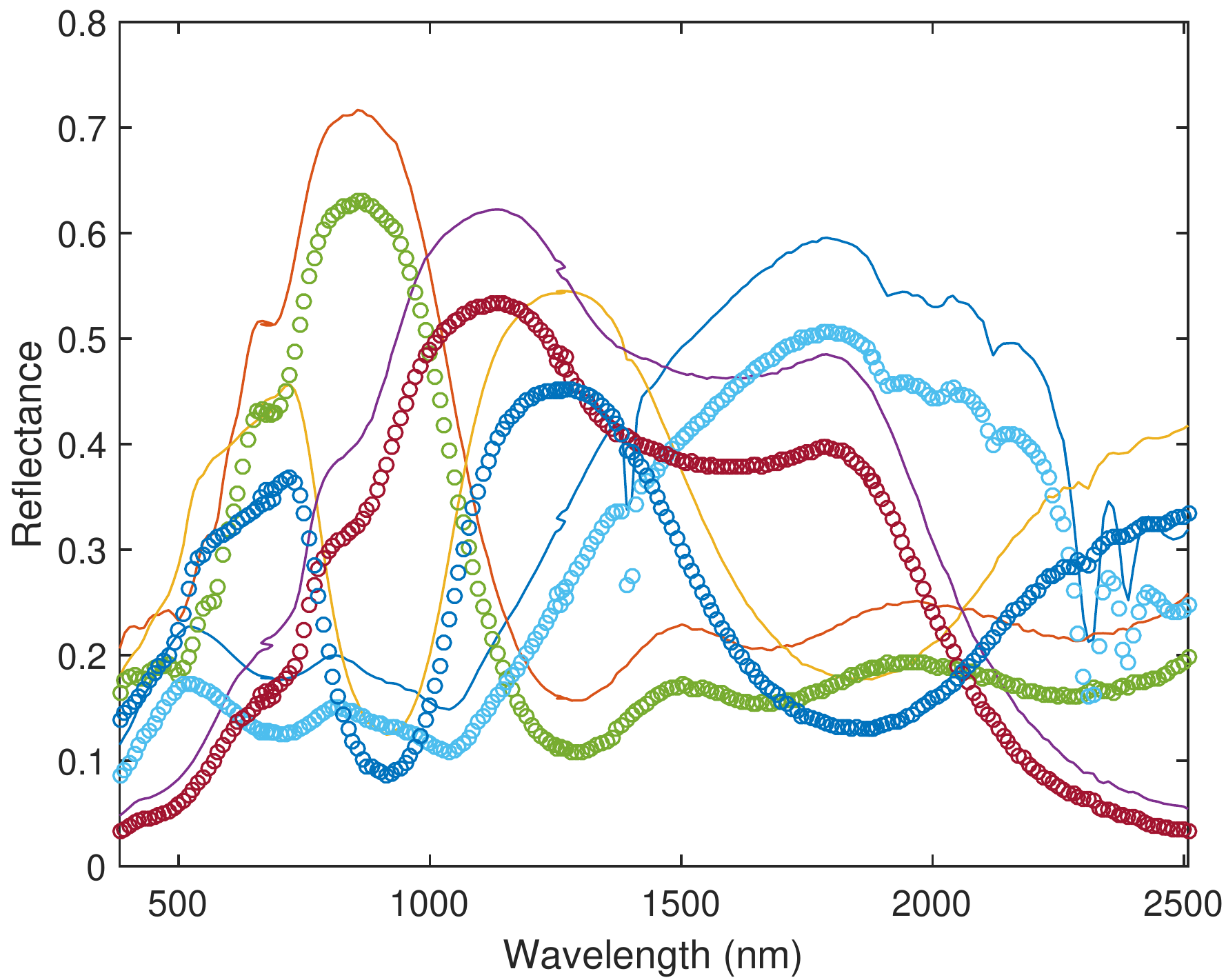}
\caption{Actual endmembers (solid lines) and their estimates with VCA (circles).}
\label{fig:Syn_End_ini}
\end{figure}

For the proposed method, the initializations of $\bfE$ and $\bfP$ are necessary.
The endmember signatures have been initialized by using the outputs of the 
VCA algorithm \cite{Nascimento2005}, {which is one of the state-of-the-art endmember extraction methods}.
The reference endmembers and their estimates provided by the VCA algorithm are displayed 
in Fig. \ref{fig:Syn_End_ini} (right). The matrix $\bfP$ is initialized with the zero matrix.

\subsubsection{Stopping rule}
As all the constraints associated with the endmembers and abundances are guaranteed to be satisfied at each update,
the main issue after several updates is to analyze the value of the objective function. 
The stopping rule used in our experiments is defined as
\begin{equation*}
\begin{split}
&L(\bfE, \bfA,\bfP) < \eta_1 \quad \textrm{ or } \\
&\frac{L(\bfE^{(t)}, \bfA^{(t)},\bfP^{(t)})-L(\bfE^{(t-1)}, \bfA^{(t-1)},\bfP^{(t-1)})}{L(\bfE^{(t-1)}, \bfA^{(t-1)},\bfP^{(t-1)})}<\eta_2
\end{split}
\end{equation*} 
where $\eta_1 =n \sigma^2$ (product of the number of pixels $n$ and the noise power $\sigma^2$) 
and $\eta_2$ has been fixed to $10^{-3}$ by cross validation.

\subsubsection{Unmixing Results}
\begin{figure}
\centering
\includegraphics[width=0.5\textwidth]{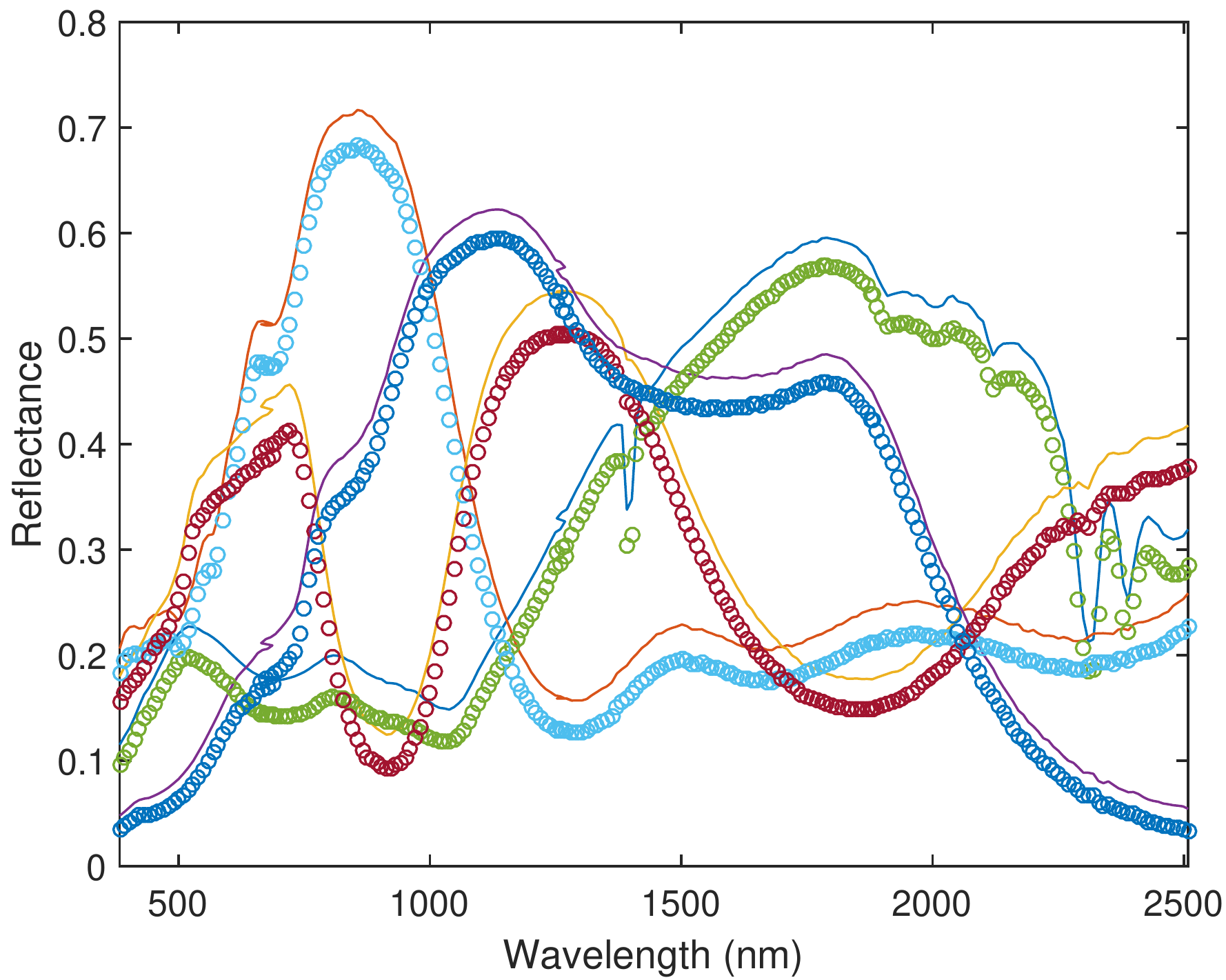}
\caption{Actual endmembers (solid lines) and their estimates using the unsupervised LU method (circles).}
\label{fig:Syn_End_LMM}
\end{figure}

\begin{figure}
\centering
\includegraphics[width=0.5\textwidth]{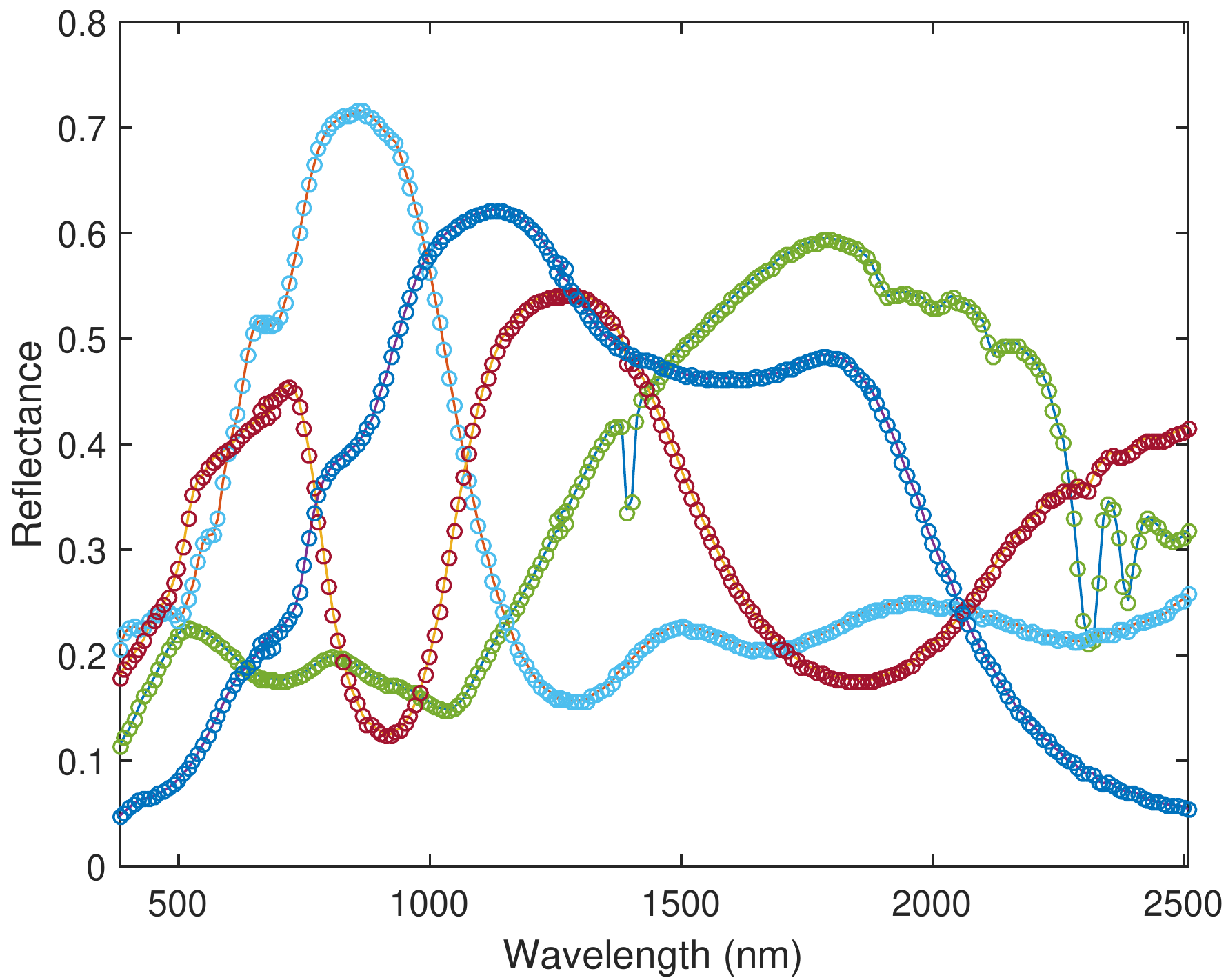}
\caption{Actual endmembers (solid lines) and their estimates using the unsupervised NLU method (circles).}
\label{fig:Syn_End}
\end{figure}

\begin{figure*}
\centering
    \subfigure{\centering
    \rotatebox{90}{\hspace{1cm} Ground-truth}}
	\subfigure{
    \includegraphics[width=0.2\textwidth]{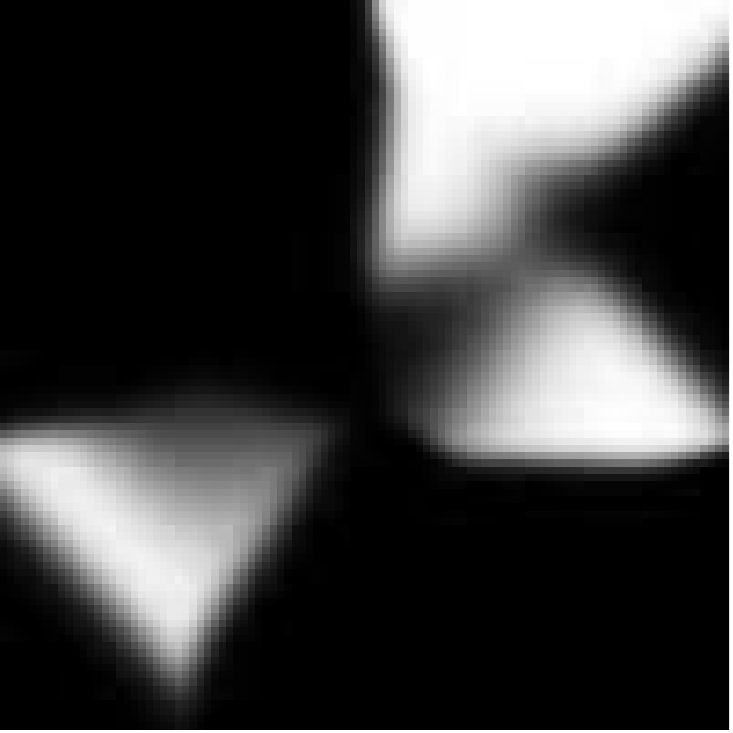}}
    \subfigure{
    \includegraphics[width=0.2\textwidth]{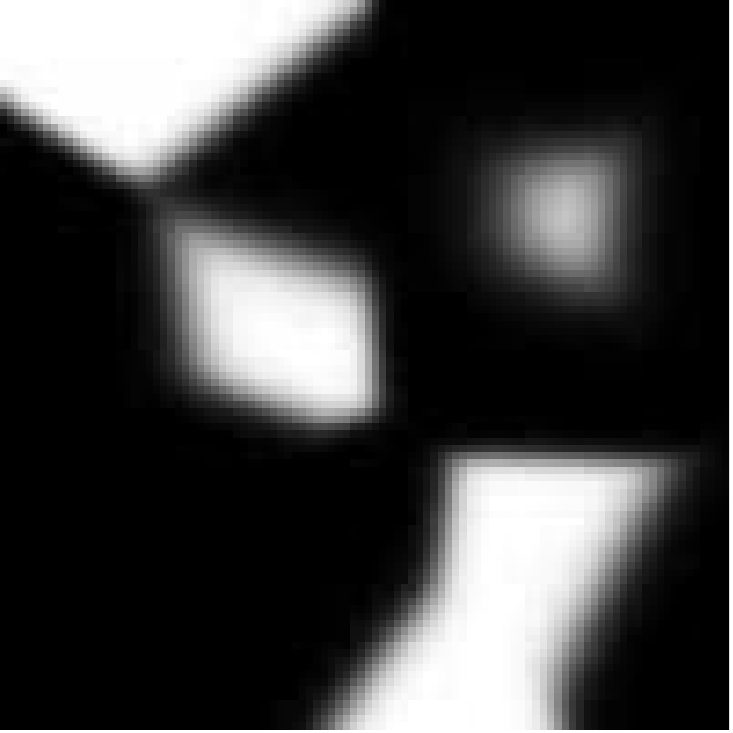}}
    \subfigure{
    \includegraphics[width=0.2\textwidth]{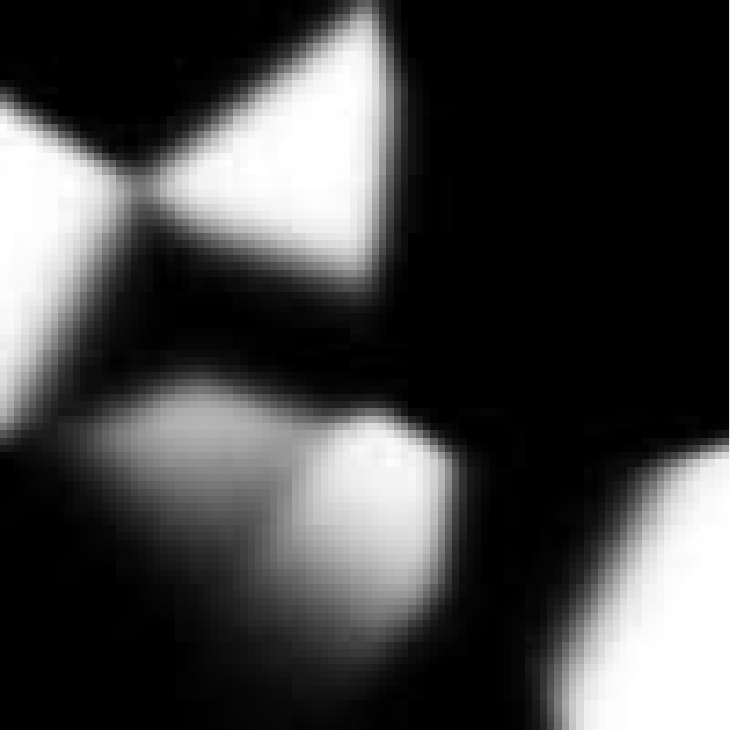}}
    \subfigure{
    \includegraphics[width=0.2\textwidth]{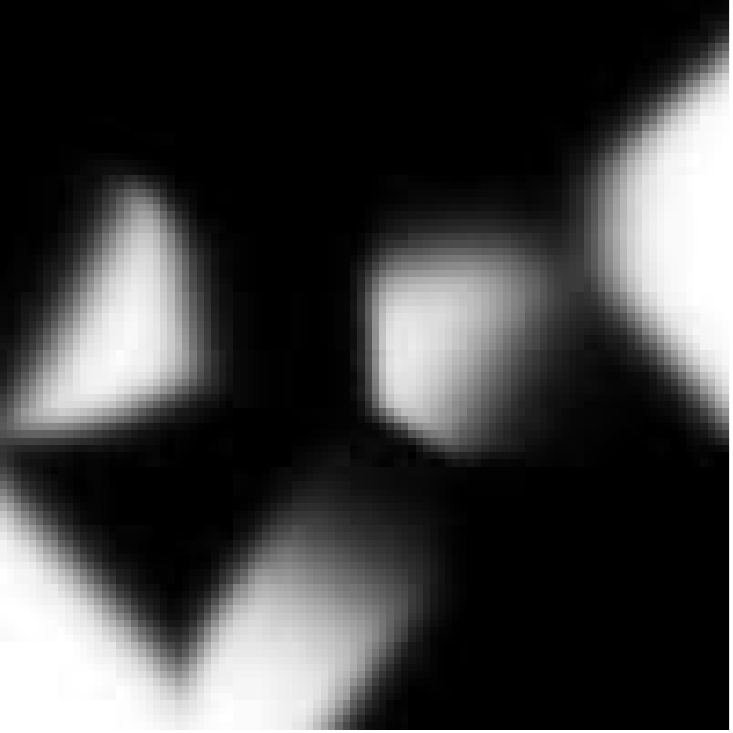}}\\
    \subfigure{\centering
    \rotatebox{90}{\hspace{0.5cm} Supervised LU}
    }
	\subfigure{
    \includegraphics[width=0.2\textwidth]{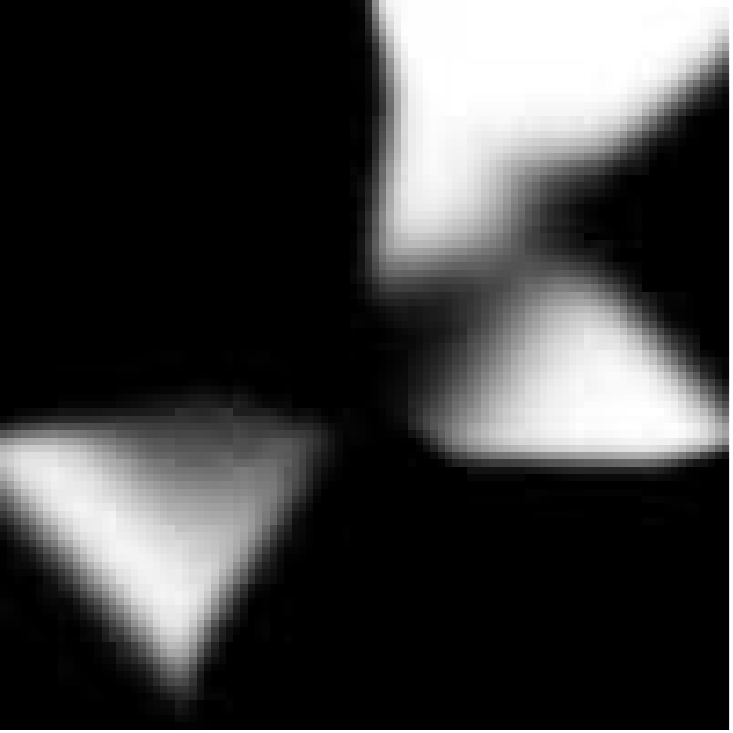}}
    \subfigure{
    \includegraphics[width=0.2\textwidth]{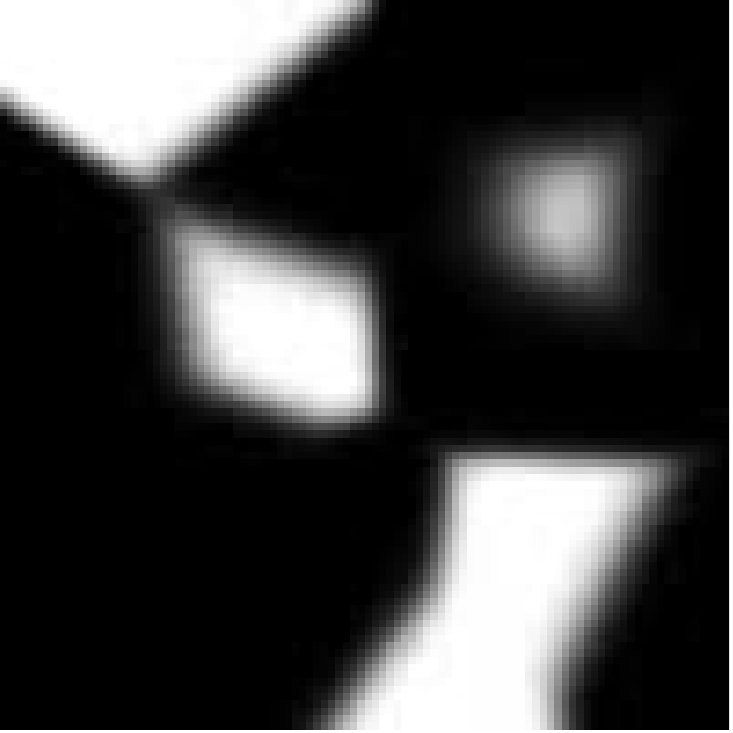}}    
    \subfigure{
    \includegraphics[width=0.2\textwidth]{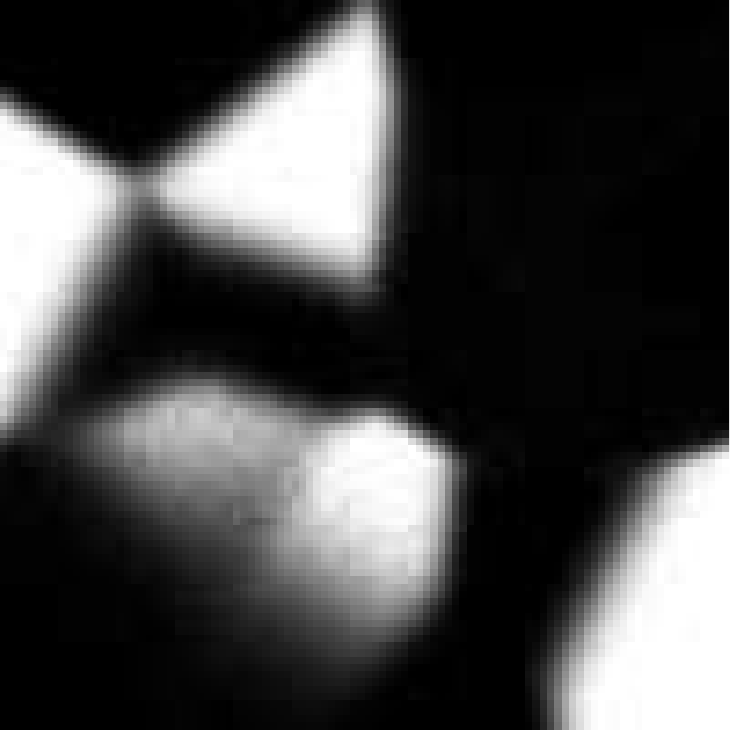}}
    \subfigure{
    \includegraphics[width=0.2\textwidth]{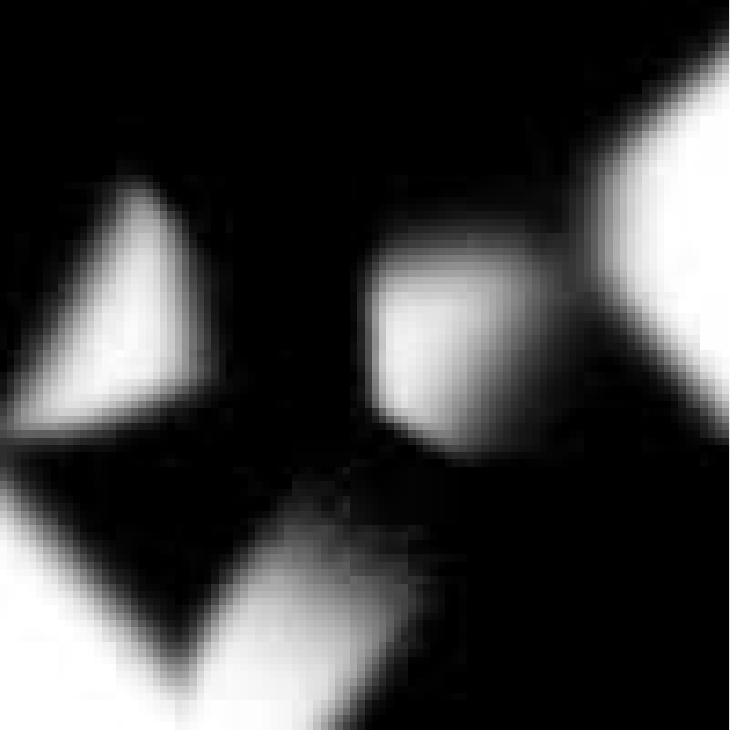}}\\
    
    \subfigure{\centering
    \rotatebox{90}{\hspace{0.3cm} Unsupervised LU}
    }
	\subfigure{
    \includegraphics[width=0.2\textwidth]{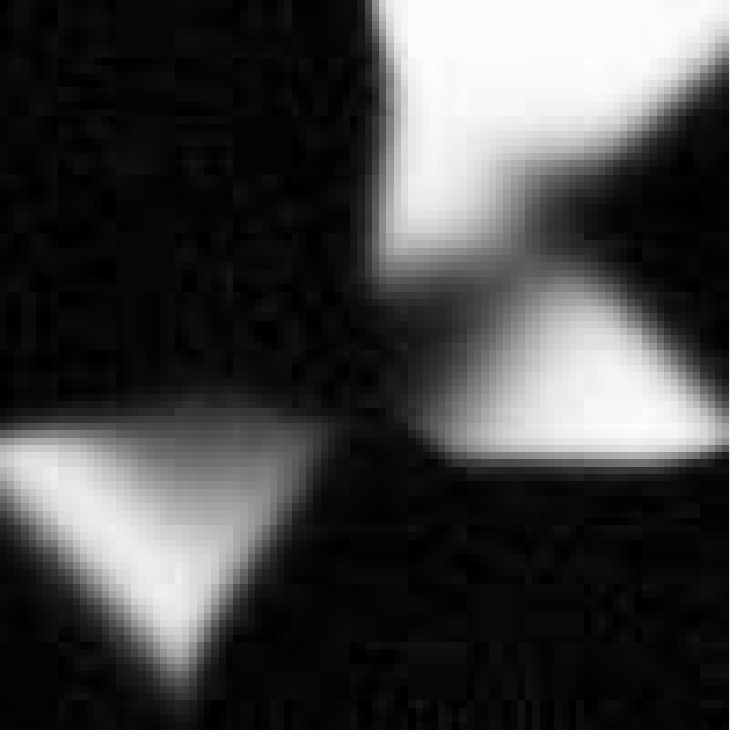}}
    \subfigure{
    \includegraphics[width=0.2\textwidth]{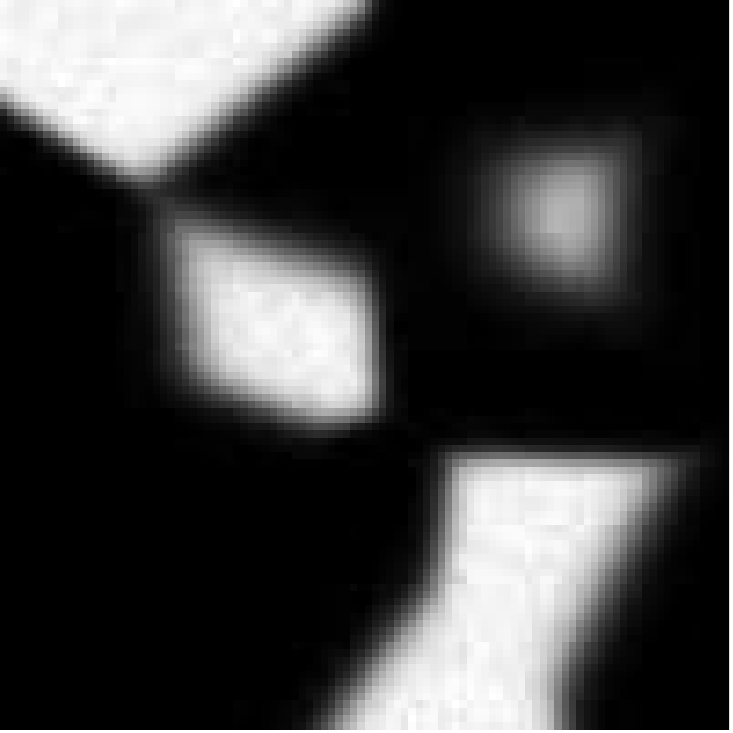}}    
    \subfigure{
    \includegraphics[width=0.2\textwidth]{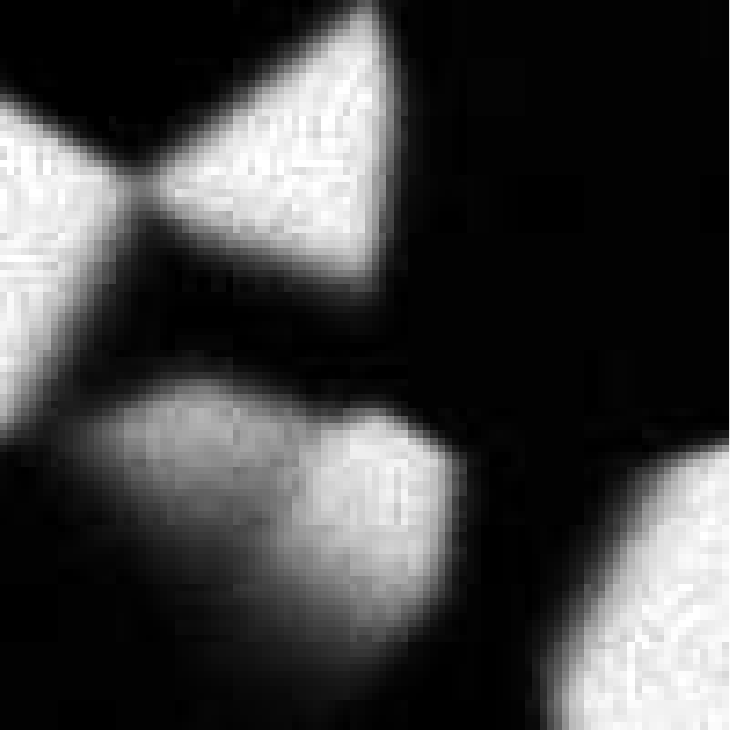}}
    \subfigure{
    \includegraphics[width=0.2\textwidth]{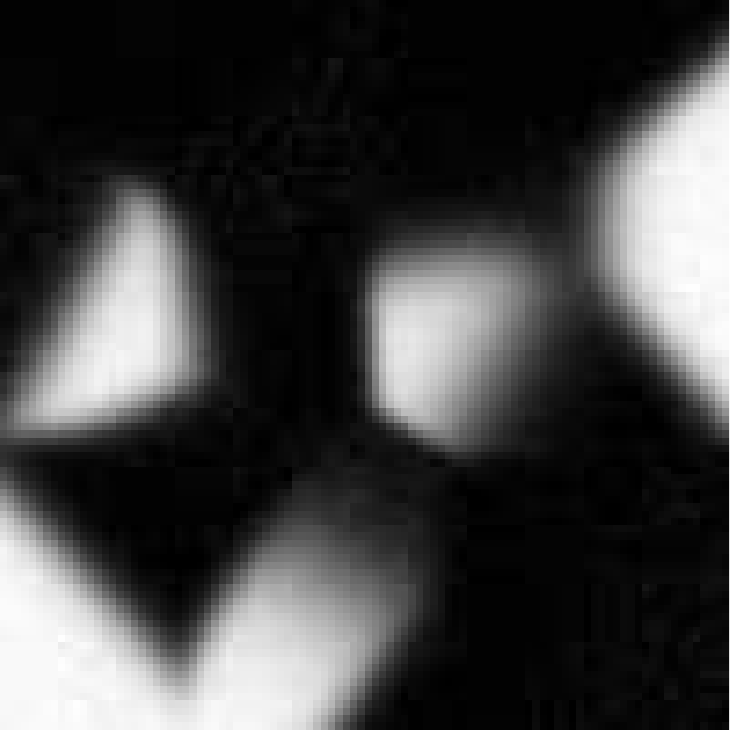}}\\
    
    \subfigure{\centering
    \rotatebox{90}{\hspace{0.3cm} Supervised NLU}
    }
    \subfigure{
    \includegraphics[width=0.2\textwidth]{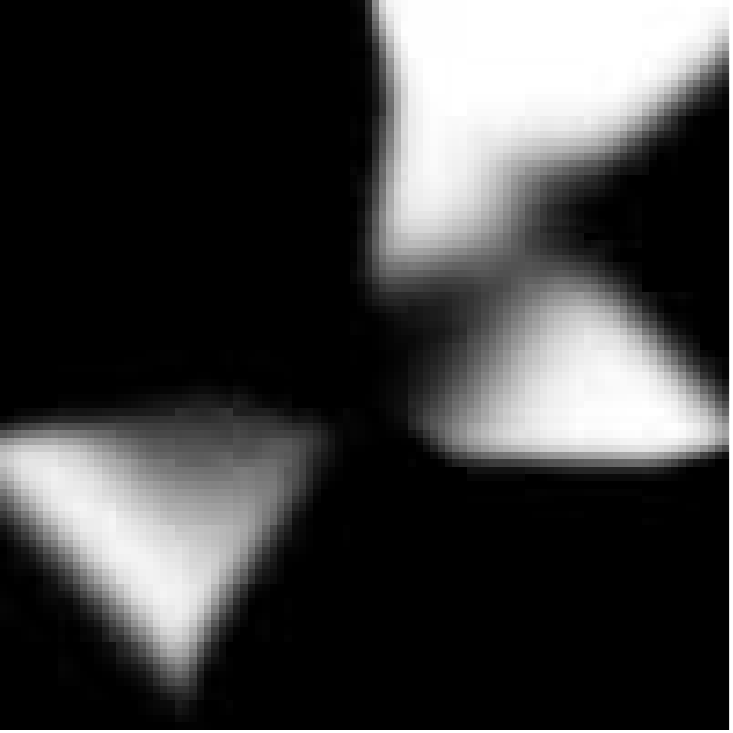}}
    \subfigure{
    \includegraphics[width=0.2\textwidth]{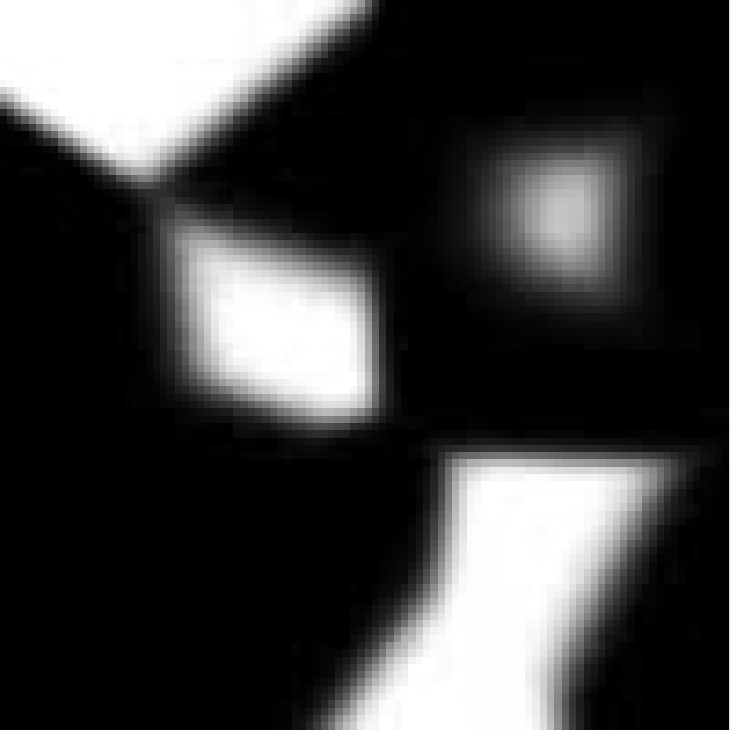}}    
    \subfigure{
    \includegraphics[width=0.2\textwidth]{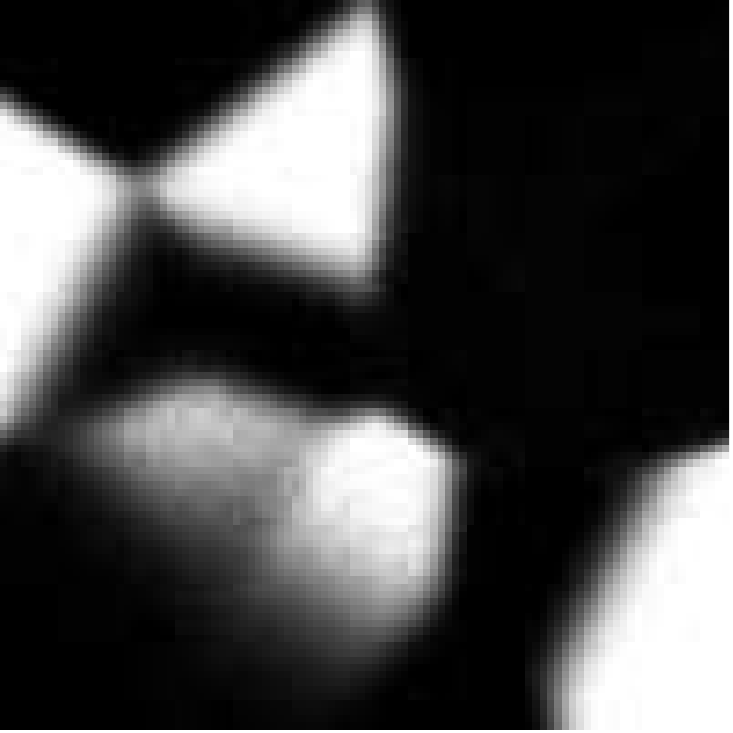}}
    \subfigure{
    \includegraphics[width=0.2\textwidth]{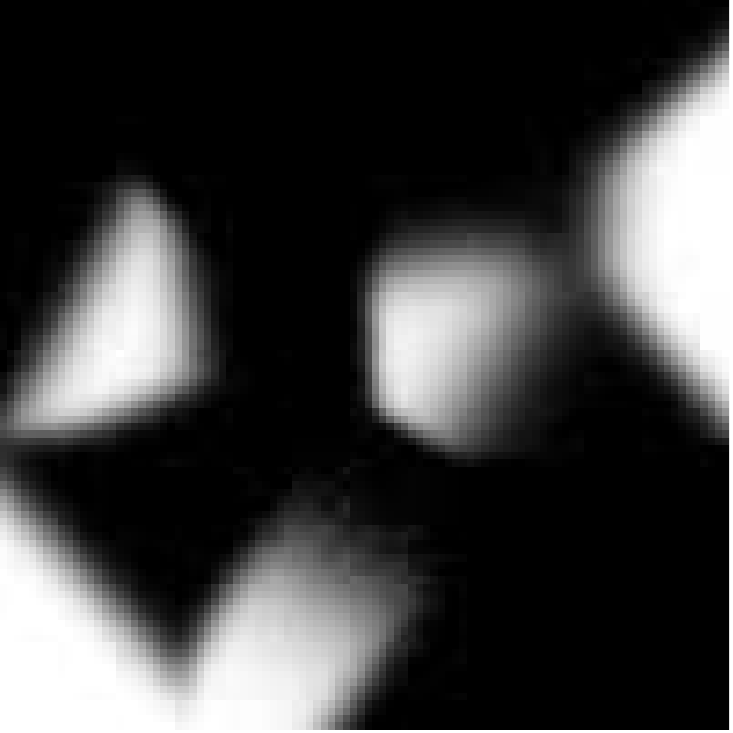}}
    
     \subfigure{\centering
    \rotatebox{90}{\hspace{0.3cm} Unsupervised NLU}
    }
    \subfigure{
    \includegraphics[width=0.2\textwidth]{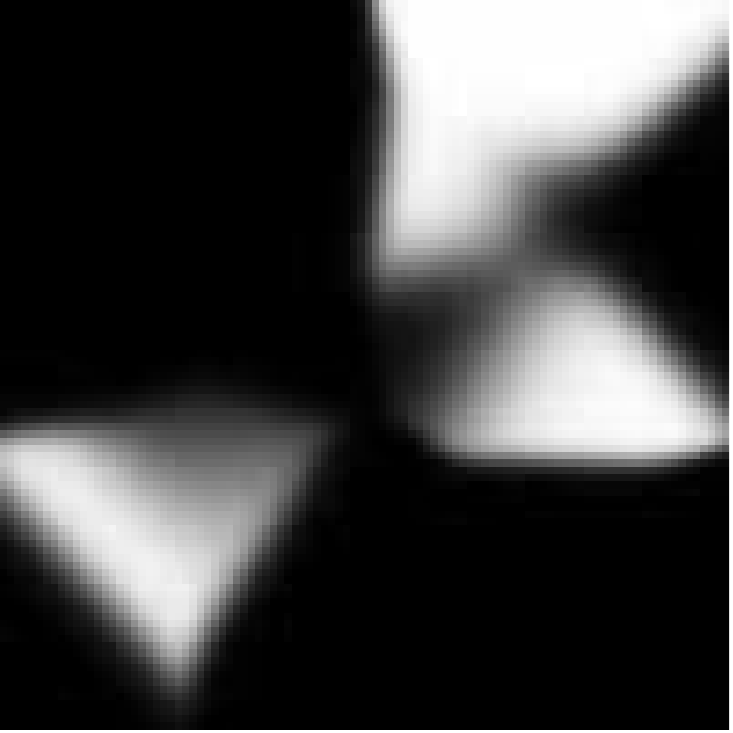}}
    \subfigure{
    \includegraphics[width=0.2\textwidth]{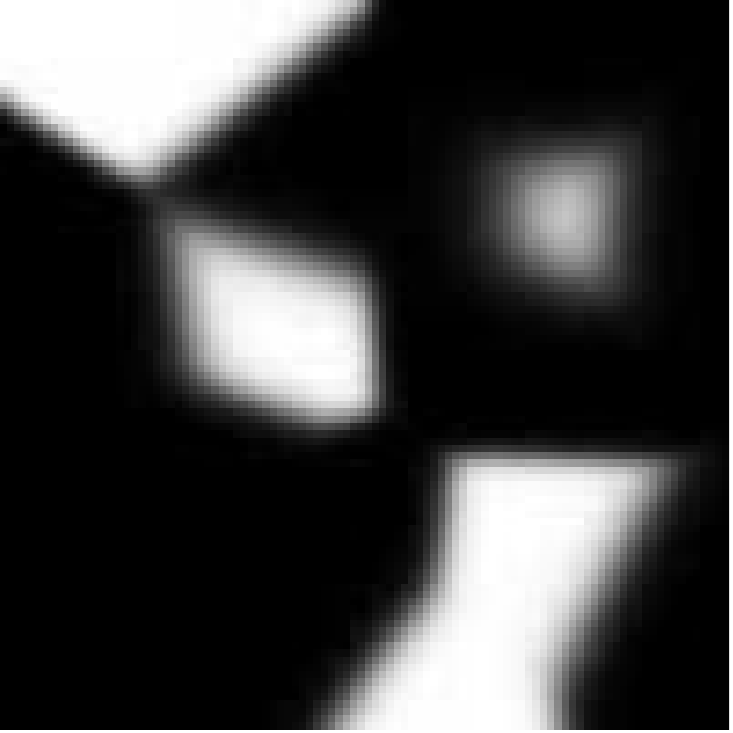}}    
    \subfigure{
    \includegraphics[width=0.2\textwidth]{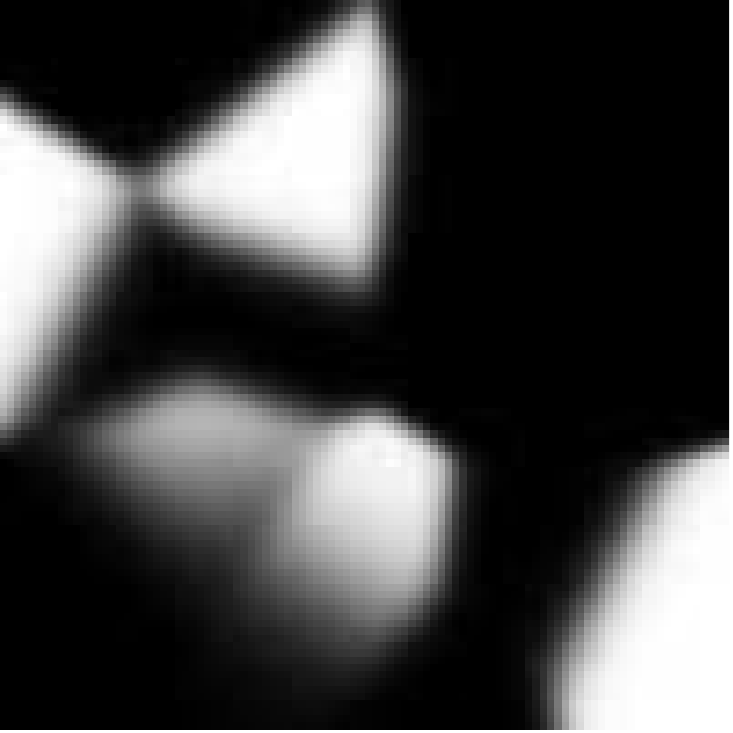}}
    \subfigure{
    \includegraphics[width=0.2\textwidth]{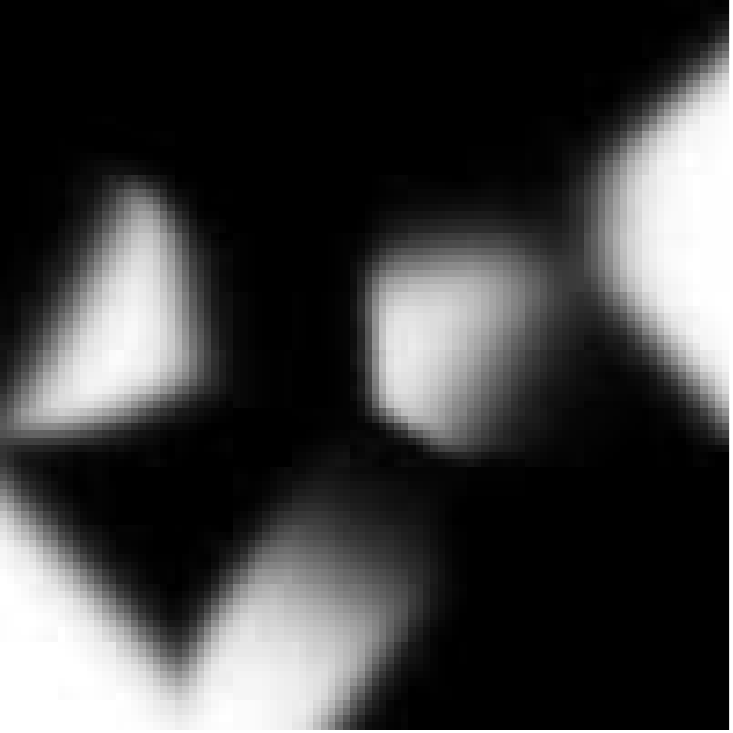}}
    \caption{Abundance maps: ground-truth (line 1), estimation by supervised LU (line 2), unsupervised LU (line 3), supervised NLU (line 4) and unsupervised NLU (line 5).}
\label{fig:Syn_Abu}
\end{figure*}

\begin{figure}
\centering
	\includegraphics[width=0.23\textwidth]{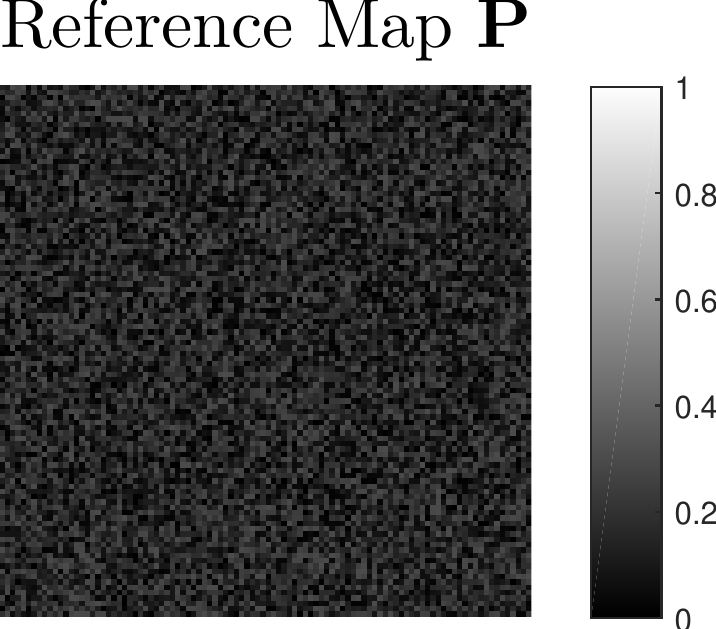}
	\includegraphics[width=0.23\textwidth]{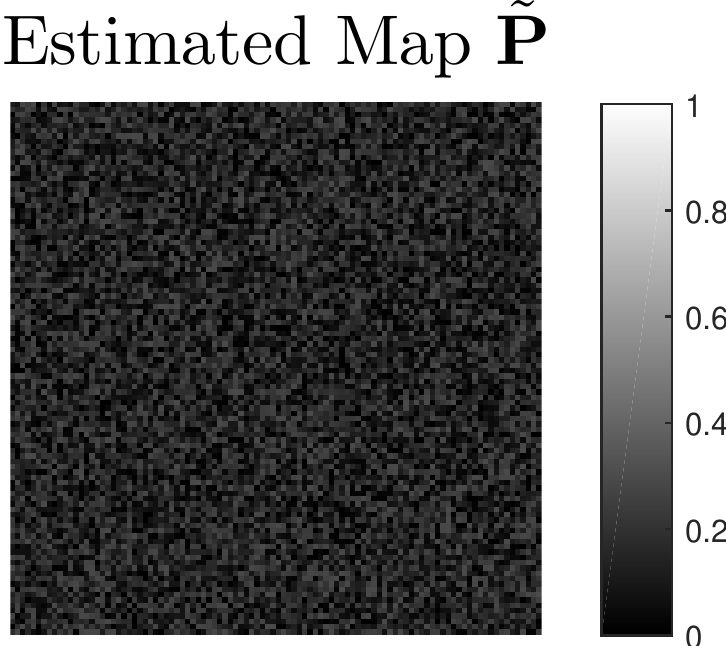}
	\includegraphics[width=0.23\textwidth]{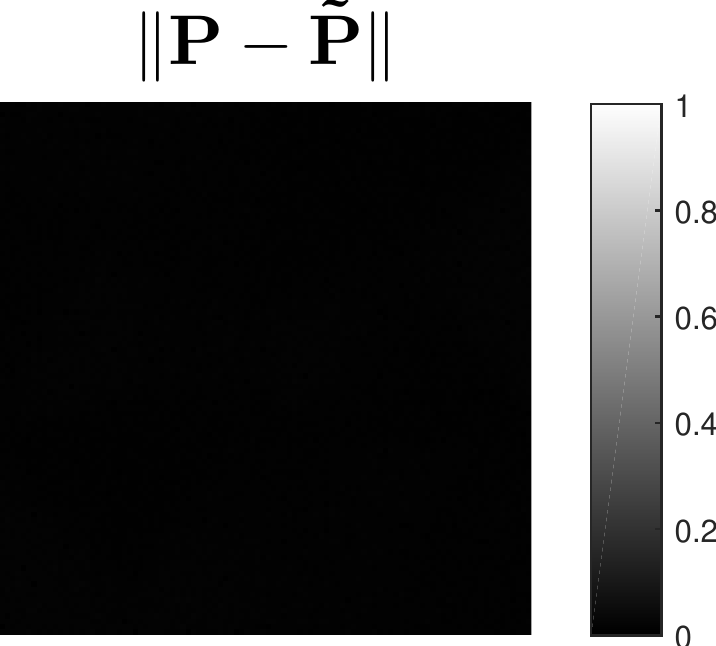}
	\includegraphics[width=0.25\textwidth]{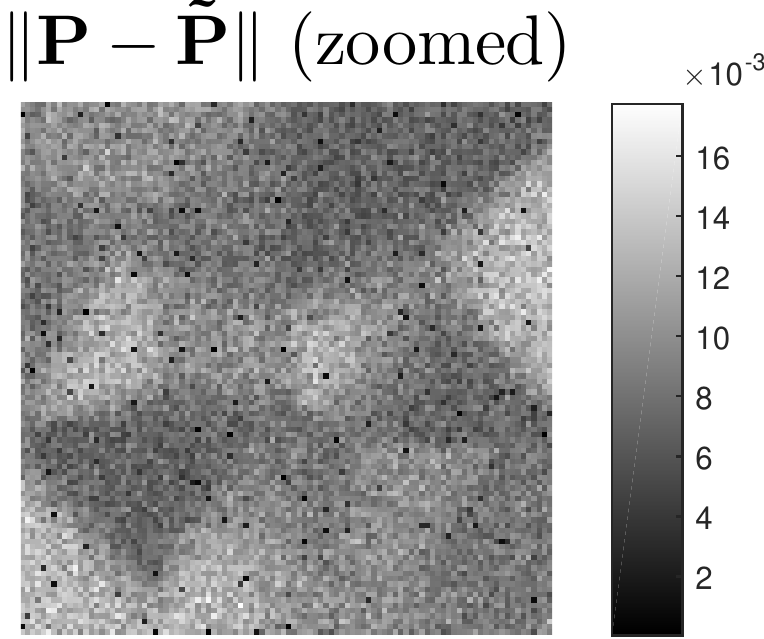}
\caption{Probability estimation: ground-truth $\bfP$ (top left), estimated $\tilde{\bfP}$ (top right) and their difference $\|\bfP-\tilde{\bfP}\|$ (bottom).}
\label{fig:Syn_Prob}
\end{figure}

The estimated endmember matrices, abundance maps and probability matrix are reported in Figs. \ref{fig:Syn_End},
\ref{fig:Syn_Abu} and \ref{fig:Syn_Prob}. To further illustrate the role of nonlinear unmixing, we have compared
our results with an LMM-based strategy. To achieve this, we simply fixed $\bfP = \bs{0}$, leading to an unsupervised
LMM spectral unmixing method. The estimated endmembers are plotted in Fig. \ref{fig:Syn_End_LMM} and the abundance maps are 
included in Fig. \ref{fig:Syn_Abu}. By fixing the endmember matrix $\bfE$ to the VCA estimates, we obtain the supervised versions
of the algorithms for linear and nonlinear unmixing. To simplify the notations, linear and nonlinear unmixing are referred to as LU and NLU thereafter.
Note that the supervised NLU can be regarded as a variant of Heylen's method in \cite{Heylen2016MLM} 
with the additional constraint that the elements of $\bfP$ belong to $[0,1]$.
Quantitative results for the estimated endmembers and abundances for both LU and NLU are reported in 
Table \ref{tb:Syn_unmixing}. 

Figs. \ref{fig:Syn_End_ini}, \ref{fig:Syn_End_LMM} and \ref{fig:Syn_End}
show that the unsupervised unmixing methods (linear and nonlinear) 
improve the estimation accuracy of the endmember signatures significantly compared to 
the output of VCA. This result demonstrates the necessity of updating
$\bfE$ jointly with the other parameters $\bfA$ and $\bfP$. Furthermore, as shown in Figs. \ref{fig:Syn_End_LMM}
and \ref{fig:Syn_End}, the estimated endmembers using NLU are much closer to the ground-truth than their
counterparts obtained using LU. The same conclusion holds for the estimation of abundances as shown in  Fig. \ref{fig:Syn_Abu}.
These results are further confirmed in the quantitative results reported in Table. \ref{tb:Syn_unmixing}.
The values of $P$ obtained for all pixels are displayed in Fig. \ref{fig:Syn_Prob} and can be used to assess the importance of nonlinear effects,
which matches the ground-truth quite well.

\newpage

\begin{table}[h!]
\renewcommand{\arraystretch}{1.1}
\setlength{\tabcolsep}{0.9mm}
\centering \caption{Comparison of Unmixing Performance for Synthetic datasets: SAM$_\bfE$ (in \textnormal{degree}), NMSE$_\bfE$ (in \textnormal{dB}), NMSE$_\bfA$ (in \textnormal{dB}) and NMSE$_\bfP$ (in \textnormal{dB}).}
\vspace{0.1cm}
\begin{tabular}{|c|cccc|}
\hline
 Methods & SAM$_\bfE$ & NMSE$_\bfE$ & NMSE$_\bfA$ & NMSE$_\bfP$ \\
\hline
\hline
LU (fixing $\bfE$) & 2.42  & -14.72 & -24.37   &/ \\  
\hline
LU (relaxing $\bfE$)&  2.08 &  {-21.09} & {-24.38} & /\\
\hline
NLU (fixing $\bfE$) &  2.42 &  {-14.72} & {-26.52} & -0.035\\
\hline
NLU (relaxing $\bfE$) &  0.13 &  {-44.05} & {-48.77} & -25.26\\
\hline
\end{tabular}
\label{tb:Syn_unmixing}
\end{table}

\subsection{Real Dataset}
\label{subsec:real}
\subsubsection{Tahoe Dataset}
\begin{figure*}
\centering
\hspace{-0.5cm}
\subfigure{
\includegraphics[width=0.25\textwidth]{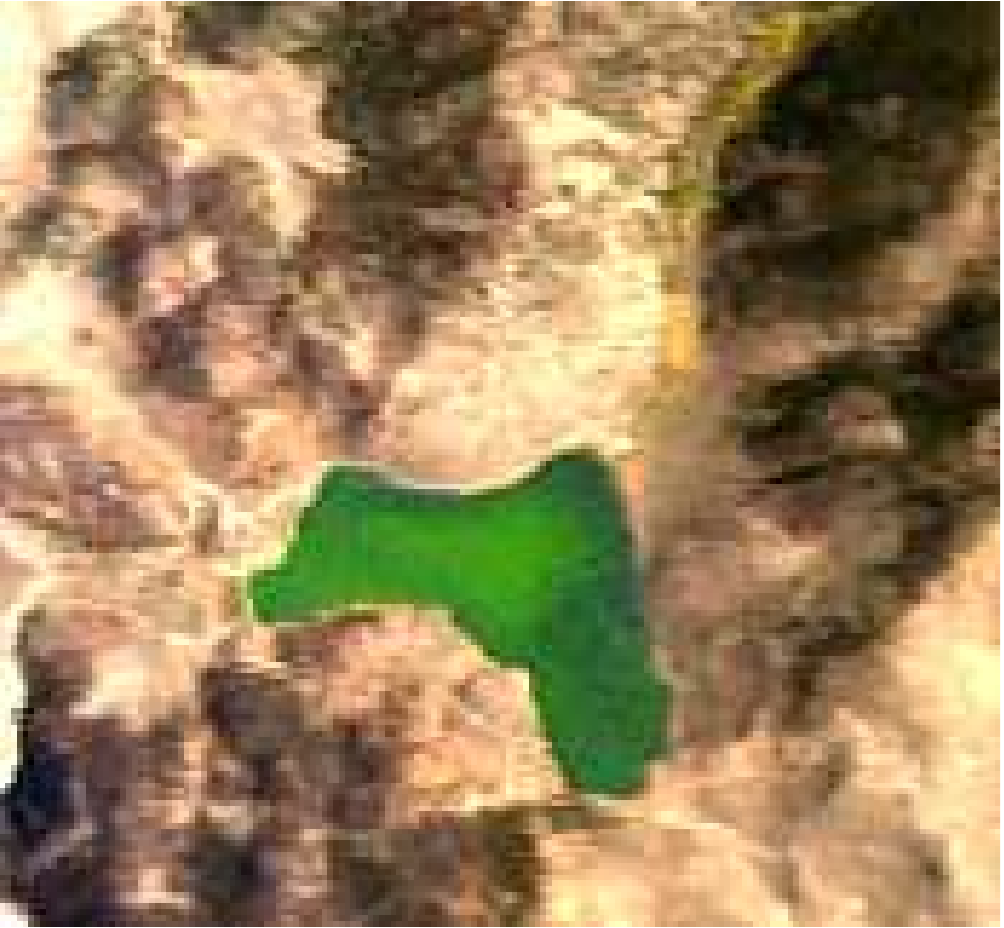}}
\subfigure{
\includegraphics[width=0.3\textwidth]{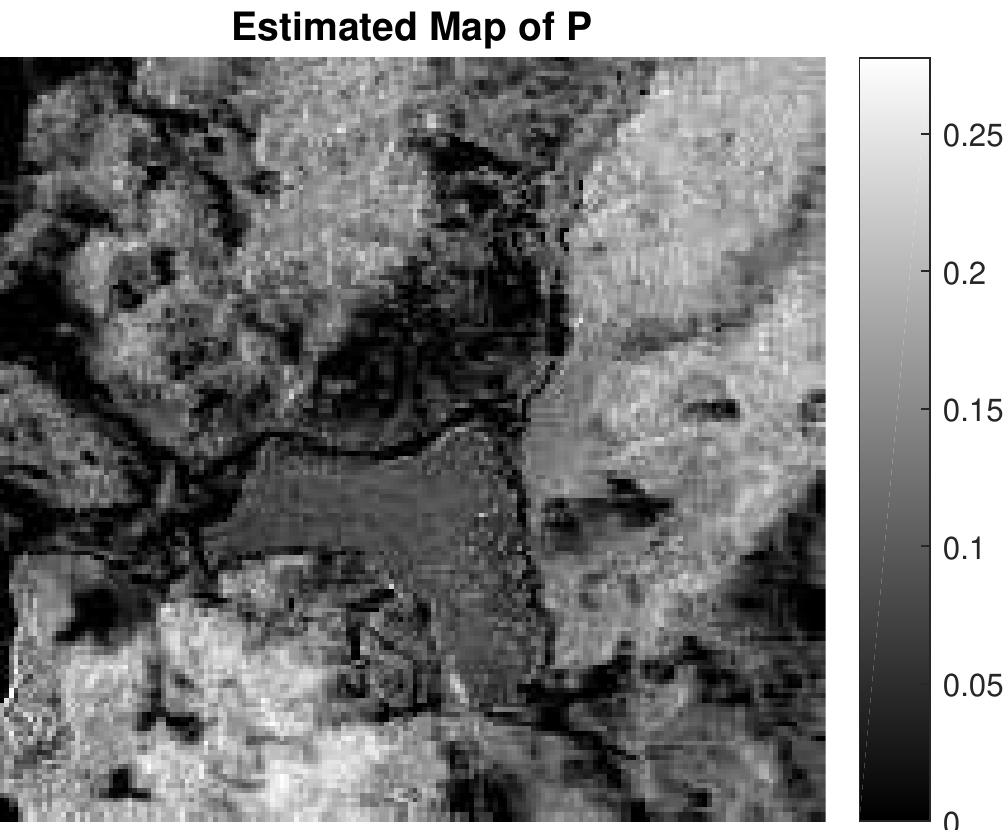}}
\subfigure{
\includegraphics[width=0.3\textwidth]{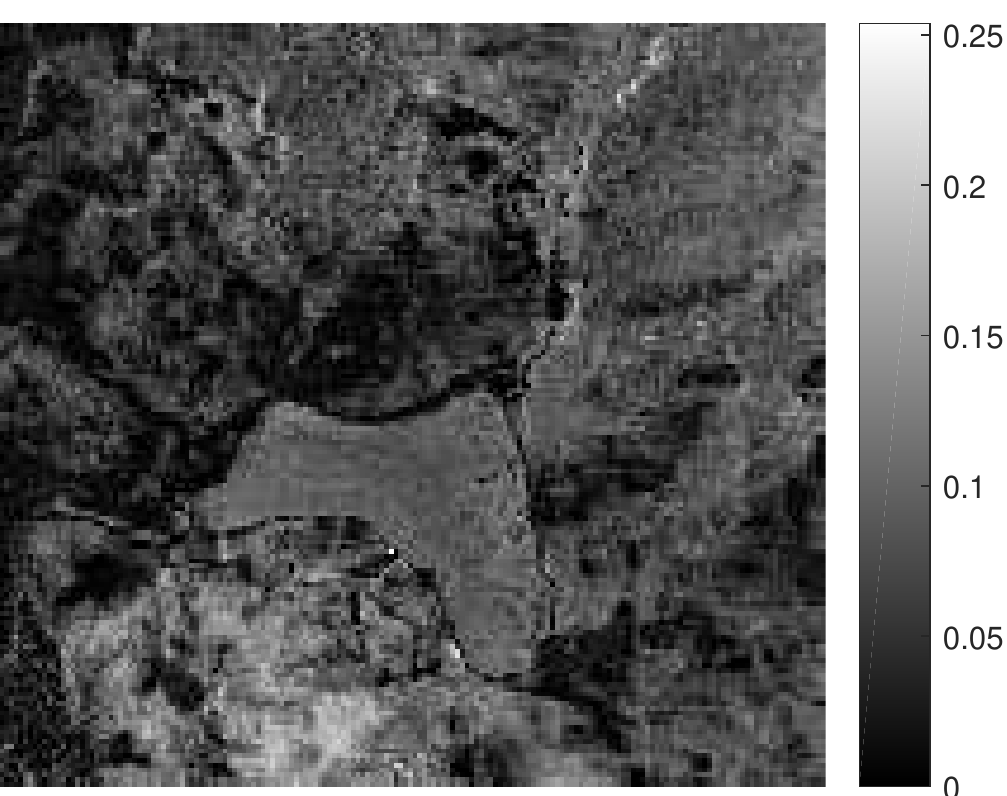}}
\caption{Left: Tahoe dataset. Middle: Estimated map $\hat{\bfP}$. Right: Sum of absolute differences between abundance maps estimated by LU and NLU.}
\label{fig:HS}
\end{figure*}

\begin{figure}
\centering
\includegraphics[width=0.5\textwidth]{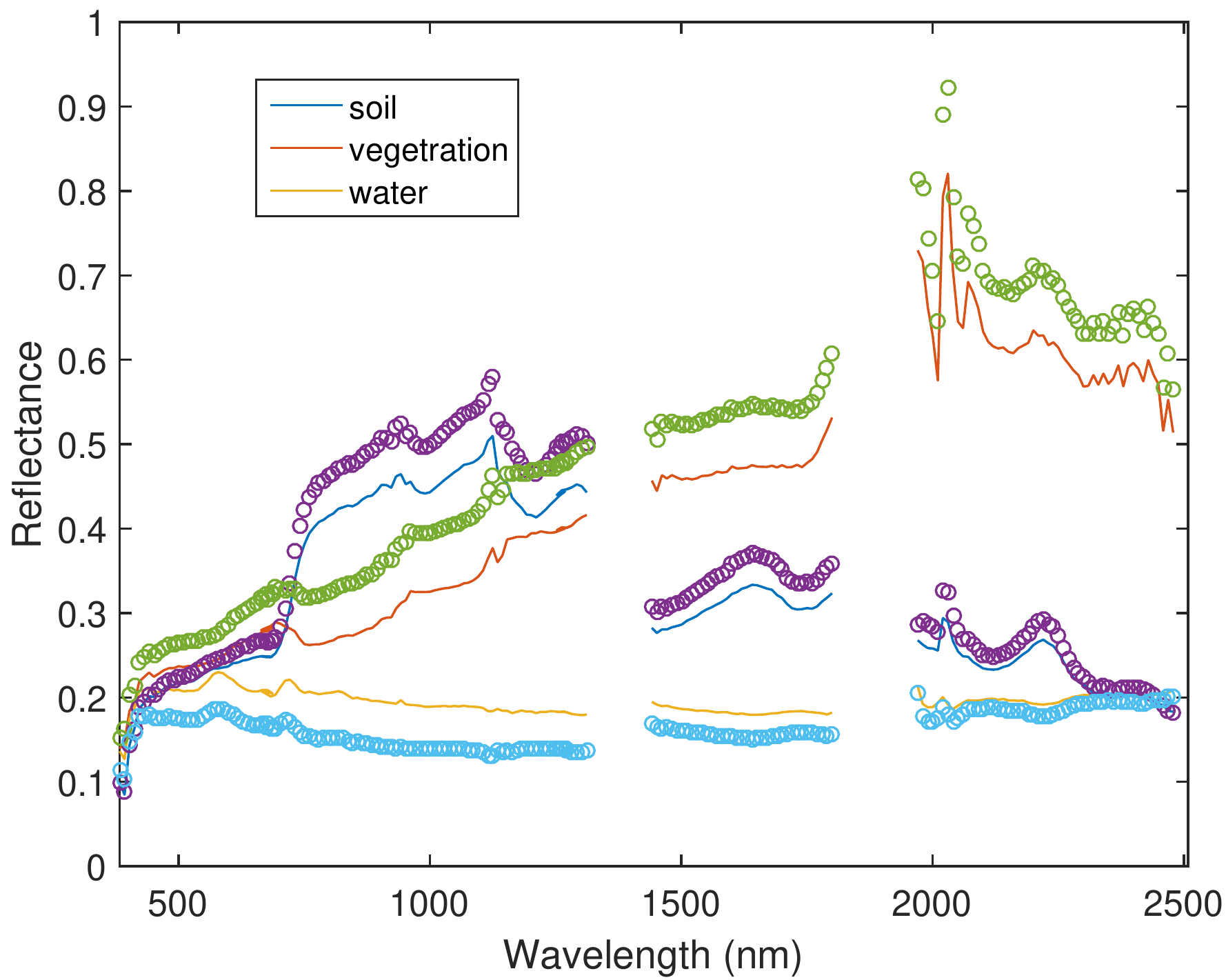}
\includegraphics[width=0.5\textwidth]{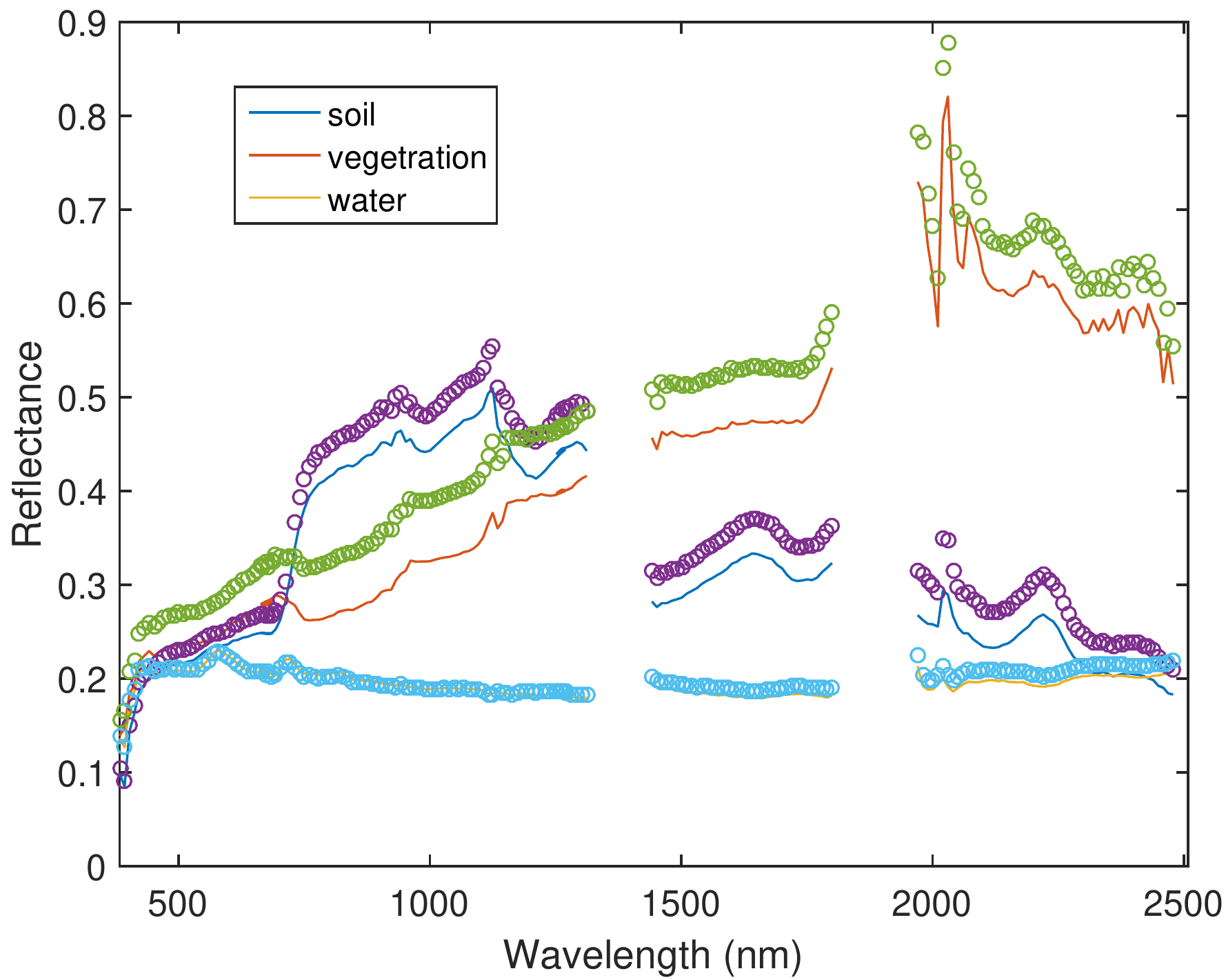}
\caption{Tahoe dataset: Estimated endmember signatures of soil, vegetation and water using (Top) VCA (solid) and LU (circle), (Bottom) VCA (solid) and NLU (circle).}
\label{fig:Real_End}
\end{figure}

In this experiment, we consider an HS image of size $161 \times 174 \times 224$
acquired over Tahoe, located along the border between California and Nevada, on October 13th, 2015 
by the JPL/NASA airborne visible/infrared imaging spectrometer (AVIRIS)\footnote{http://aviris.jpl.nasa.gov/}.
This image was initially composed of $224$ bands that have been
reduced to $192$ bands ($d=192$) after removing the water vapor absorption bands
as well as the highly noisy bands. The spatial resolution of this HS image is around $20$m per pixel
and it mainly contains water, soil and vegetation.
A composite color image of the scene of interest is shown in the left of Fig. \ref{fig:HS}. In this experiment, 
the number of endmembers is fixed to be $m=3$ according to our available prior knowledge. The three endmembers 
were estimated using VCA and are shown with solid lines in Fig. \ref{fig:Real_End}.
Note that $\bfP=\bs{0}$ was used for the initialization of $\bfP$ as in Section \ref{subsec:synthetic}.

\begin{figure*}
\centering
	\subfigure{\centering
    \rotatebox{90}{\hspace{0.5cm} Unsupervised LU}}
    \subfigure{
    \includegraphics[width=0.3\textwidth]{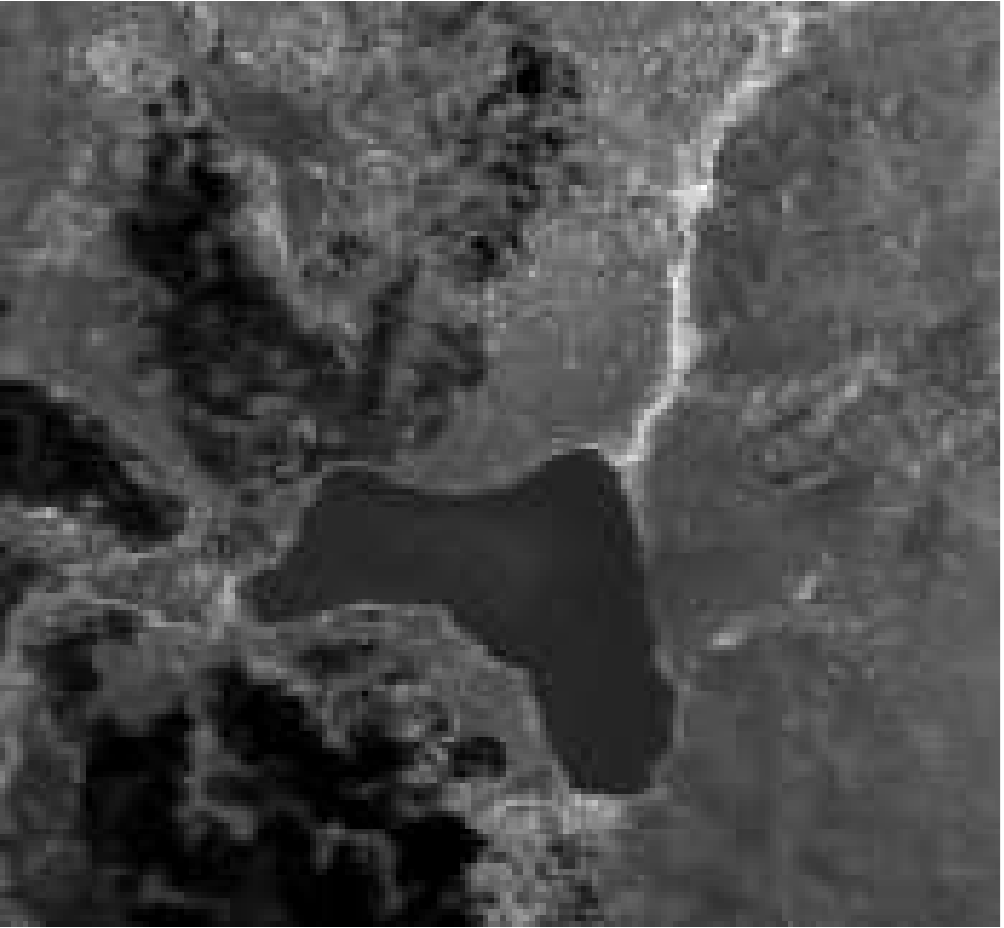}}
    \subfigure{
    \includegraphics[width=0.3\textwidth]{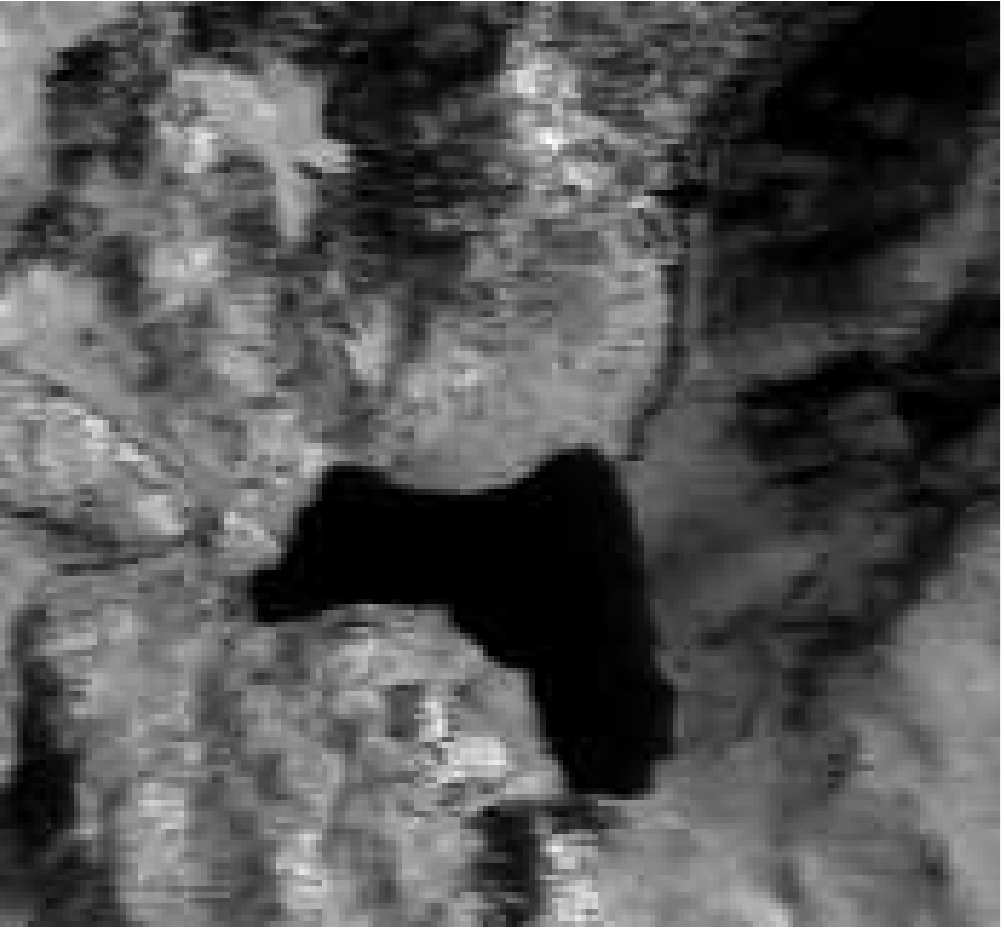}}
    \subfigure{
    \includegraphics[width=0.3\textwidth]{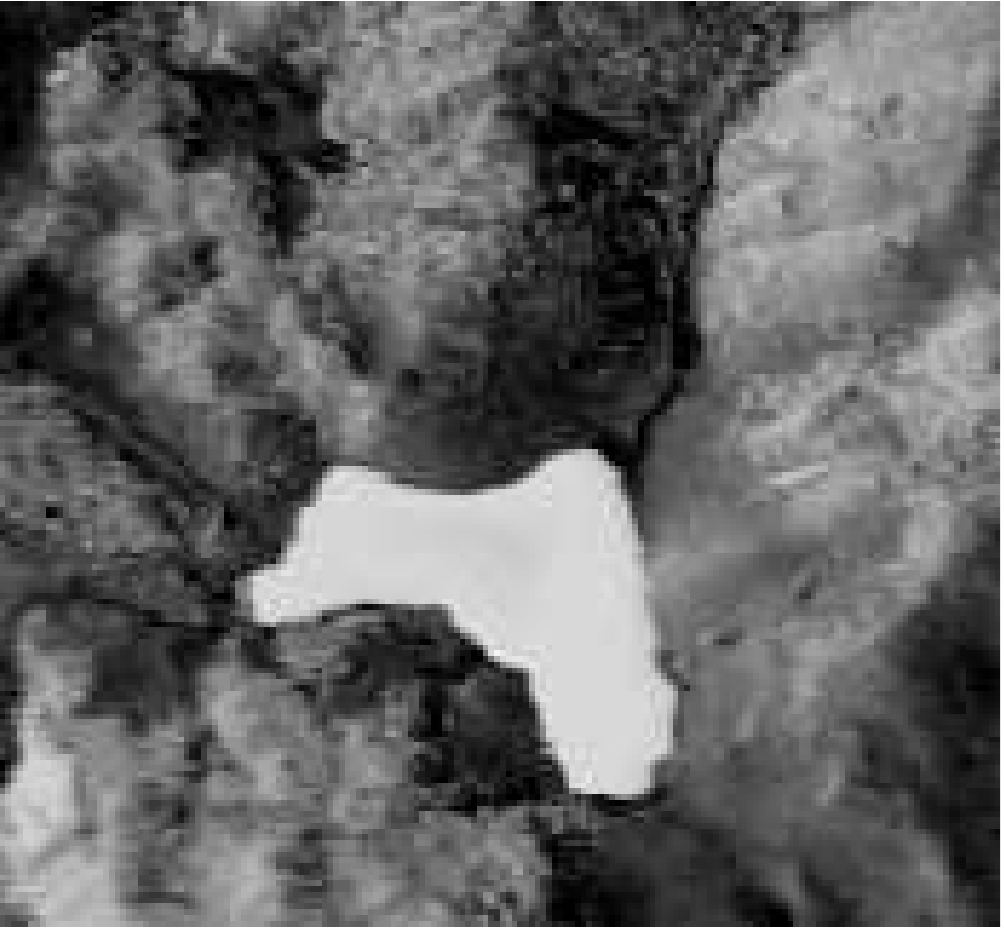}}\\
   	\subfigure{\centering
    \rotatebox{90}{\hspace{0.5cm} Unsupervised NLU}}
    \subfigure{
    \includegraphics[width=0.3\textwidth]{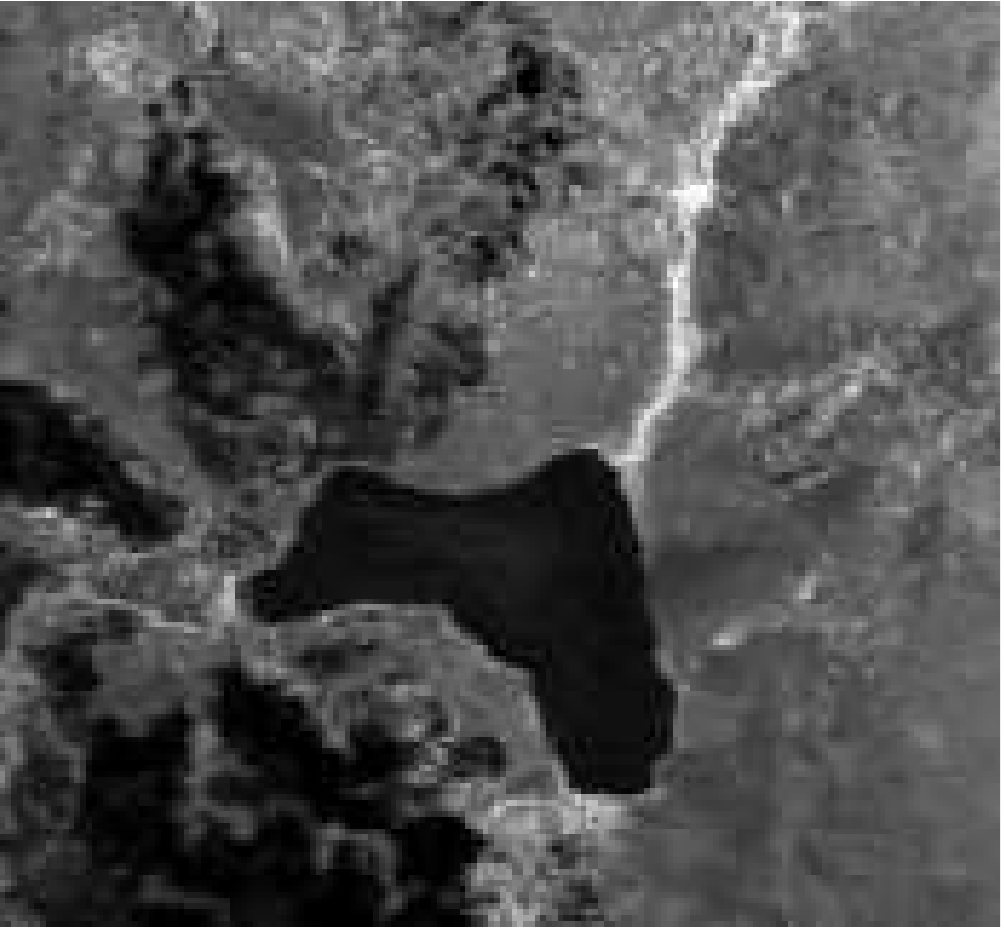}}
    \subfigure{
    \includegraphics[width=0.3\textwidth]{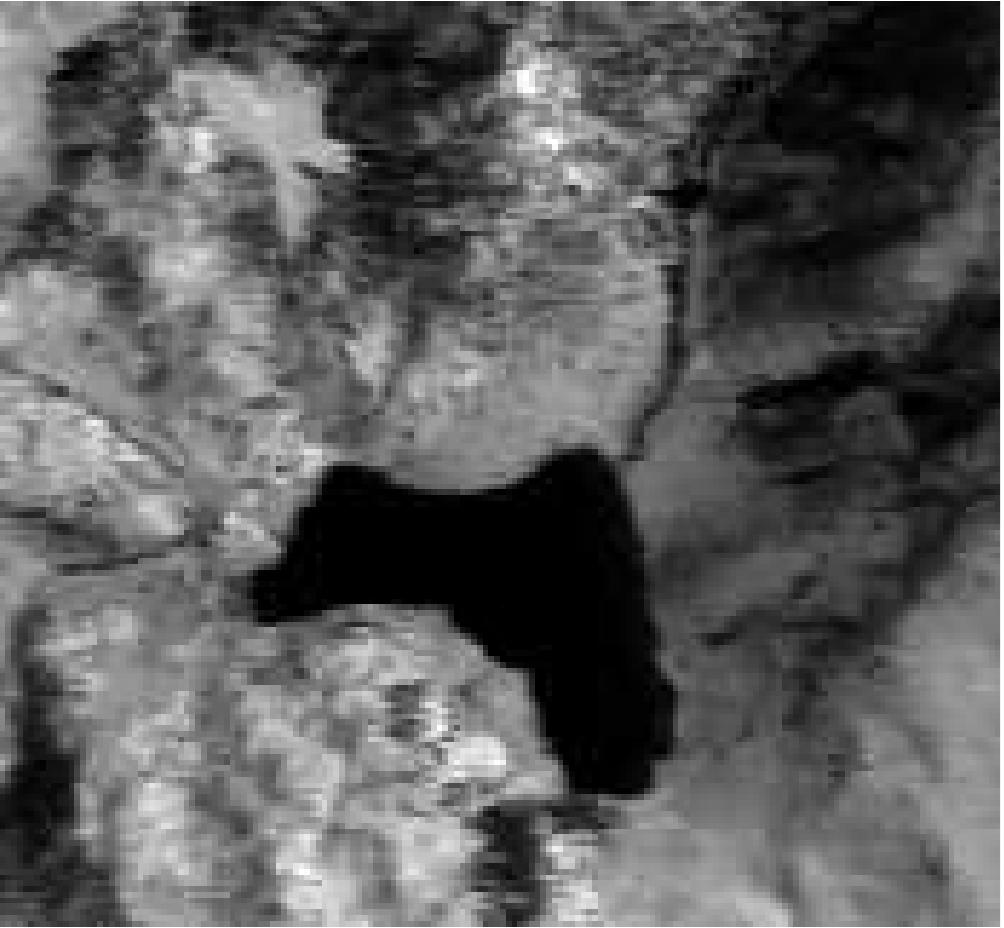}}
    \subfigure{
    \includegraphics[width=0.3\textwidth]{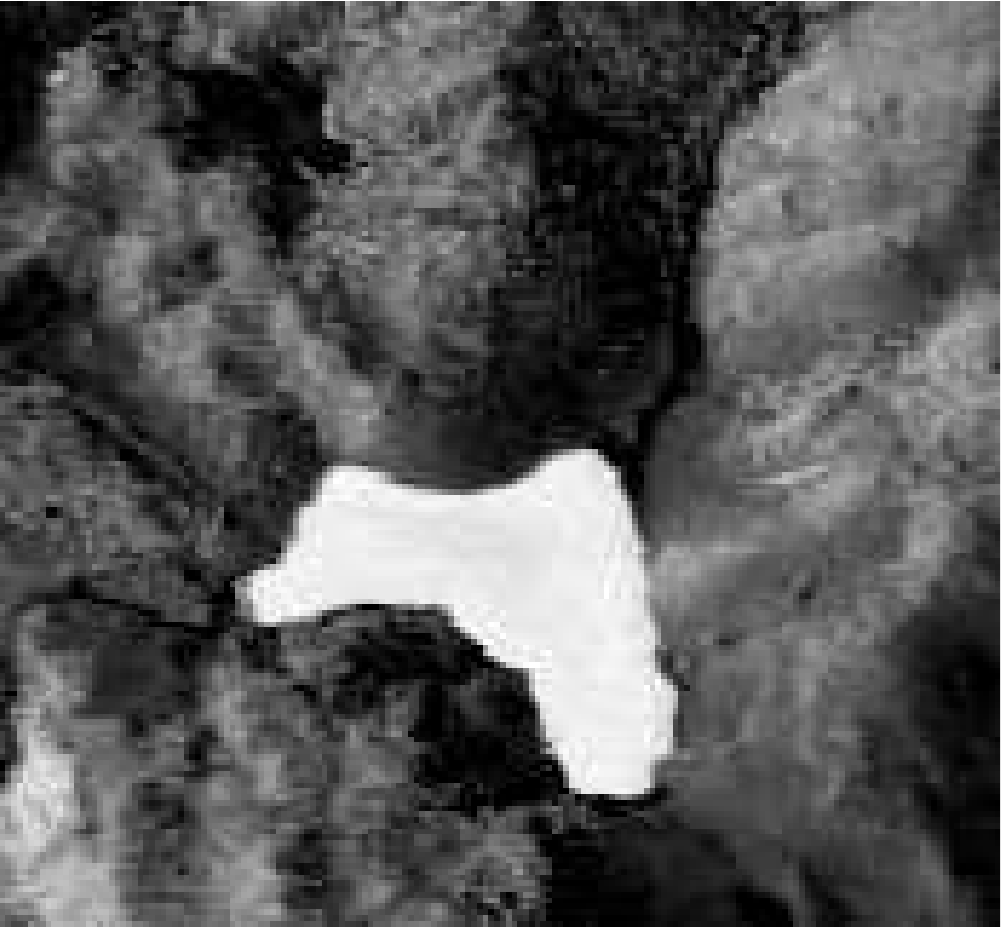}}\\
   	\subfigure{\centering
    \rotatebox{90}{\hspace{1cm} Difference}}
   	\subfigure{
    \includegraphics[width=0.3\textwidth]{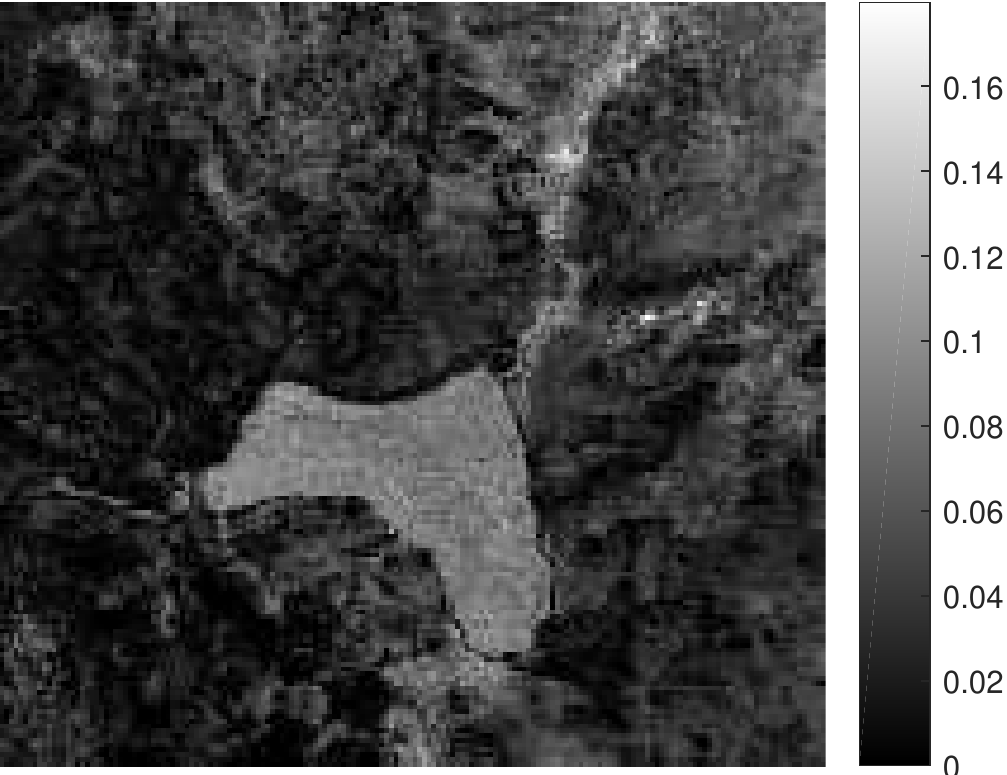}}
    \subfigure{
    \includegraphics[width=0.3\textwidth]{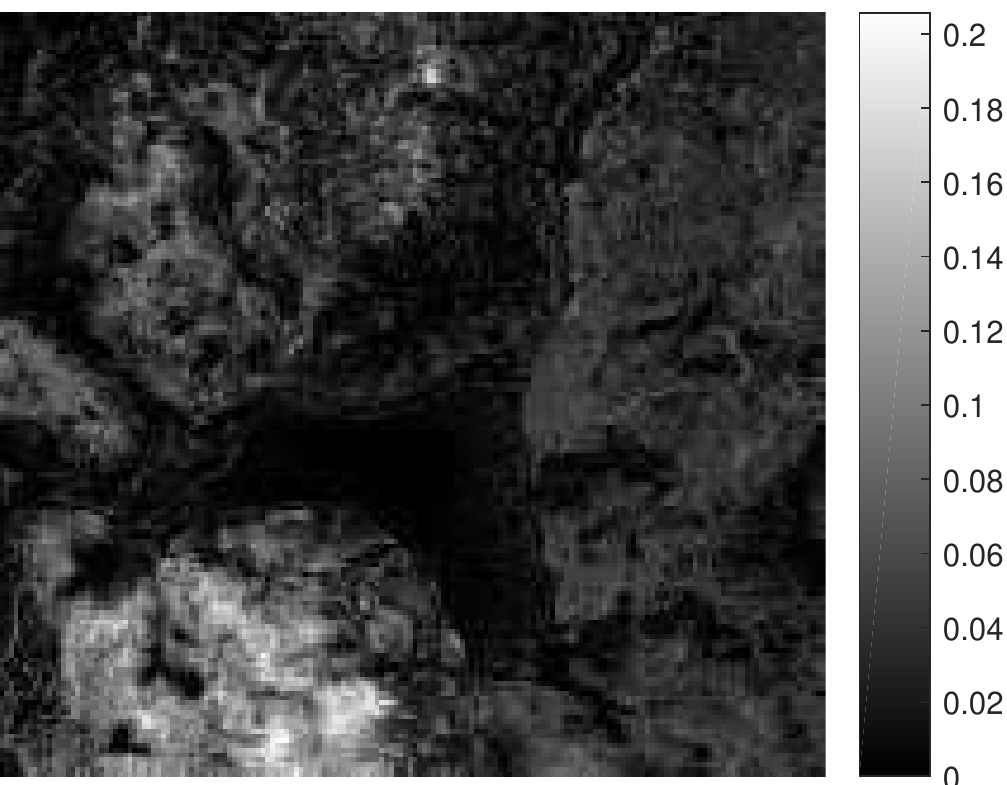}}
    \subfigure{
    \includegraphics[width=0.3\textwidth]{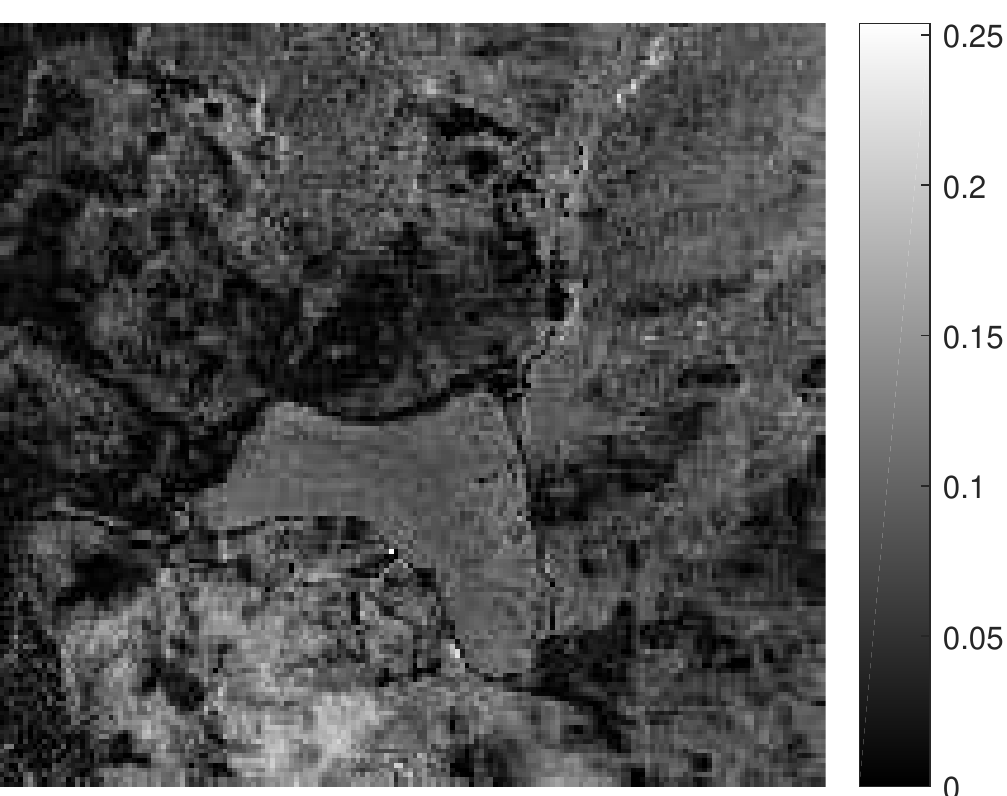}}
    \caption{Estimated abundance maps with LU and NLU methods (from left to right): soil, vegetation and water.}
\label{fig:Real_Abu}
\end{figure*}

The proposed NLU and LU (obtained by fixing $\bfP=\bs{0}$) have been implemented to process the observed image.
As there is no ground-truth for this image, unmixing results are first studied qualitatively by 
displaying the endmembers and abundances. Figs. \ref{fig:Real_End} show that the
the estimated signatures of soil and vegetation using 
LU and NLU are similar while those of water are slightly different. The corresponding 
abundance maps obtained by LU and NLU shown in the first two rows of Figs. \ref{fig:Real_Abu} 
(the color scales are exactly the same) are globally similar. However, some differences can be observed in
the last row of Fig. \ref{fig:Real_Abu}. 
It is interesting to note that the abundance maps of NLU have larger contrast than the ones obtained with 
LU, which is usually expected for good unmixing results. For example, the lake part is expected to have small
values in the abundance map of soil and to have big values in abundance map of water.
As is shown in the first and second rows of Figs. \ref{fig:Real_Abu}, the lake part in NLU is darker than the
one in LU for the soil map and is brighter than the one in LU for the water map, showing that the results obtained by 
NLU consist more reasonably with our experience. 
In order to appreciate the interest of using a nonlinear unmixing strategy,
Fig. \ref{fig:HS} shows the estimated map of $\bfP$ (in the middle) which reflects the distribution of nonlinearity
and the sum of absolute differences between abundance maps estimated by LU and NLU.
It is interesting to note that the areas with large differences in abundance maps accord
well with the areas associated with large values of $\bfP$. In these shared areas, 
nonlinear effects can be expected as there exist multiple interactions 
between endmembers in the ridge of mountain, in the shadow of mountain in the lake, etc.


\subsubsection{Cuprite Dataset}
\begin{figure*}
\centering
\subfigure{
\includegraphics[width=0.25\textwidth]{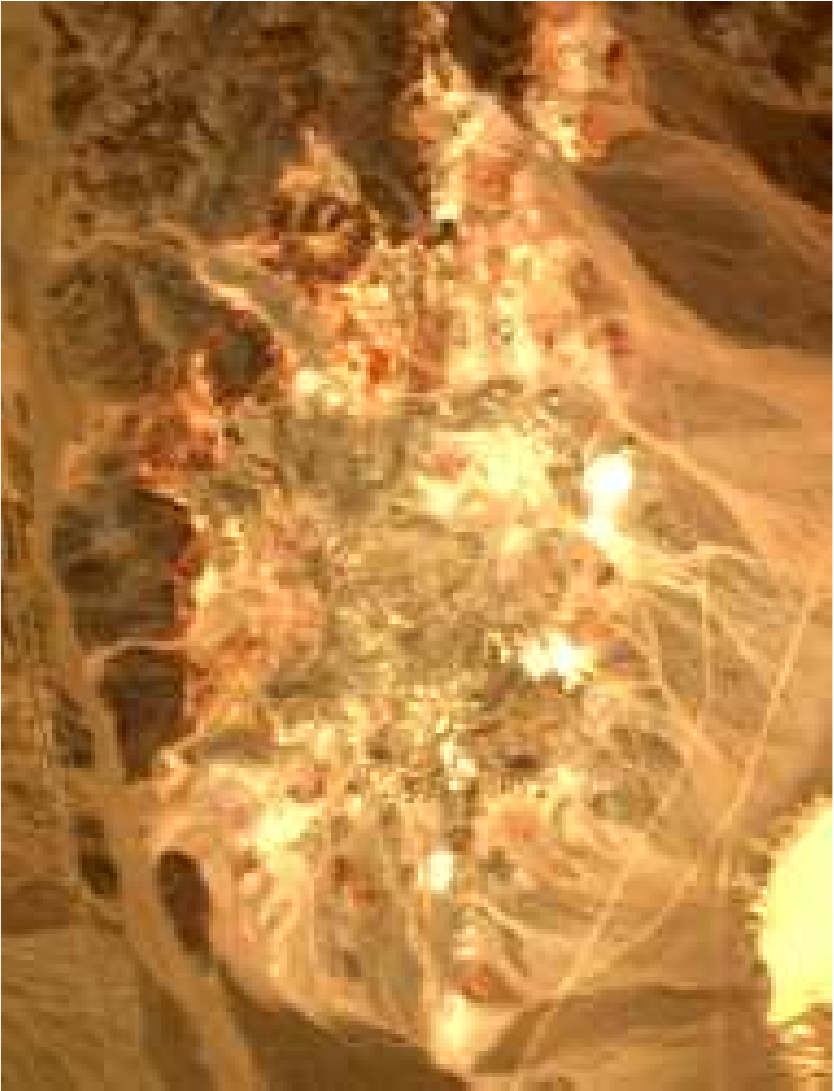}}
\subfigure{
\includegraphics[width=0.3\textwidth]{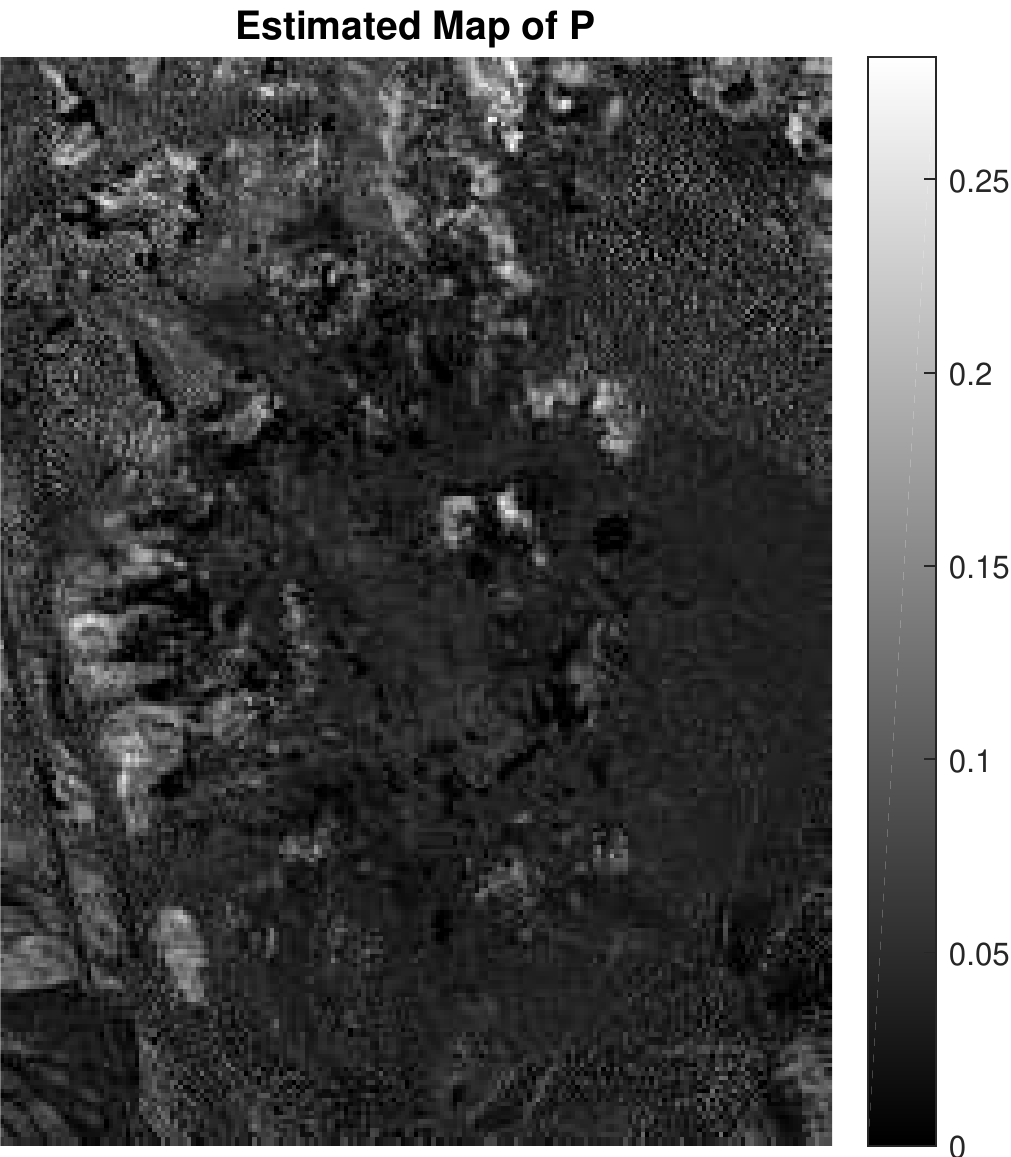}}
\subfigure{
\includegraphics[width=0.3\textwidth]{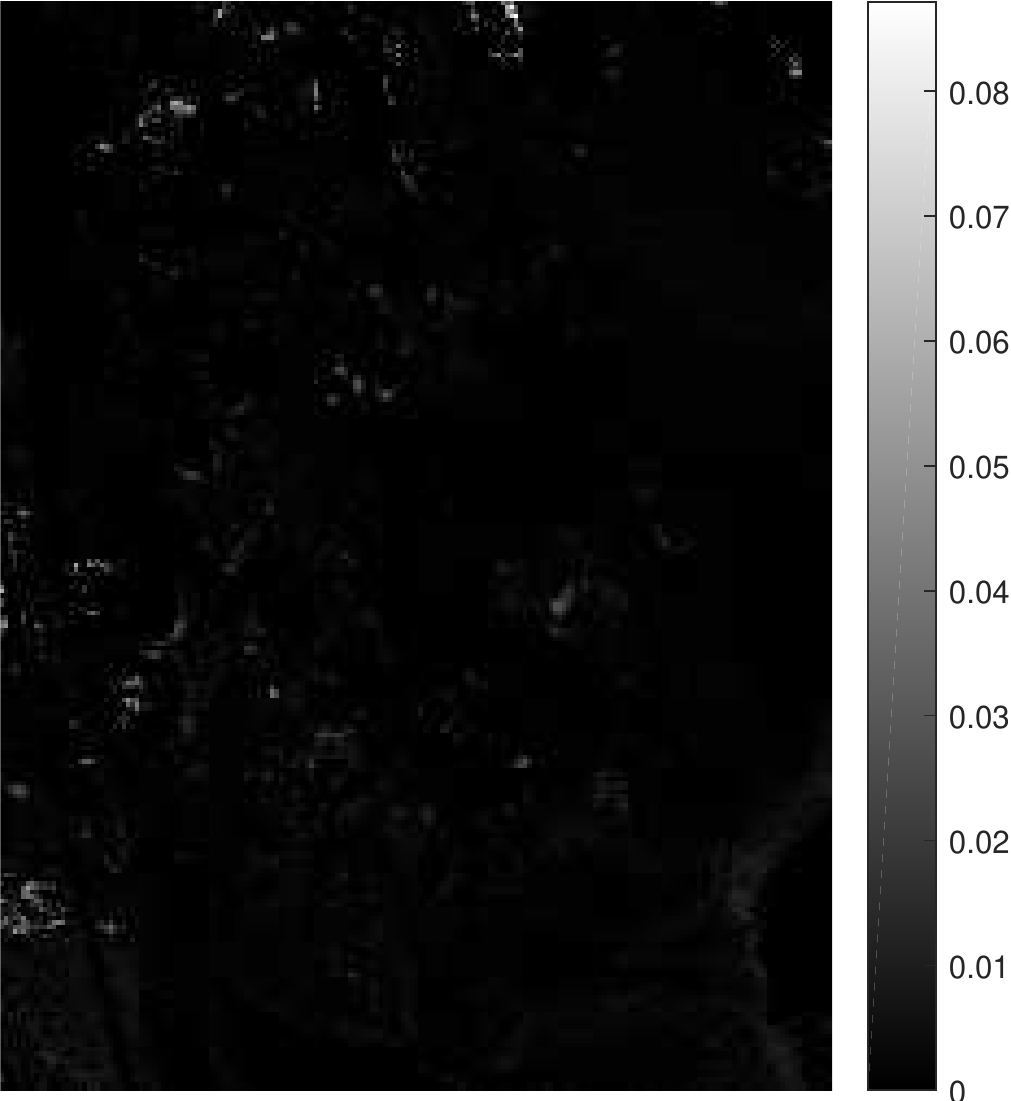}}
\caption{Left: Cuprite dataset. Middle: Estimated map $\hat{\bfP}$. Right: Sum of absolute differences between abundance maps estimated by LU and NLU.}
\label{fig:HS_Cuprite}
\end{figure*}

\begin{figure}[h!]
\centering
\includegraphics[width=0.5\textwidth]{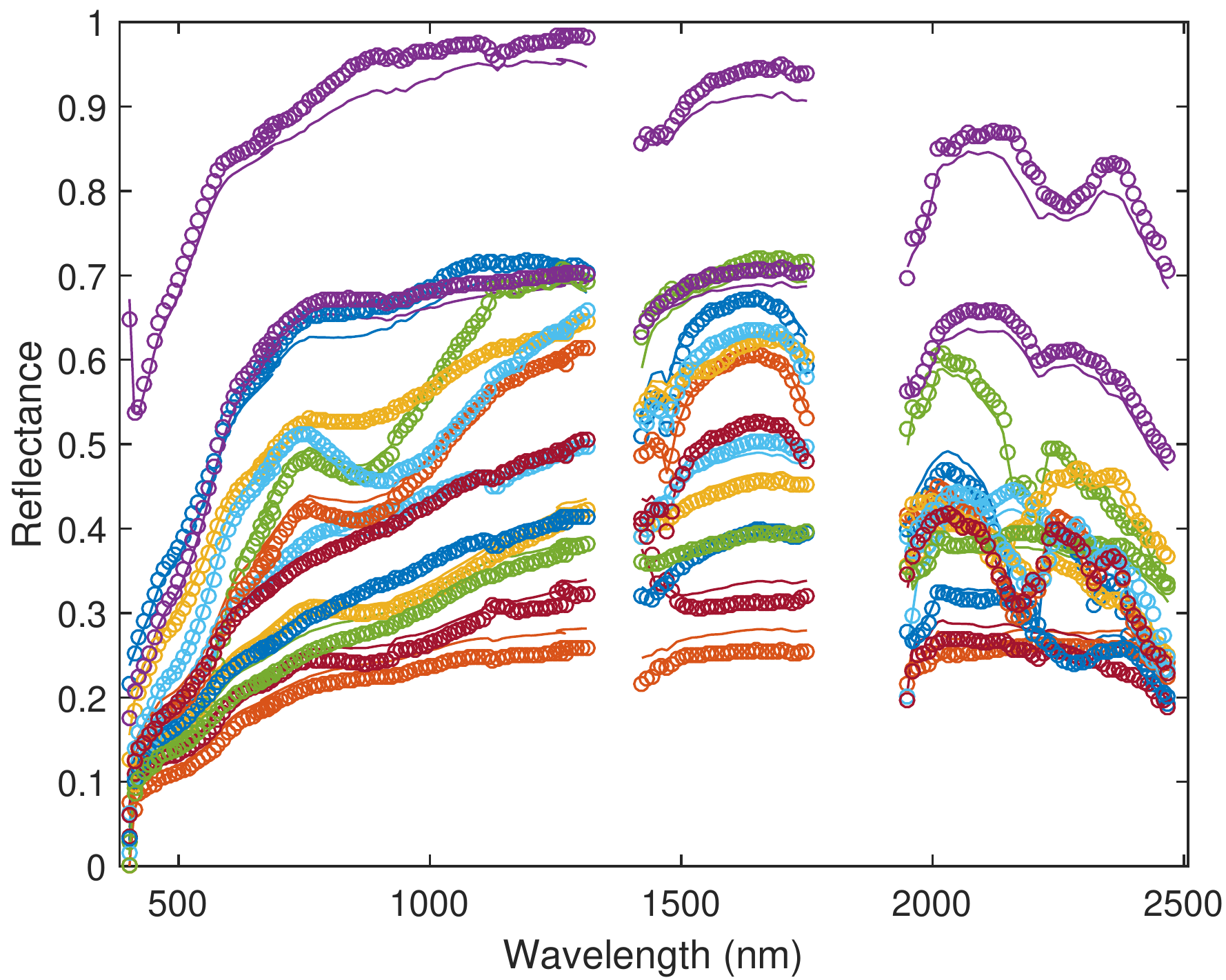}
\includegraphics[width=0.5\textwidth]{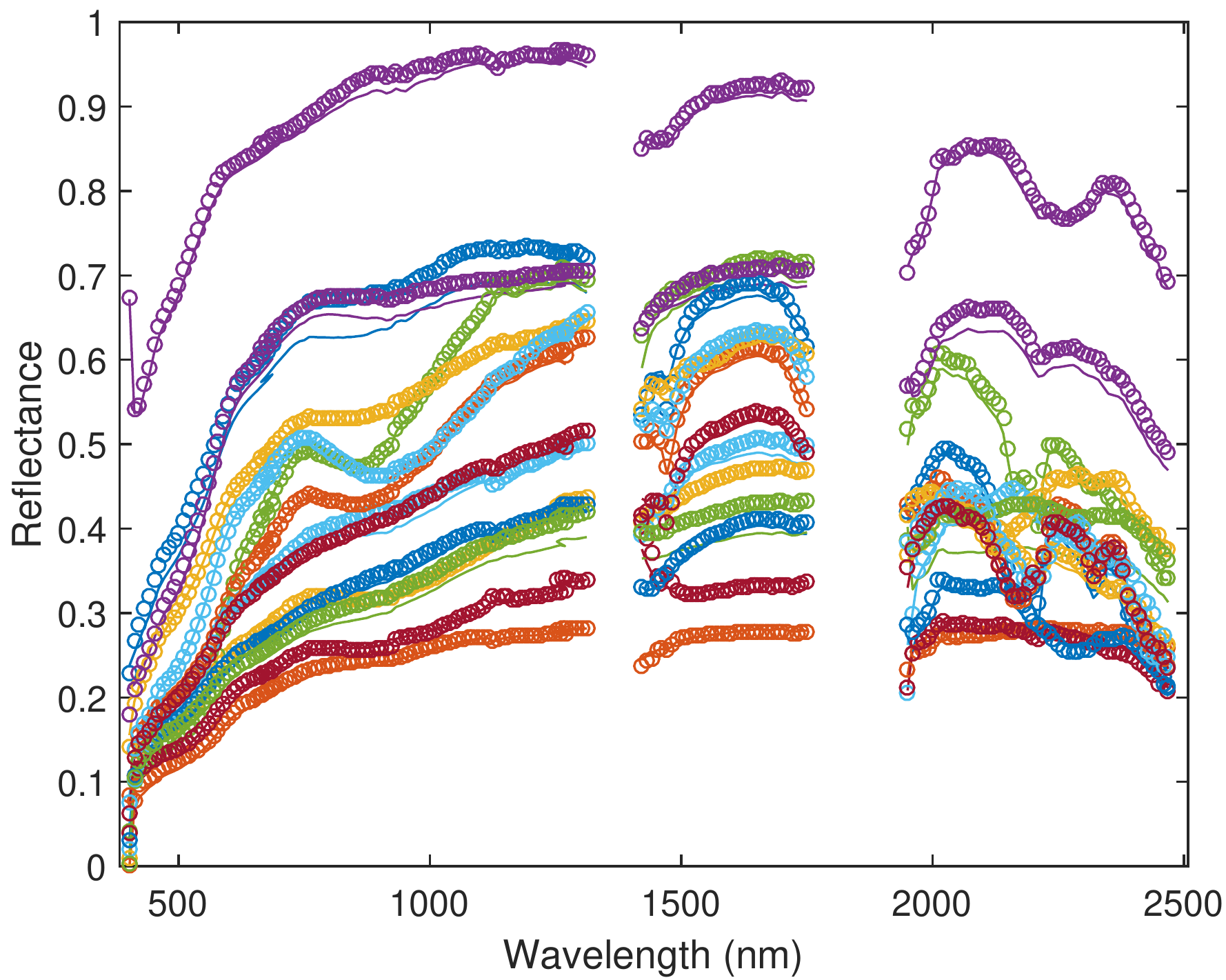}
\caption{Cuprite dataset: Extracted endmember signatures using (Top) VCA (solid) and LU (circle), (Bottom) VCA (solid) and NLU (circle).}
\label{fig:End_Cuprite}
\end{figure}

\begin{figure*}[t!]
\centering
    \subfigure[Alunite]{
    \includegraphics[width=0.18\textwidth]{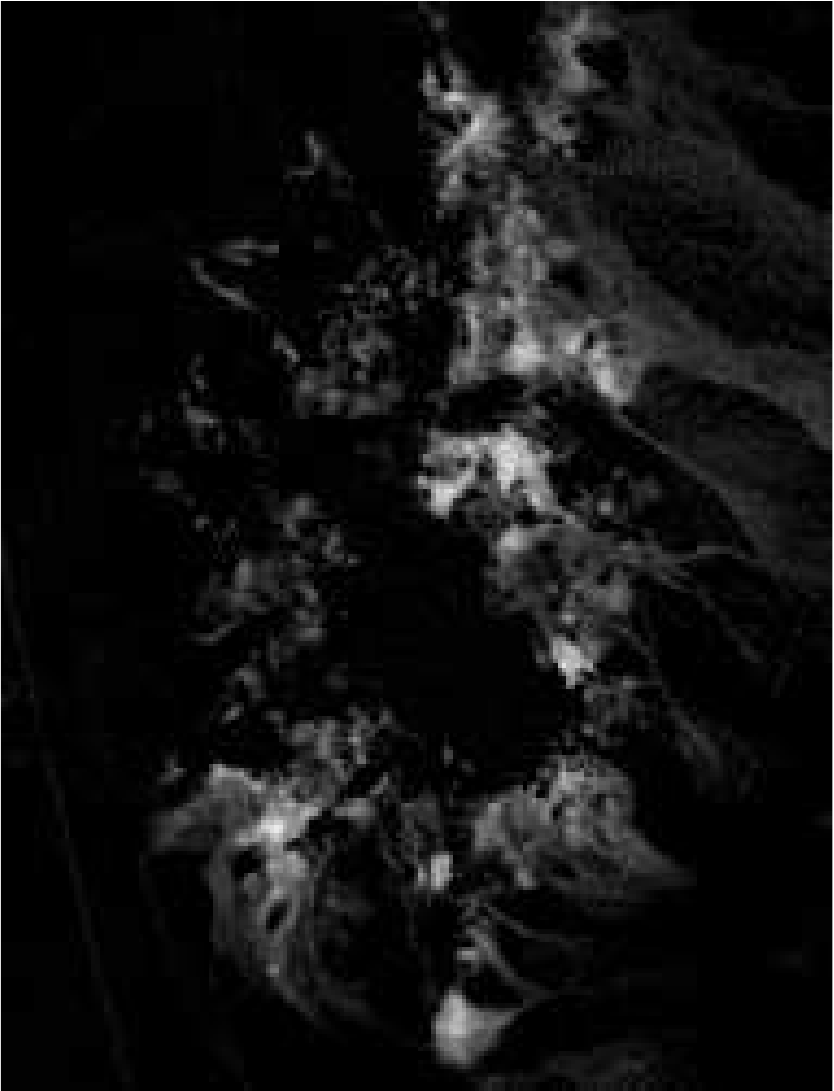}}
    \subfigure[Sphene]{
    \includegraphics[width=0.18\textwidth]{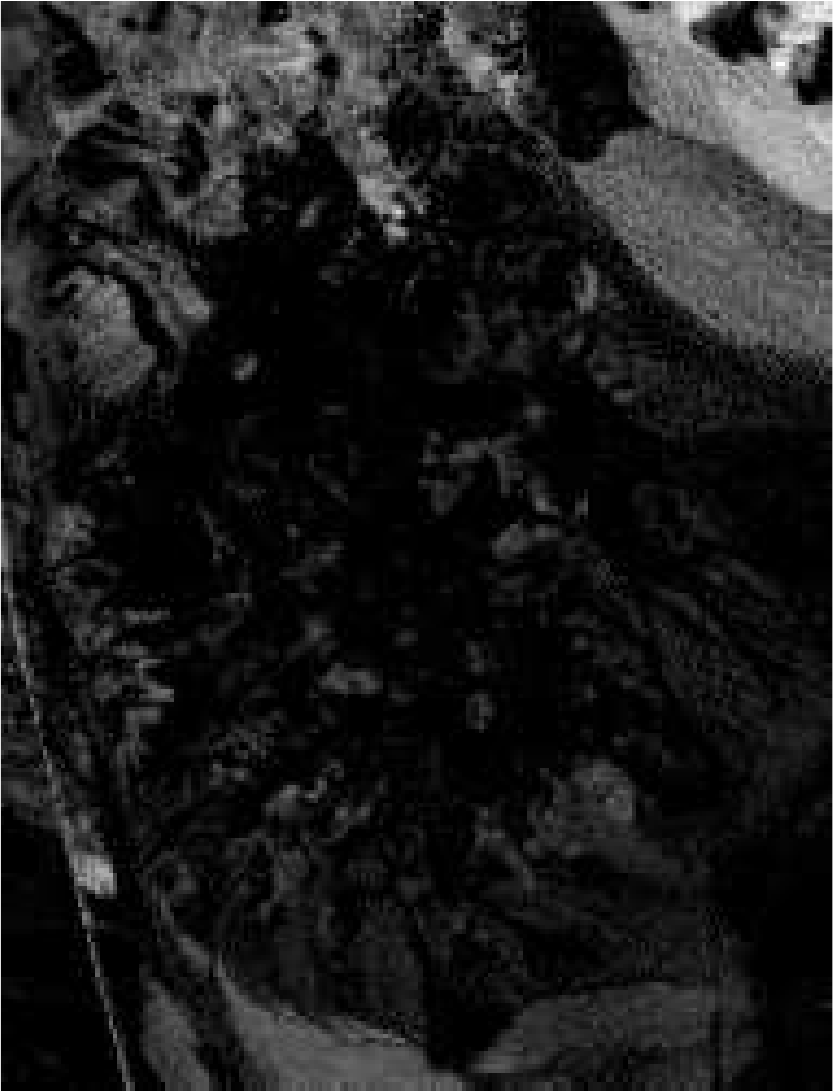}}
    \subfigure[Andradite]{
    \includegraphics[width=0.18\textwidth]{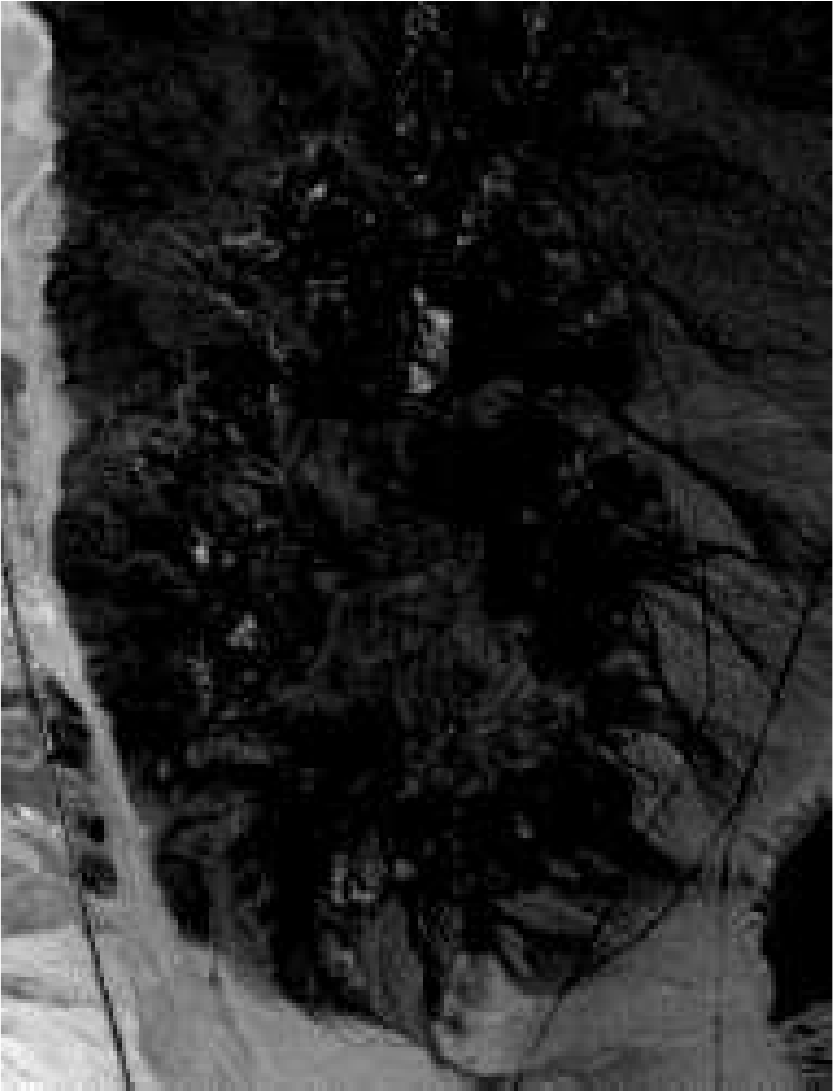}}
    \subfigure[Muscovite]{
    \includegraphics[width=0.18\textwidth]{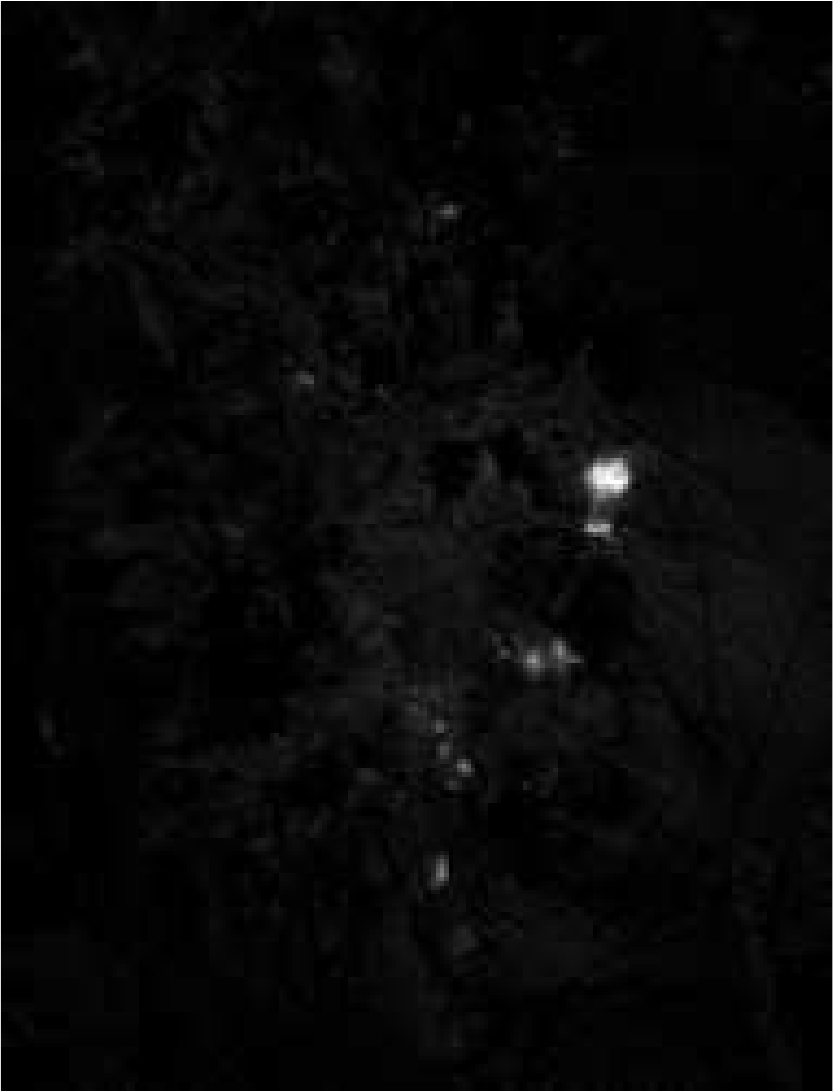}}
    \subfigure[Kaolinite \#1]{
    \includegraphics[width=0.18\textwidth]{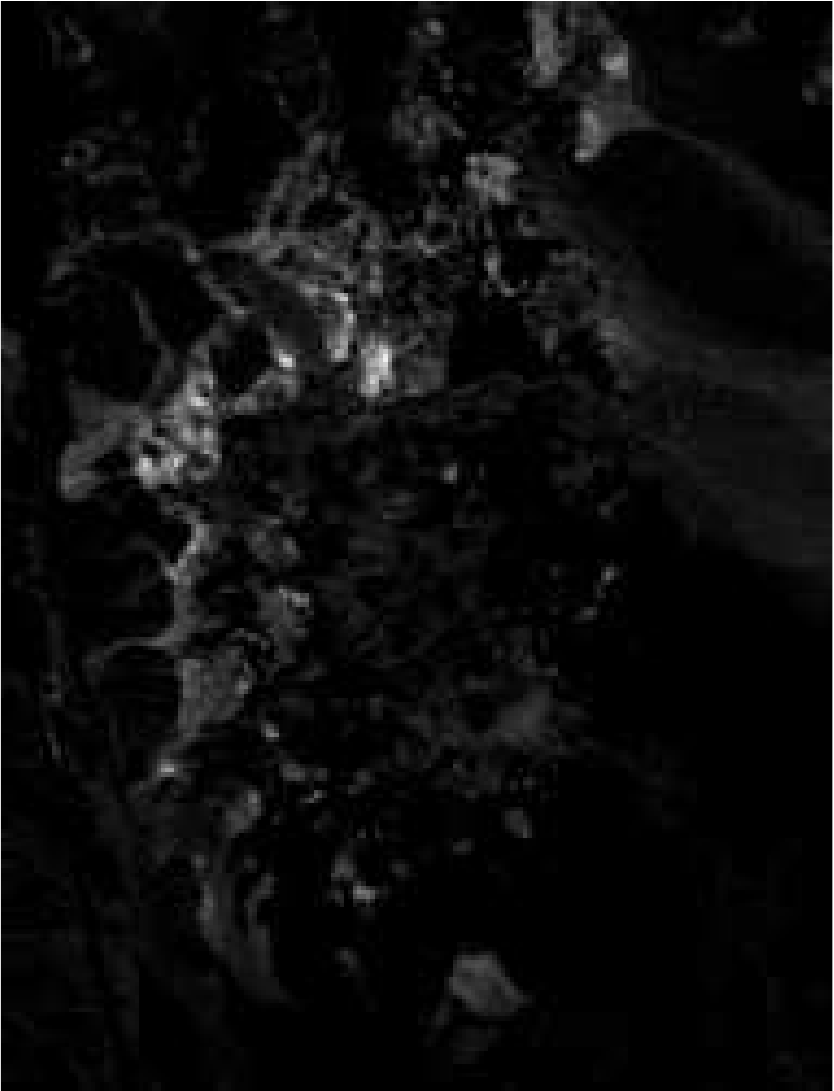}}\\
    \subfigure[Nontronite]{
    \includegraphics[width=0.18\textwidth]{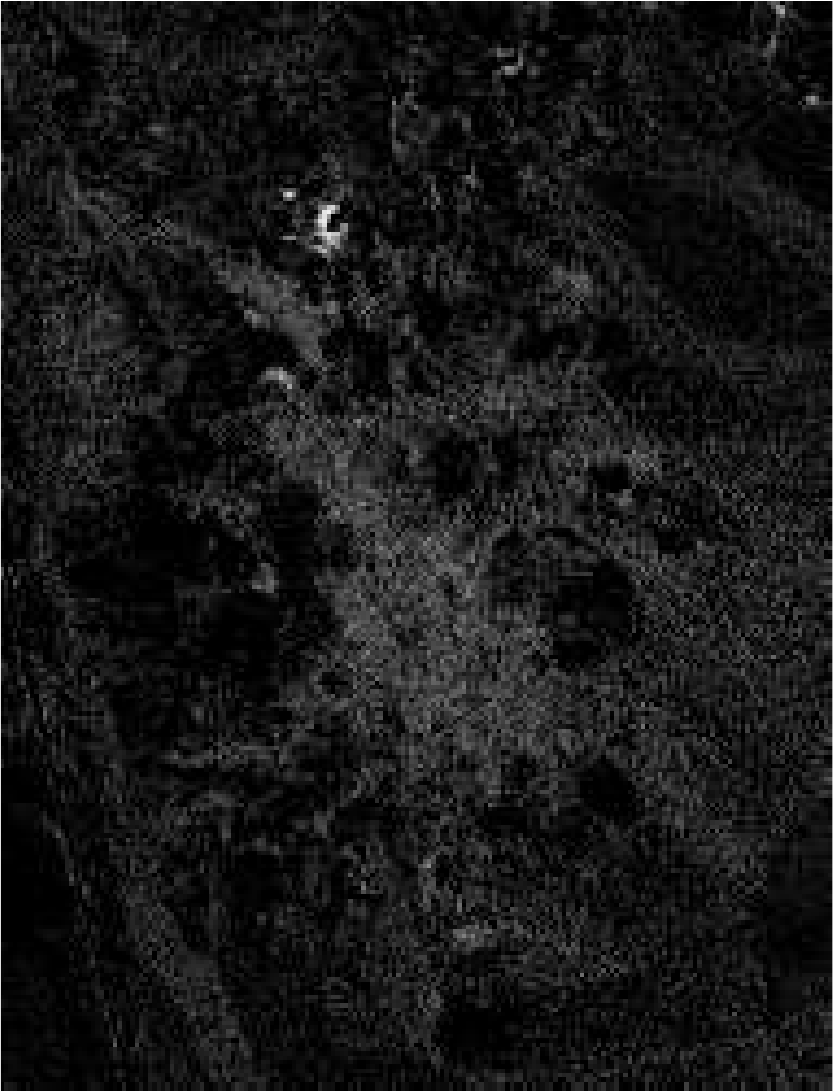}}
    \subfigure[Kaolinite \#4]{
    \includegraphics[width=0.18\textwidth]{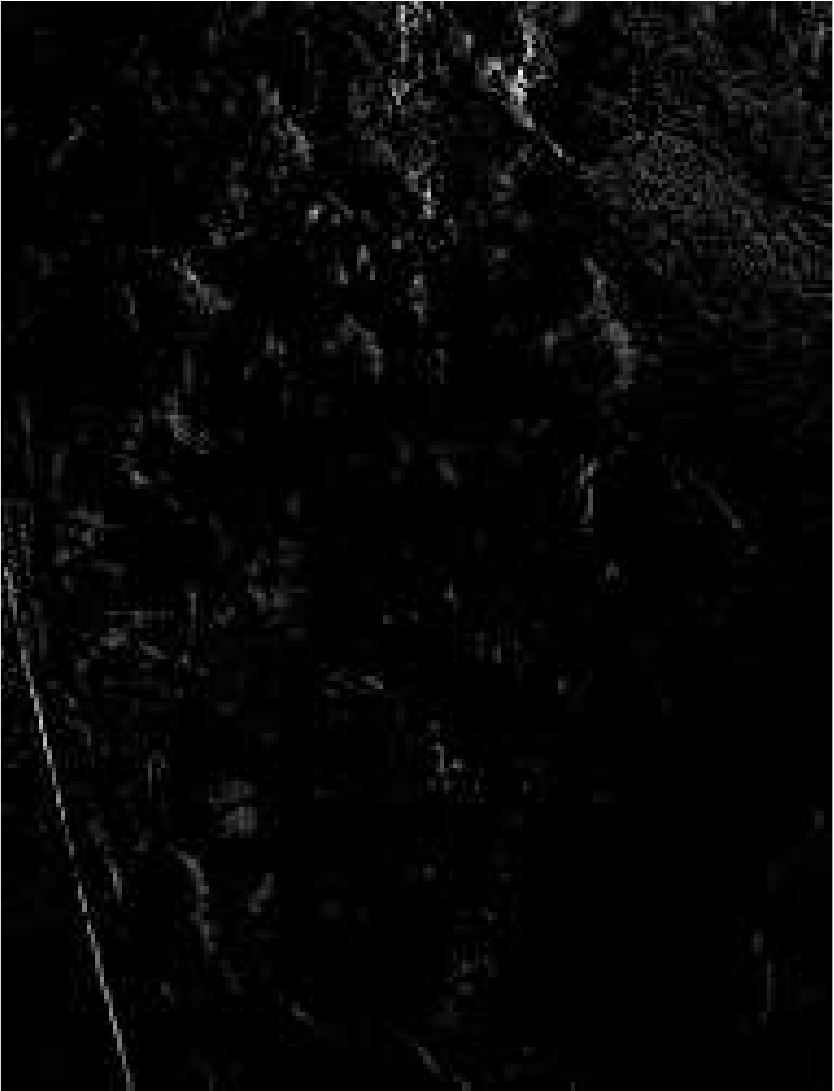}}
    \subfigure[Kaolinite \#3]{
    \includegraphics[width=0.18\textwidth]{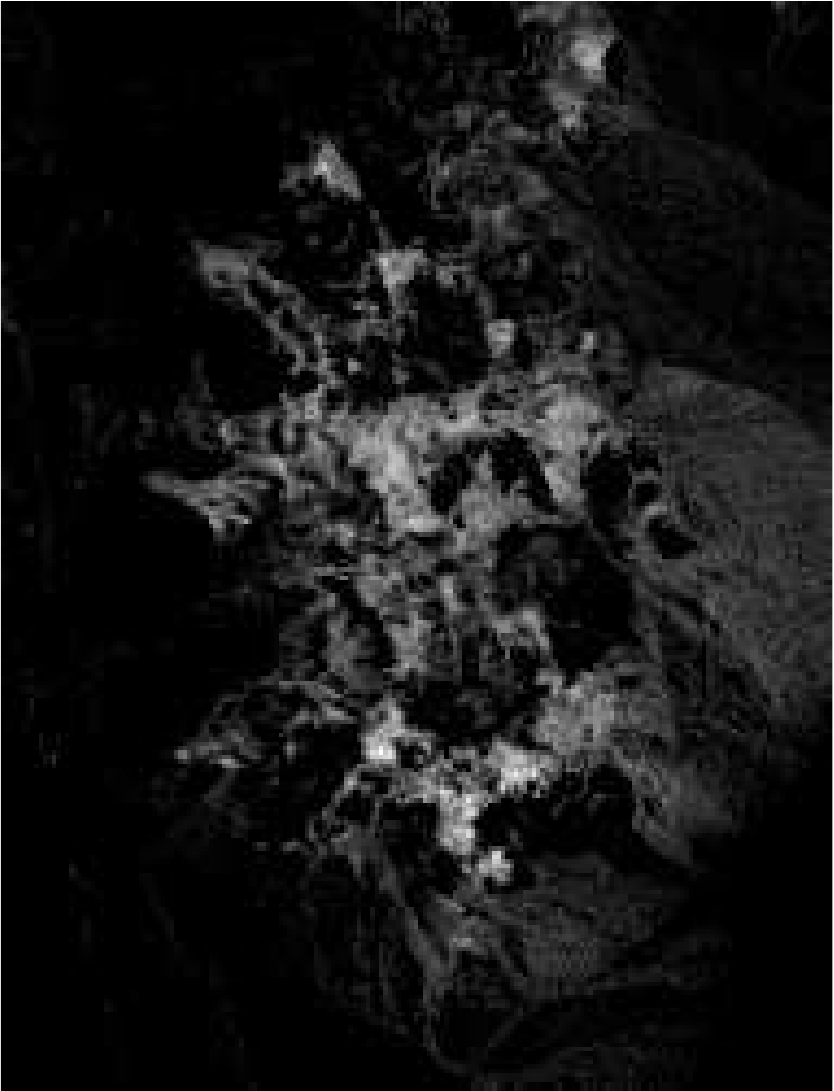}}
    \subfigure[Pyrope \#2]{
    \includegraphics[width=0.18\textwidth]{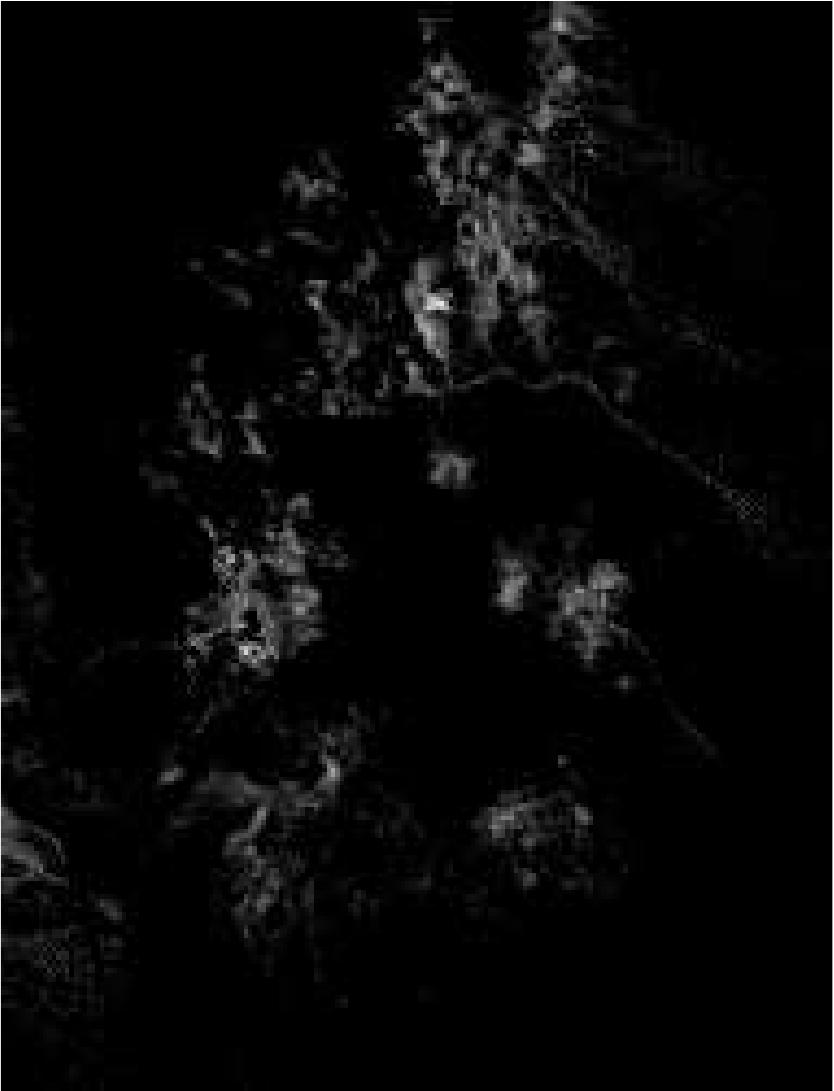}}
    \subfigure[Buddingtonite]{
    \includegraphics[width=0.18\textwidth]{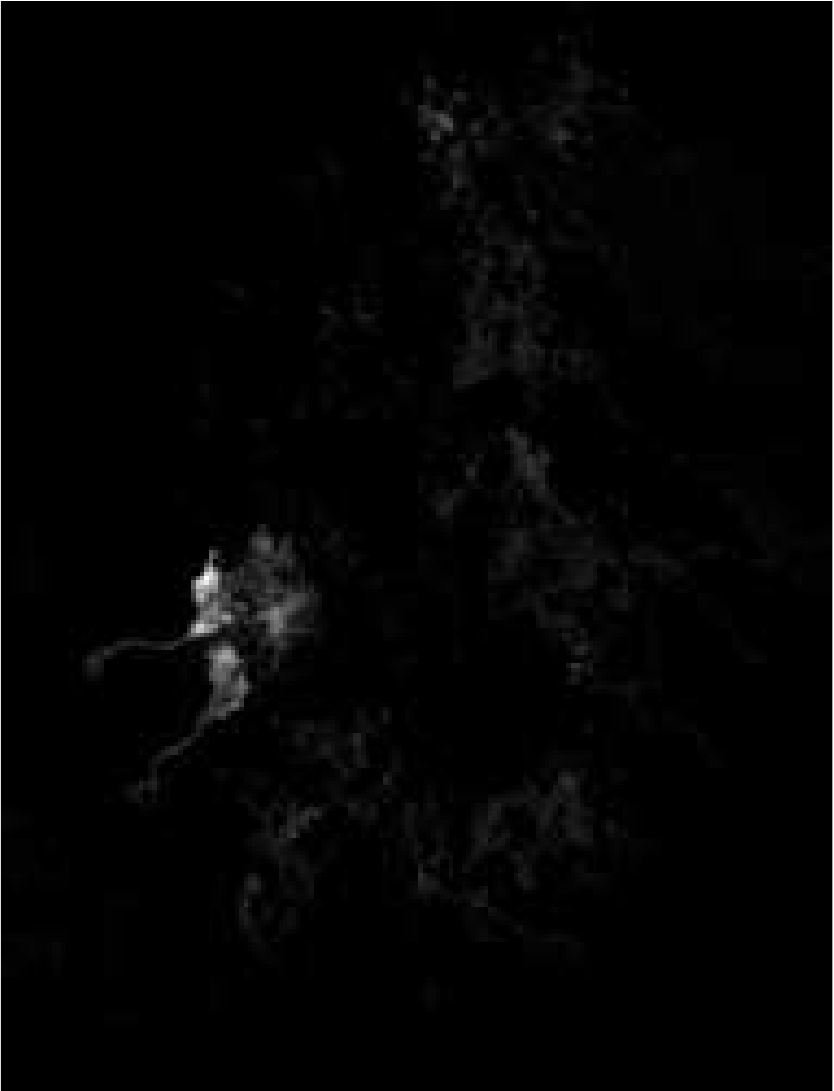}}\\
    \subfigure[Montmorillonite]{
    \includegraphics[width=0.18\textwidth]{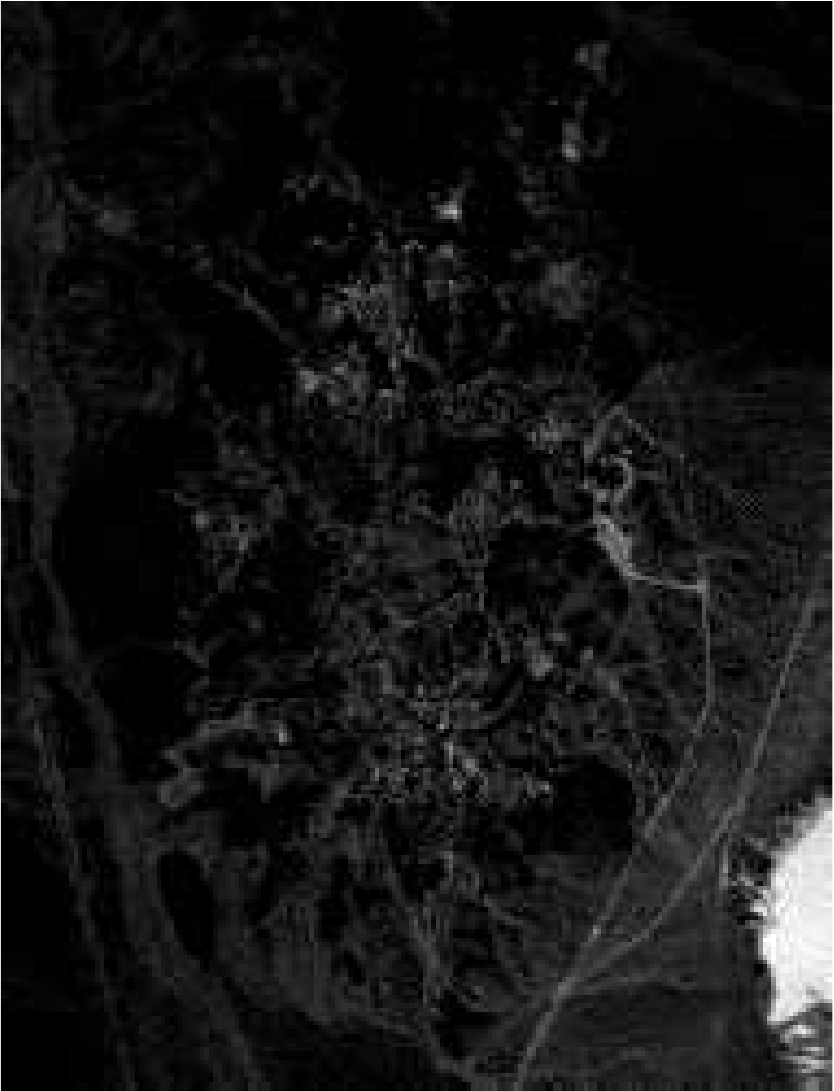}}
    \subfigure[Pyrope \#1]{
    \includegraphics[width=0.18\textwidth]{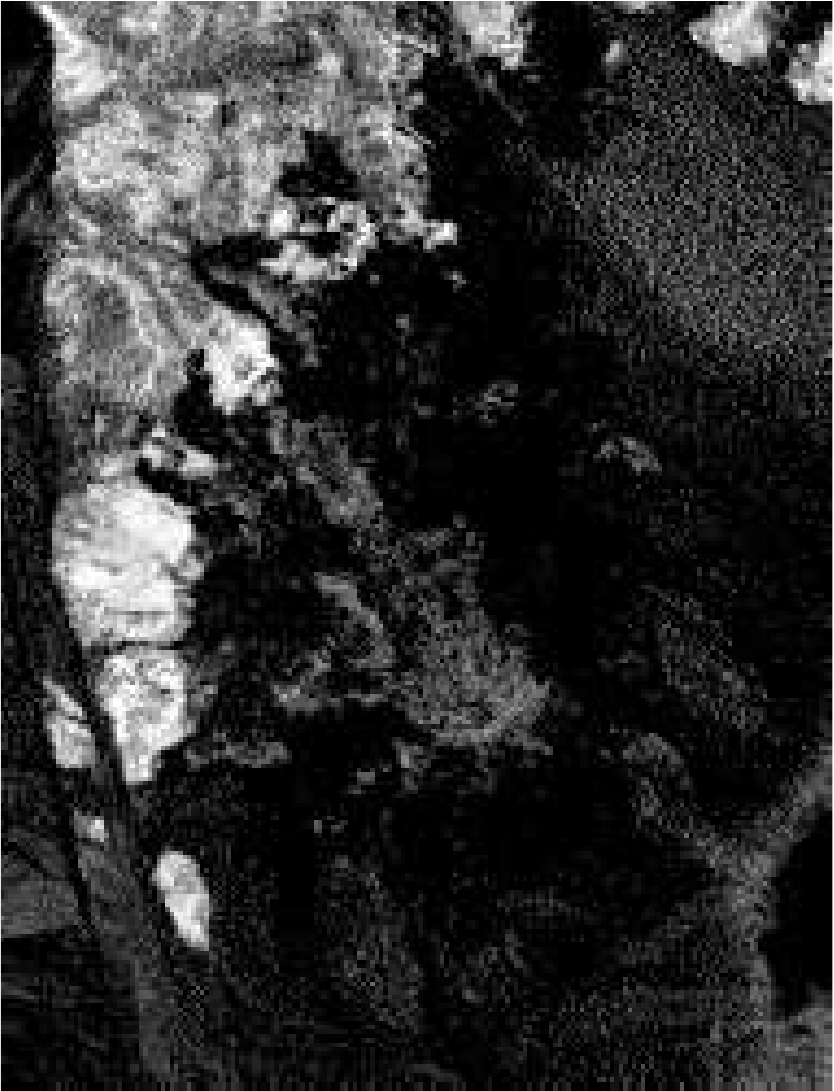}}
    \subfigure[Kaolinite \#2]{
    \includegraphics[width=0.18\textwidth]{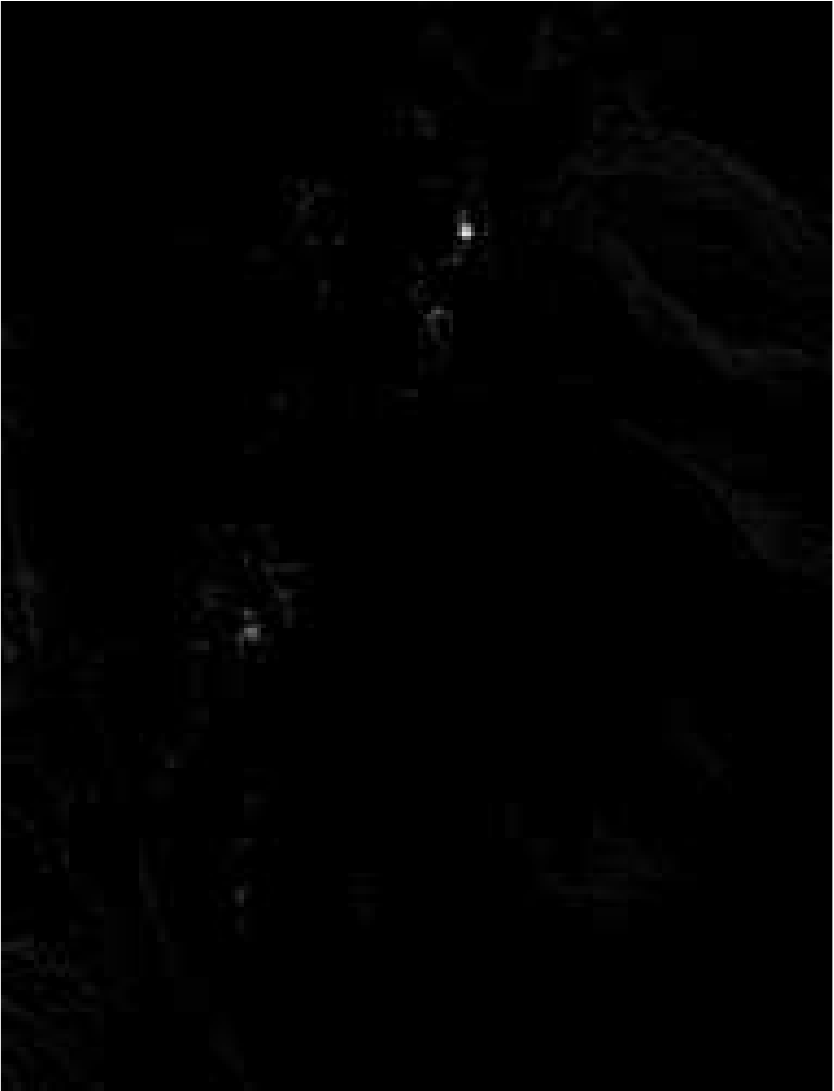}}
    \subfigure[Dumortierite]{
    \includegraphics[width=0.18\textwidth]{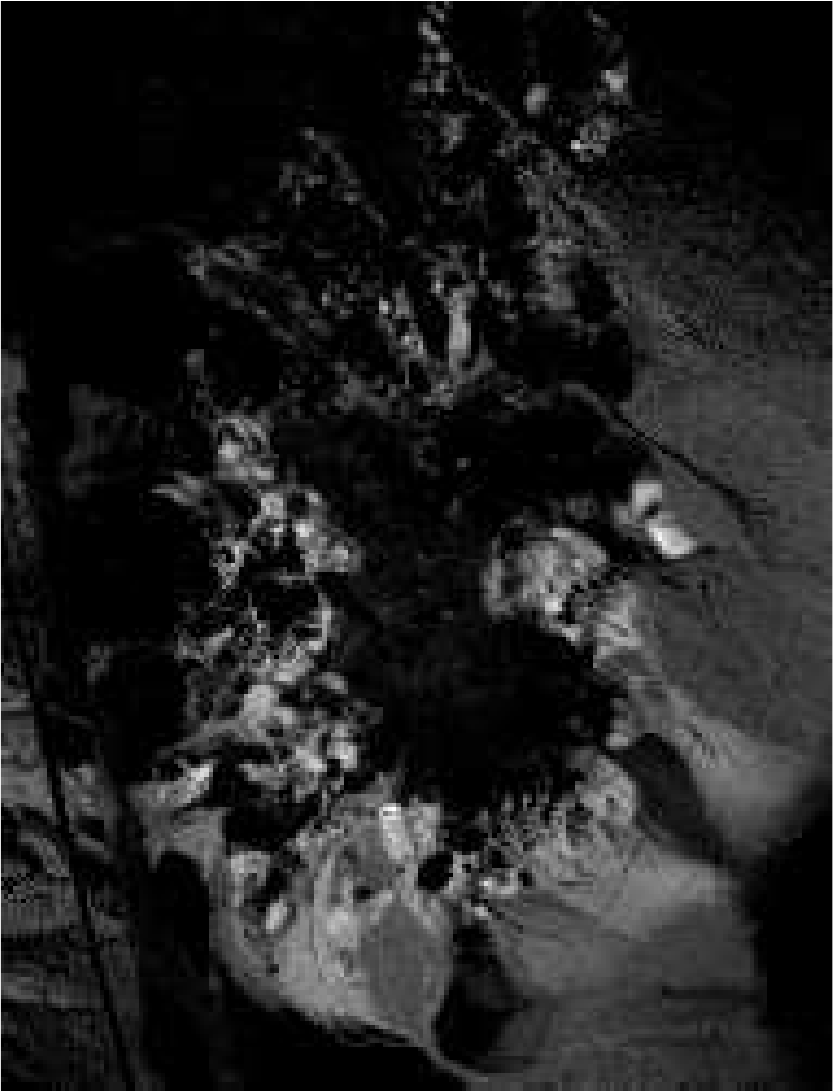}}
    \caption{Cuprite dataset: abundance maps estimated by LU.}
\label{fig:Real_Abu_Cuprite_LU}
\end{figure*}

\begin{figure*}[t!]
\centering
    \subfigure[Alunite]{
    \includegraphics[width=0.18\textwidth]{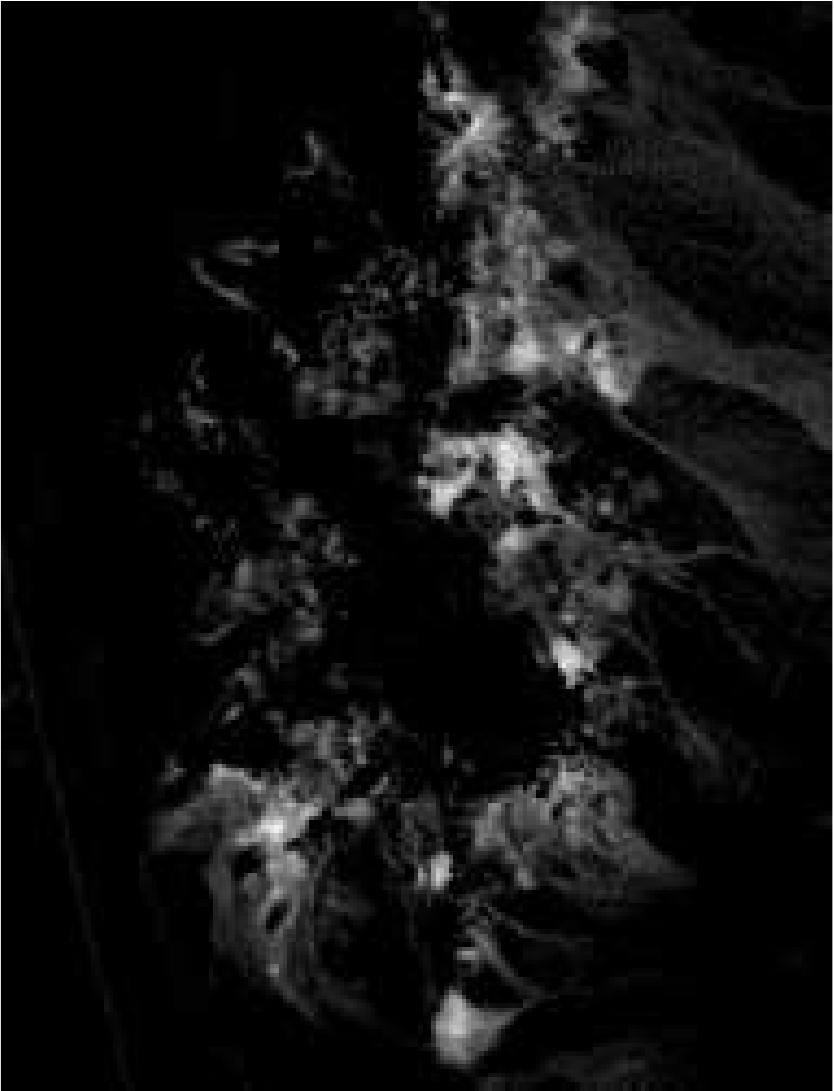}}
    \subfigure[Sphene]{
    \includegraphics[width=0.18\textwidth]{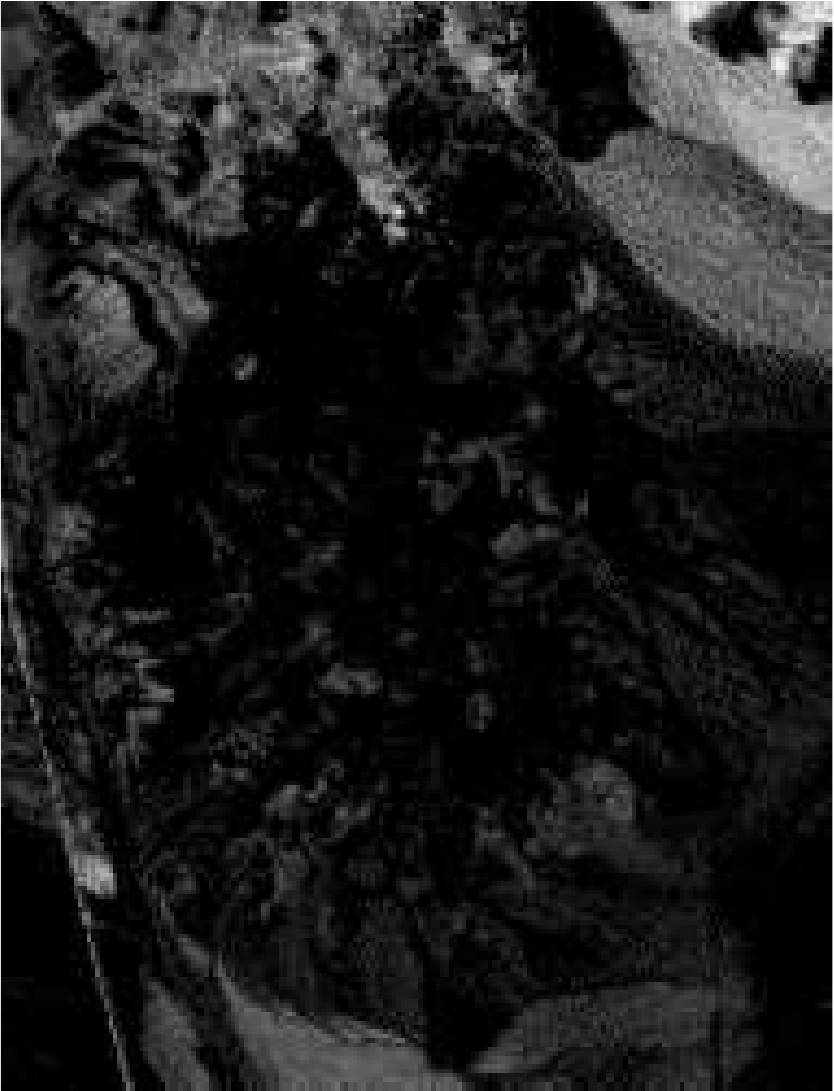}}
    \subfigure[Andradite]{
    \includegraphics[width=0.18\textwidth]{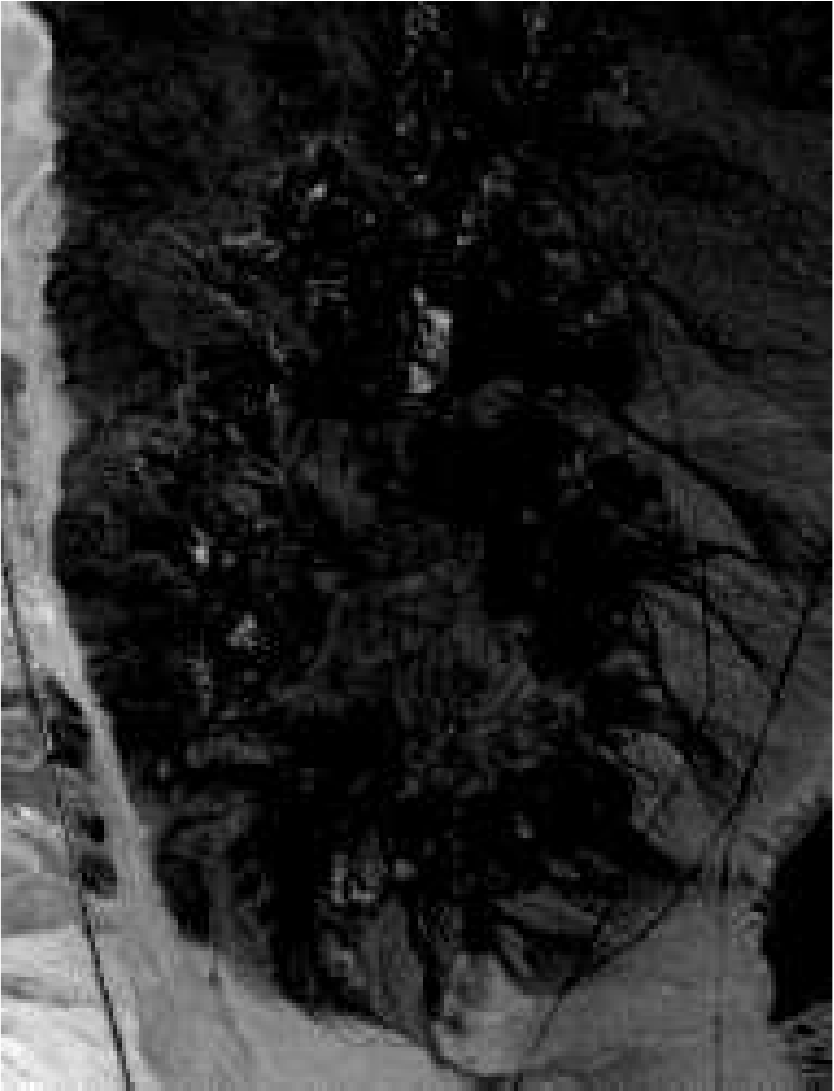}}
    \subfigure[Muscovite]{
    \includegraphics[width=0.18\textwidth]{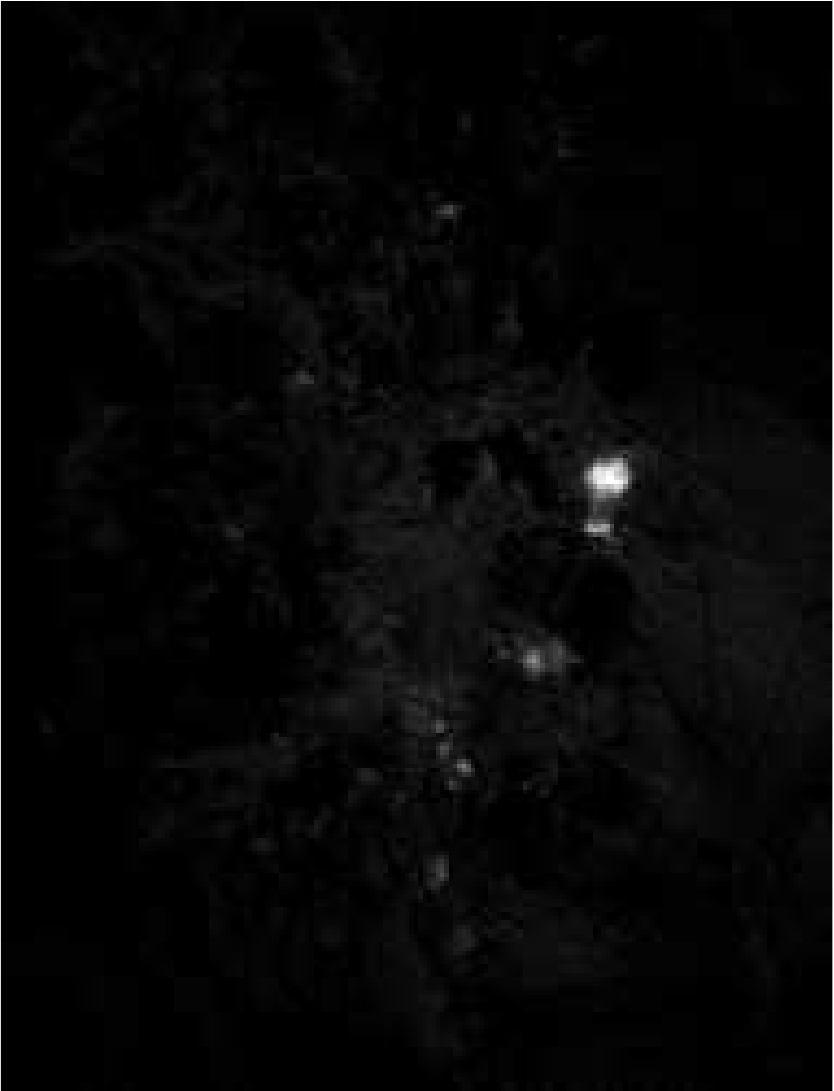}}
    \subfigure[Kaolinite \#1]{
    \includegraphics[width=0.18\textwidth]{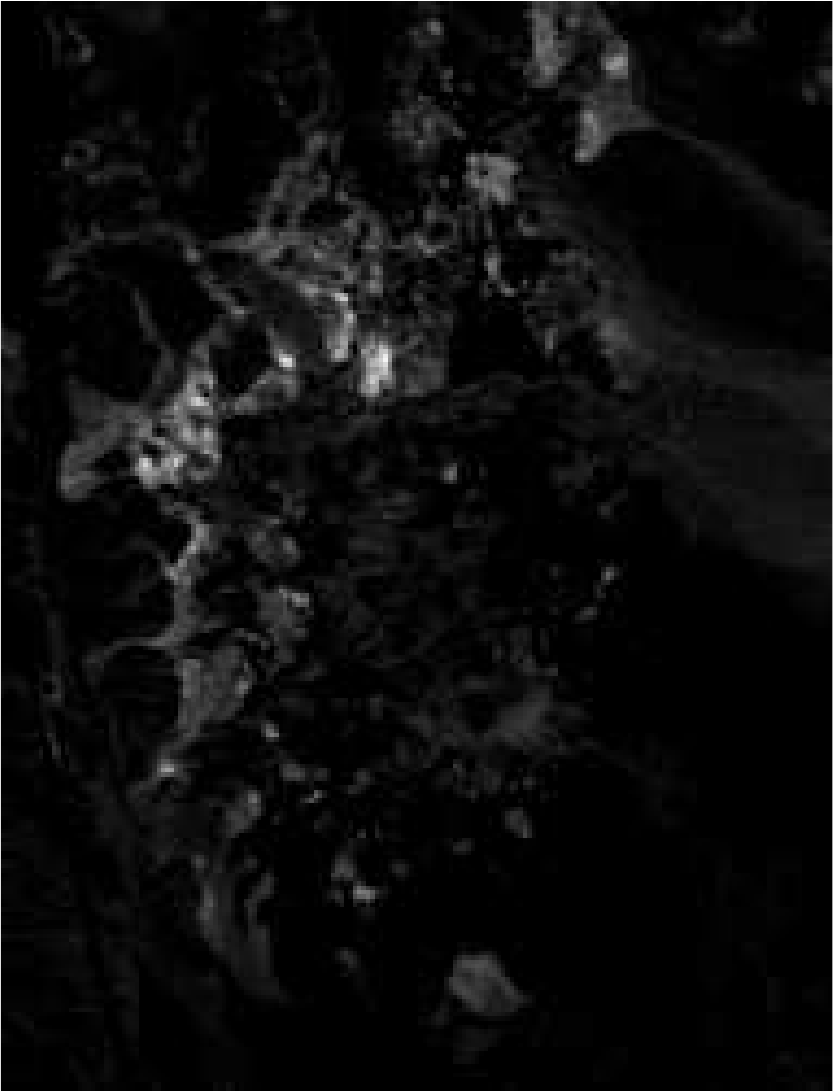}}\\
    \subfigure[Nontronite]{
    \includegraphics[width=0.18\textwidth]{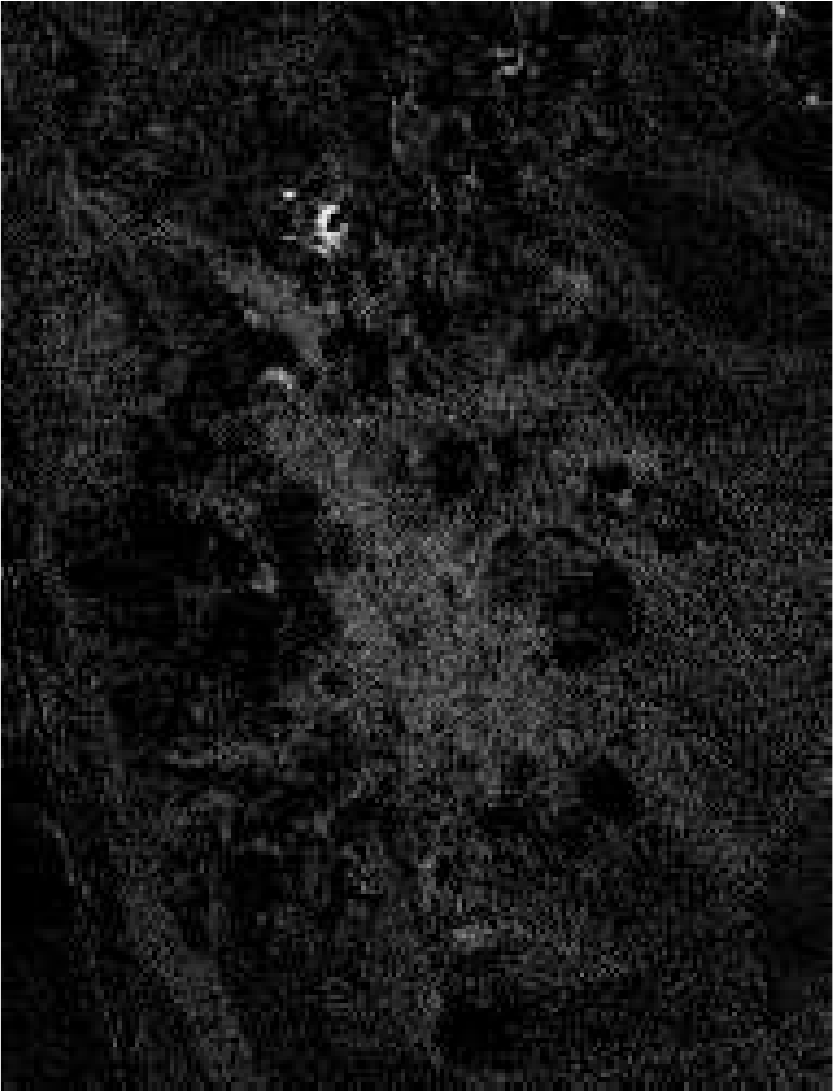}}
    \subfigure[Kaolinite \#4]{
    \includegraphics[width=0.18\textwidth]{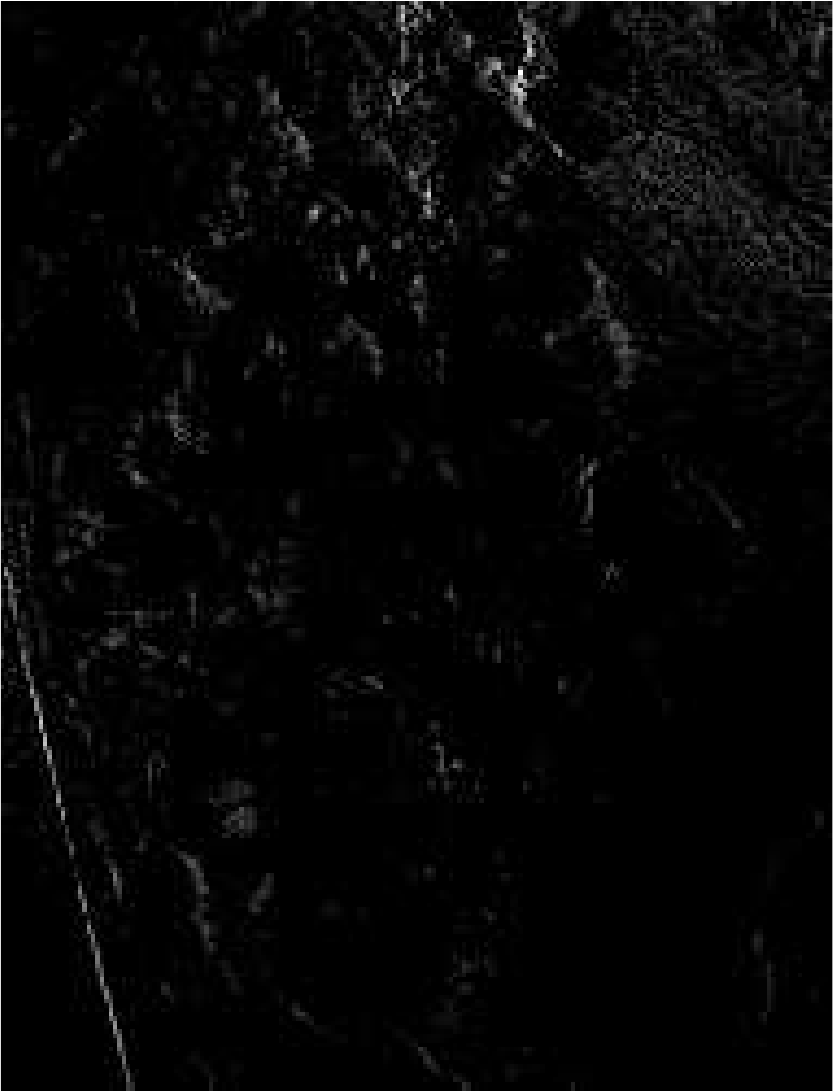}}
    \subfigure[Kaolinite \#3]{
    \includegraphics[width=0.18\textwidth]{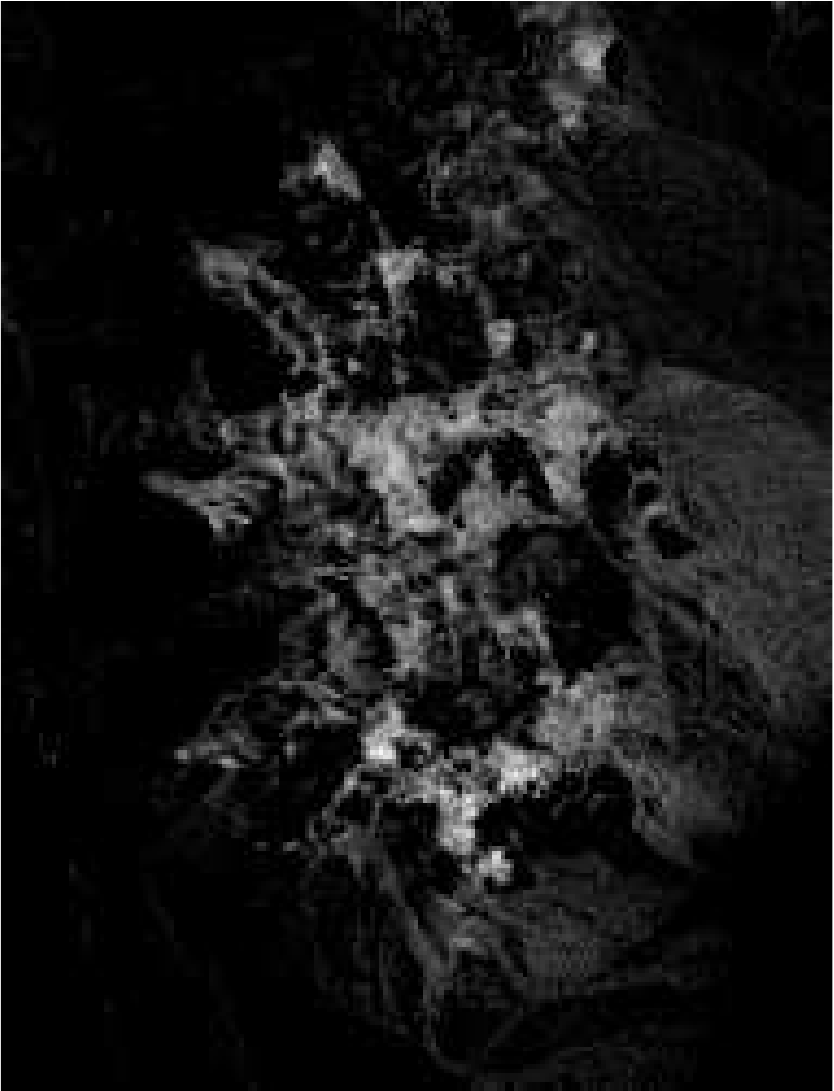}}
    \subfigure[Pyrope \#2]{
    \includegraphics[width=0.18\textwidth]{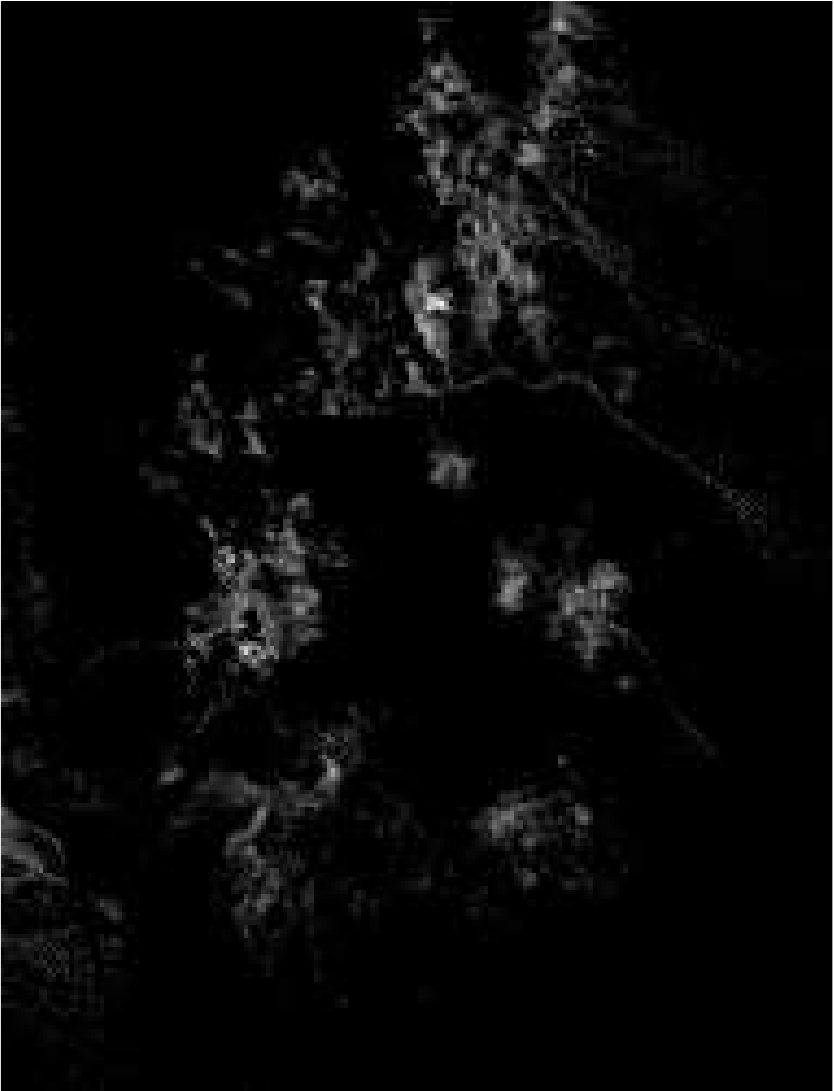}}
    \subfigure[Buddingtonite]{
    \includegraphics[width=0.18\textwidth]{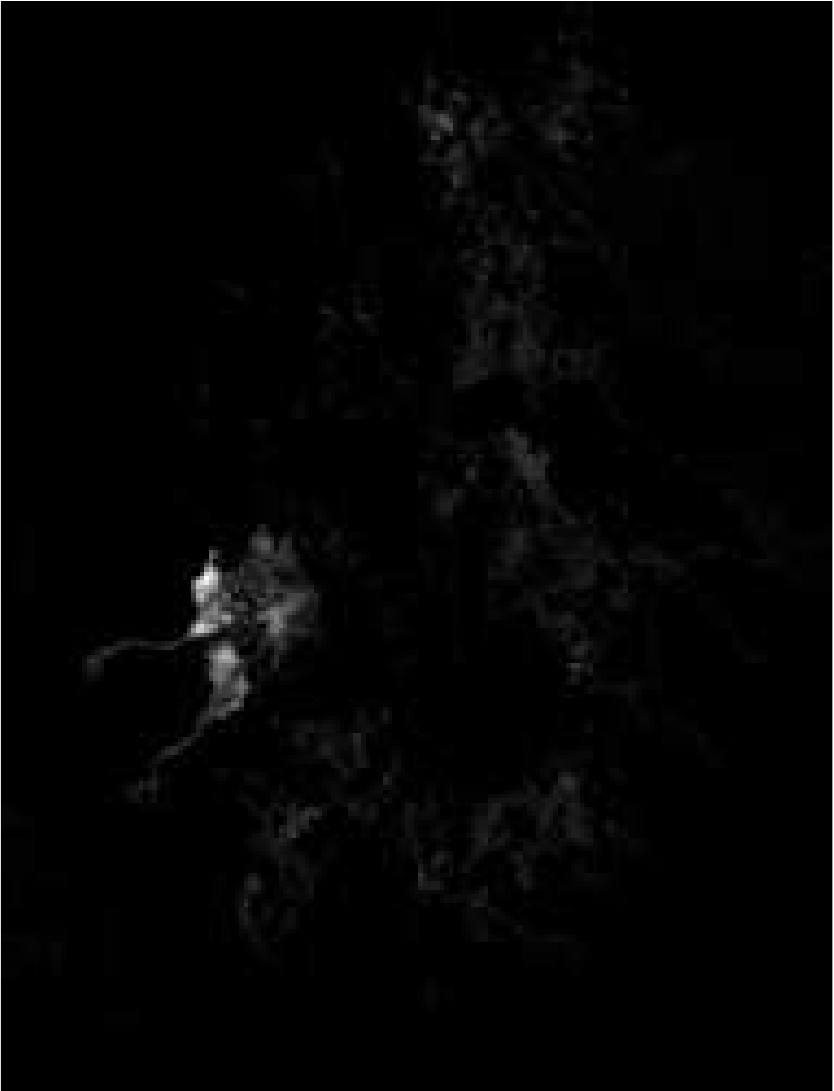}}\\
    \subfigure[Montmorillonite]{
    \includegraphics[width=0.18\textwidth]{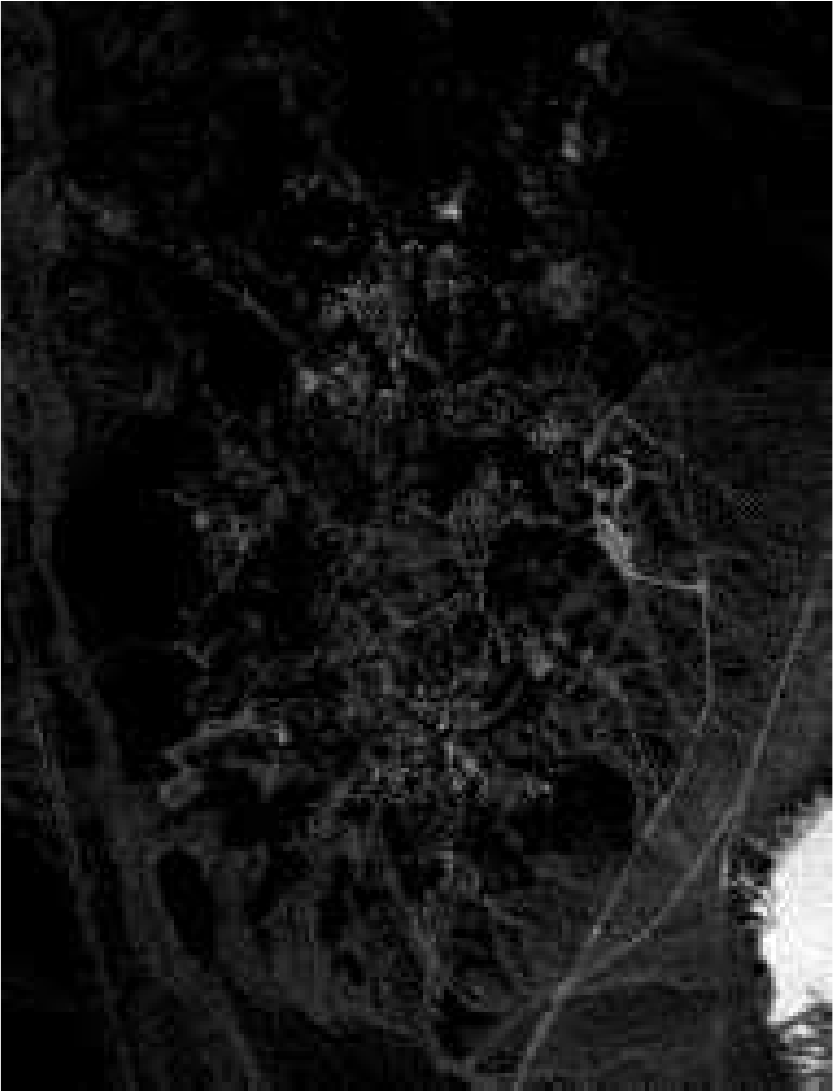}}
    \subfigure[Pyrope \#1]{
    \includegraphics[width=0.18\textwidth]{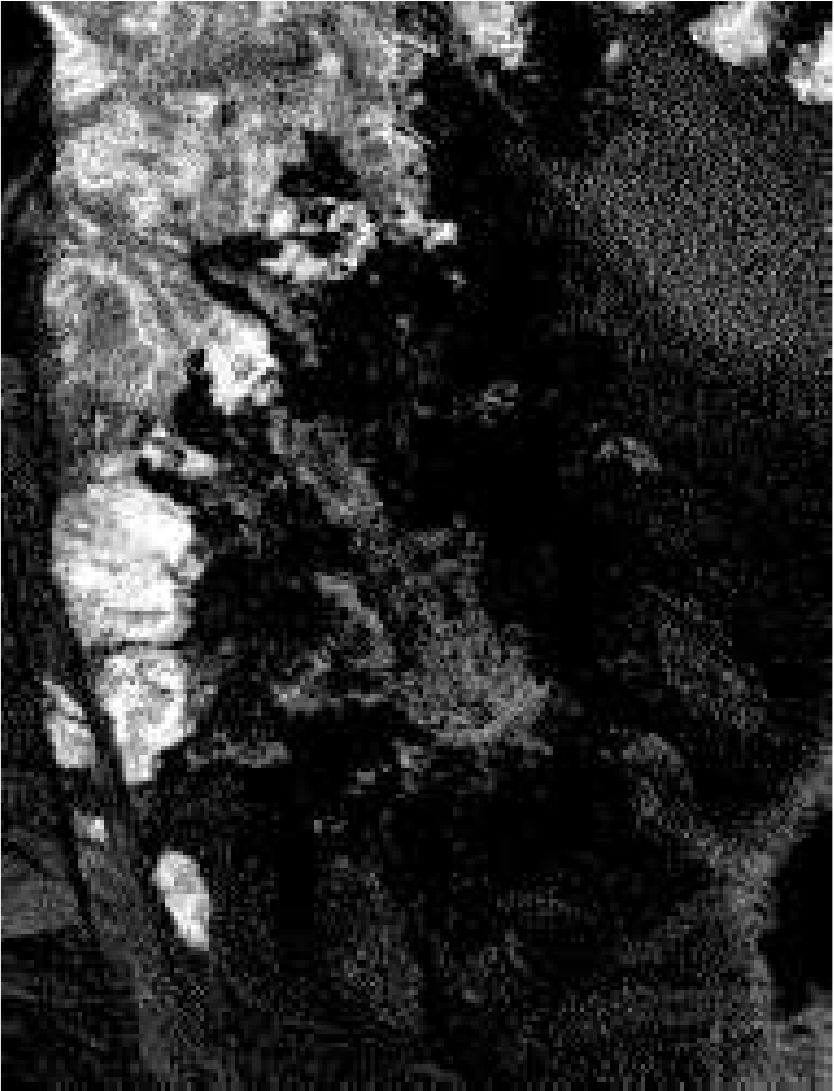}}
    \subfigure[Kaolinite \#2]{
    \includegraphics[width=0.18\textwidth]{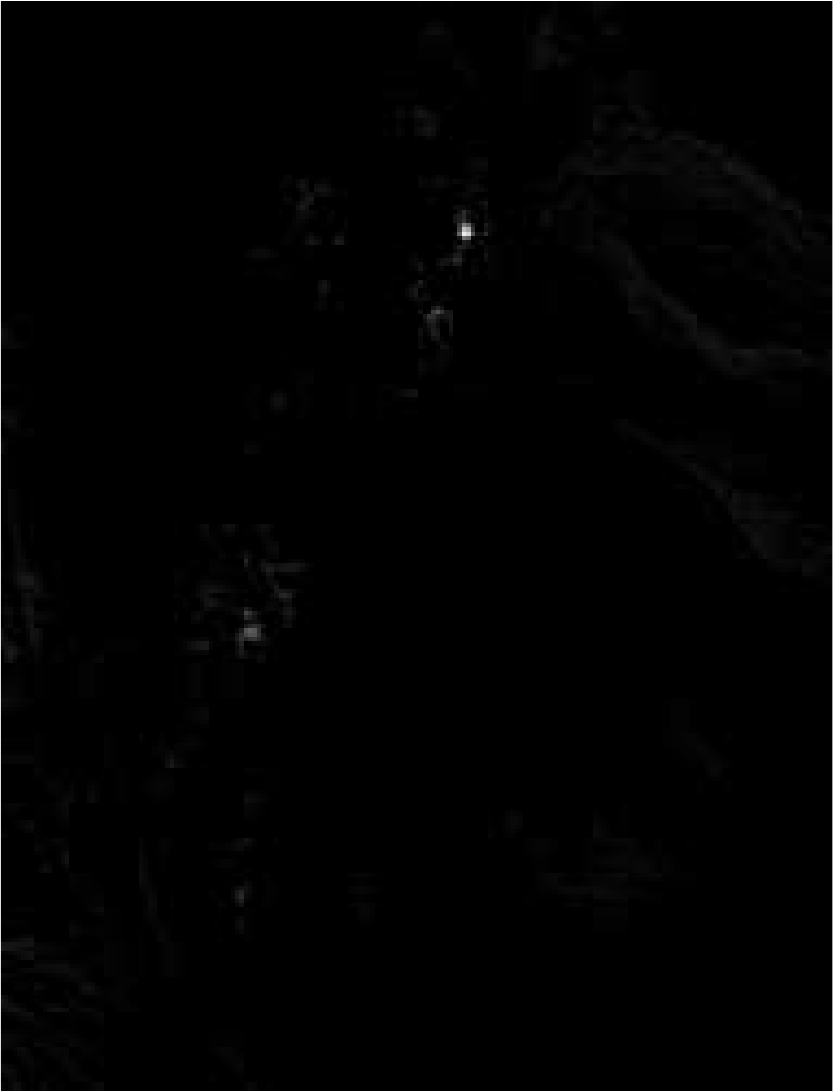}}
    \subfigure[Dumortierite]{
    \includegraphics[width=0.18\textwidth]{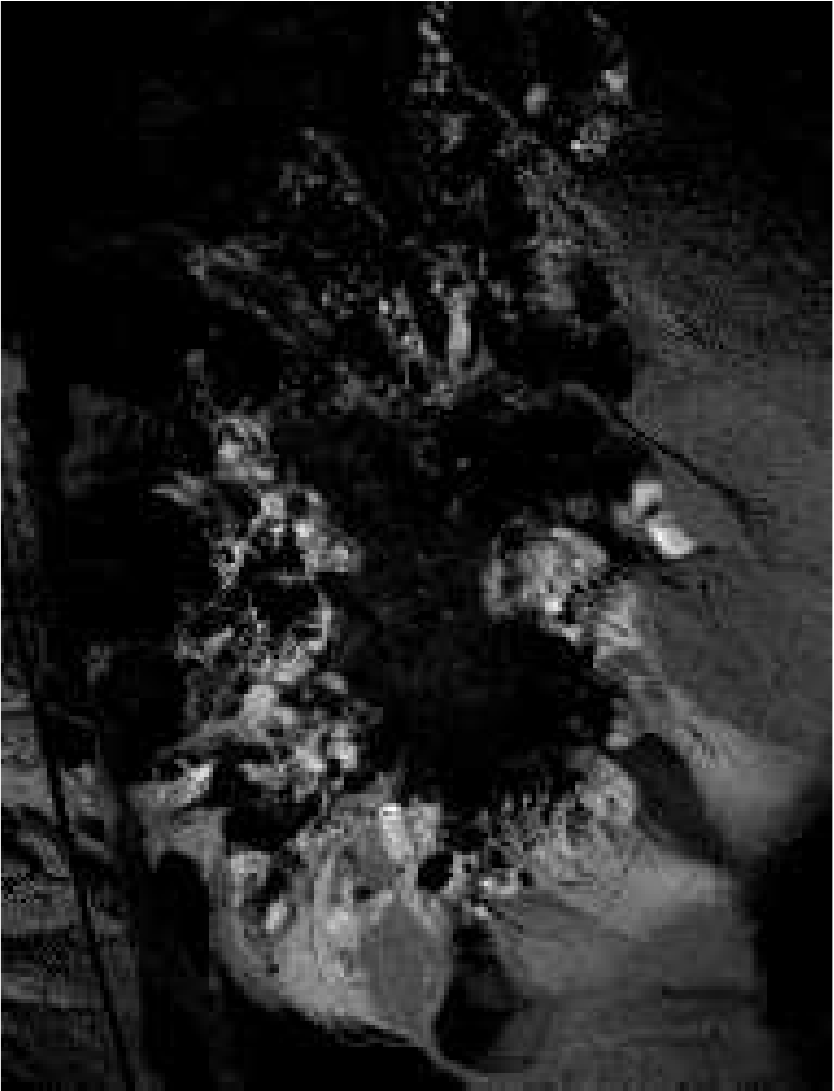}}
    \caption{Cuprite dataset: abundance maps estimated by NLU.}
\label{fig:Real_Abu_Cuprite_NLU}
\end{figure*}

This section investigates the performance of the proposed NLU method for unmixing the well-known Cuprite HS image. 
This image, which has received a lot of interest in the remote sensing and geoscience literature, was acquired over Cuprite
field by AVIRIS. 
It corresponds to a mining area in southern Nevada composed of several minerals and some vegetation, located 
approximately $200$km northwest of Las Vegas.
The image considered in this experiment consists of $250 \times 191$ pixels of $n_{\lambda} =188$ spectral bands 
obtained after removing the water vapor absorption bands.
A composite color image of the scene of interest is shown in the left of Fig. \ref{fig:HS_Cuprite}.
According to \cite{Nascimento2005}, the number of endmembers has been set to $m=14$. 
The estimated endmember signatures from VCA, LU and NLU are displayed in Fig. \ref{fig:End_Cuprite} and 
the $m=14$ corresponding abundance maps estimated using LU and NLU are shown in Figs. \ref{fig:Real_Abu_Cuprite_LU}
and \ref{fig:Real_Abu_Cuprite_NLU}. Visually, VCA, LU and NLU provide similar endmembers
and abundance maps.
The estimated map of probability $\bfP$ from NLU and the differences between the estimated LU and NLU
abundance maps are shown in Figs. \ref{fig:HS_Cuprite} (middle and right).
Again, the areas with large differences in the abundance maps
overlap with the ones with large nonlinear effects. Nonlinear effects can be expected in this scenario as there exist 
intimate mixtures, in which a light ray can interact many times with the different mineral grains.

\section{Conclusion}
This paper studied a new unsupervised nonlinear spectral unmixing method based on 
a recent multilinear model. The nonlinear unmixing problem was formulated
as a constrained optimization problem with respect to the endmembers, abundances and
pixel-dependant transition probabilities. A gradient projection within block coordinate descent method 
was proposed to estimate these three sets of variables jointly. Each step
of the proposed method was carefully addressed to guarantee the convergence to a stationary point
of the objective function. 
Experiments implemented on both synthetic and real datasets confirmed that 
the multilinear model based nonlinear unmixing allows us to detect and analyze 
the non-linearities present in a hyperspectral image.
Further work will be devoted to investigate a similar nonlinear unmixing algorithm
accounting for spatial correlations for the abundances and the transition probabilities.

\section*{Acknowledgments}
This work was supported by DSO Singapore, and in part by the HYPANEMA ANR Project under Grant ANR-12-BS03-003, and the
Thematic Trimester on Image Processing of the CIMI Labex, Toulouse, France, under Grant ANR-11-LABX-0040-CIMI within the Program ANR-11-IDEX-0002-02.
The authors would also like to thank Rob Heylen for insightful discussion on the design of the multilinear model.

\bibliographystyle{ieeetran}
\bibliography{strings_all_ref,biblio_all}
\end{document}